%% file: main.tex
\documentclass[a4paper,10pt,conference]{ieeeconf}
\usepackage{FG2027}
\input{preamble}

\input{paper-info}

\FGfinalcopy
\IEEEoverridecommandlockouts
\title{\LARGE \bf
\PaperTitle
}
\author{\PaperAuthors}

\begin{document}
\pagestyle{empty}
\maketitle
\thispagestyle{empty}

\input{sec/0_abstract}
\input{sec/1_intro}
\input{sec/2_related}
\input{sec/3_method}
\input{sec/4_experiments}
\input{sec/5_results}
\input{sec/6_conclusion}

\FloatBarrier
{\small
\bibliographystyle{ieee}
\bibliography{main}
}
\clearpage
\useRomanappendicesfalse
\appendices
\input{sec/7_appendix}
\clearpage
\input{sec/8_appendix_tables}
\input{sec/9_appendix_figures}
\end{document}

%% file: preamble.tex
\usepackage{graphicx}
\usepackage{placeins}
\usepackage{booktabs}
\usepackage{amsmath,amssymb}
\usepackage{xspace}
\usepackage{xcolor}

\usepackage{caption}
\renewcommand{\thetable}{\arabic{table}}
\makeatletter\def\fnum@table{Table~\thetable}\makeatother
\usepackage{cite}

\usepackage{url}
\makeatletter\let\NAT@parse\undefined\makeatother
\definecolor{linkblue}{rgb}{0.21,0.49,0.74}
\usepackage[colorlinks,breaklinks,allcolors=linkblue]{hyperref}

%% file: paper-info.tex
\def\FGPaperID{186}
\newcommand{\PaperTitle}{S2HAND: Accuracy-Preservation Trade-Offs for Hand Specialization in Whole-Body Pose Models}
\newcommand{\PaperAuthors}{%
Singyu Muk$^{1}$, Xuanren Song$^{2}$, and Kalin Stefanov$^{3,*}$%
\thanks{$^{1}$Singyu Muk is an Independent Researcher.
{\tt\small ronggong000@gmail.com}}%
\thanks{$^{2}$Xuanren Song is with Tsinghua University.
{\tt\small kowdataanalyze@gmail.com}}%
\thanks{$^{3}$Kalin Stefanov is with Japan Advanced Institute of Science and Technology.
{\tt\small kalins@jaist.ac.jp}}.
$^{*}$Corresponding author.}%

%% file: sec/0_abstract.tex
\begin{abstract}

Unified whole-body pose estimators jointly predict body, face, and hand landmarks, but often degrade on small, blurred, or occluded hands. Attaching a standalone hand specialist introduces extra detection overhead, coordinate alignment, and wrist-stitching artifacts. We introduce S2HAND, a hand-specialized Sapiens2-5B framework that preserves the native 308-keypoint output interface. Specifically, S2HAND-MT adapts a lightweight 0.87M-parameter pose branch to route 40 finger heatmaps while strictly preserving the shared Transformer, auxiliary task heads, and the other 268 body, foot, and face heatmaps. S2HAND-Pose further fine-tunes three Fisher-selected Transformer blocks (267.13M parameters), trading exact preservation for stronger hand recovery: it lifts macro AUC from 57.83\% to 65.73\% across HInt TEST, COCO WholeBody VAL, and FreiHAND EVAL, and surpasses HInt-adapted Hand5/HRNet and stitched Hand5 systems on all five Native Pipeline benchmarks. On COCO, S2HAND-Pose improves hand AP by 12.89 points and body-plus-foot AP by 0.19 points. Across a 512-image multi-task audit, it maintains 99.60\% foreground segmentation agreement with the public checkpoint, with quantified normal and pointmap drift. Its 2D coordinates also lower 3D mesh error in frozen Pose2Mesh transfer, while our optional TailGuard re-cropping module cuts S2HAND-MT's severe failure rate on NewDays from 13.45\% to 10.43\%. S2HAND thus resolves the whole-body hand bottleneck with either exact non-hand preservation (MT) or empirically audited trade-offs (Pose).

\end{abstract}

%% file: sec/1_intro.tex
\section{Introduction}
\label{sec:intro}

Whole-body pose models predict body, foot, face, and hand landmarks in one pass, providing a unified interface for full-person perception. Yet hands remain a bottleneck: they are small, often blurred or occluded, and have densely packed joints. A whole-body model may therefore capture the body well while failing on the hands.

A common workaround is to attach a standalone hand model. However, this requires an extra detector and crop pipeline, followed by aligning coordinates, handedness, confidences, and joint order. The wrist also creates a seam: a body wrist may misalign with the hand model's fingers, while a hand wrist may disconnect from the elbow. Instead, we improve hand estimation directly within a single whole-body model.

We build S2HAND\footnote{Code, checkpoints, and dataset resources: \url{https://github.com/ronggong000/S2HAND}.} on Sapiens2-5B~\cite{khirodkar2026sapiens2}, which predicts 308 whole-body keypoints. For deployments requiring unchanged non-hand outputs, S2HAND-MT trains a small pose branch and replaces only the 40 finger heatmaps. The other 268 heatmaps, including both wrists, follow the frozen public path, leaving all other task heads untouched. S2HAND-Pose shares this routing and also tunes three Transformer blocks selected by a training-only Fisher probe. It yields higher hand accuracy within the same 308-point interface, though non-hand preservation is audited empirically rather than guaranteed. We also introduce TailGuard, an optional re-crop policy for low-confidence S2HAND-MT predictions.

S2HAND-Pose improves Native Pipeline AUC across all five sets over Public Sapiens2, HInt-adapted Hand5/HRNet, and stitched Hand5 systems, while EffHandEgoNet remains stronger on H2O. We audit body, face, segmentation, normal, and pointmap outputs, and test frozen 3D and recognition consumers.
Our contributions are:
\begin{itemize}
\item Two whole-body adaptations with distinct accuracy--preservation trade-offs: exact protected routing in S2HAND-MT and Fisher-guided block tuning in S2HAND-Pose;
\item TailGuard, a conditional re-crop policy that reduces severe S2HAND-MT errors; and
\item Evaluations across datasets, crop protocols, protected outputs, and downstream tasks, alongside code, checkpoints, and the paired 2D/3D H2SMV test set.
\end{itemize}

%% file: sec/2_related.tex
\section{Related Work}

\noindent \textbf{Hand and whole-body pose estimation.}
Hand pose estimation must recover fine articulation under self-occlusion, object occlusion, and motion blur~\cite{doosti2019handsurvey,moon2020interhand}.
Hand-specific methods such as RTMPose and HRNet crop the hand region before applying a dedicated estimator~\cite{jiang2023rtmpose,sun2019hrnet}.
HandOS unifies detection, 2D pose, and 3D reconstruction in one system, while ExtPose incorporates 2D-pose cues and temporal context~\cite{chen2025handos,chen2025extpose}.
Because these methods rely on different inputs and cropping rules, we separate their published scores from controlled reruns.
Benchmark datasets including FreiHAND, HInt, H2O, DexYCB, and AssemblyHands span single-hand, in-the-wild occlusion, hand--object, and egocentric settings~\cite{zimmermann2019freihand,pavlakos2024hamer,kwon2021h2o,chao2021dexycb,ohkawa2023assemblyhands}.

Whole-body methods predict body, feet, face, and hands through a unified keypoint interface~\cite{cao2017openpose,jin2020wholebody}.
Regional refinement, multi-scale features, distillation, and efficient coordinate heads help bridge the scale gap between the torso and small extremities~\cite{yang2023dwpose,jiang2023rtmpose,jiang2024rtmw}.
Sapiens and Sapiens2 scale this design with high-resolution Transformers and task-specific heads~\cite{khirodkar2024sapiens,khirodkar2026sapiens2}.
S2HAND retains the Sapiens2 whole-body interface rather than replacing it with a standalone hand model.

\noindent \textbf{Efficient specialization and output preservation.}
Prompt tuning, adapters, and LoRA adapt pretrained models while freezing most weights~\cite{jia2022vpt,chen2022adaptformer,hu2022lora}.
Gradient- and Fisher-based methods identify sensitive parameters for targeted updates or protection~\cite{kirkpatrick2017ewc,sung2021fishmask,he2023spt}.
Continual and multi-task learning instead mitigate forgetting or gradient conflict via regularization, distillation, or gradient projection~\cite{li2016lwf,hinton2015distilling,yu2020pcgrad}.
While these approaches reduce tuning cost or Empirical forgetting, they cannot guarantee identical outputs once shared features change.
S2HAND-MT enforces exact preservation via deterministic routing, whereas S2HAND-Pose updates shared blocks and is audited empirically.

\noindent \textbf{Localization and conditional computation.}
Top-down pose accuracy depends heavily on detector quality and crop context~\cite{xiao2018simple,jiang2023rtmpose}.
Confidence-aware cascades allocate extra compute only to hard samples~\cite{enomoro2021cascade,teerapittayanon2016branchynet}.
TailGuard adapts this principle via a side-specific re-crop on low-confidence hand predictions, reporting detector overhead and failure rates separately.

\noindent \textbf{2D-conditioned downstream transfer.}
Action recognition, sign language, 3D lifting, and mesh recovery models often consume predicted 2D joints~\cite{yan2018stgcn,li2025unisign,martinez2017simple3d}.
While RGB mesh regressors infer geometry directly from images~\cite{kanazawa2018hmr}, Pose2Mesh maps 2D skeletons to mesh vertices~\cite{choi2020pose2mesh}.
We use frozen 2D-driven consumers to test whether improved 2D hand coordinates transfer without downstream retraining.

%% file: sec/3_method.tex
\section{Method}
\label{sec:method}

\begin{figure}[t]
\centering
\includegraphics[width=\linewidth]{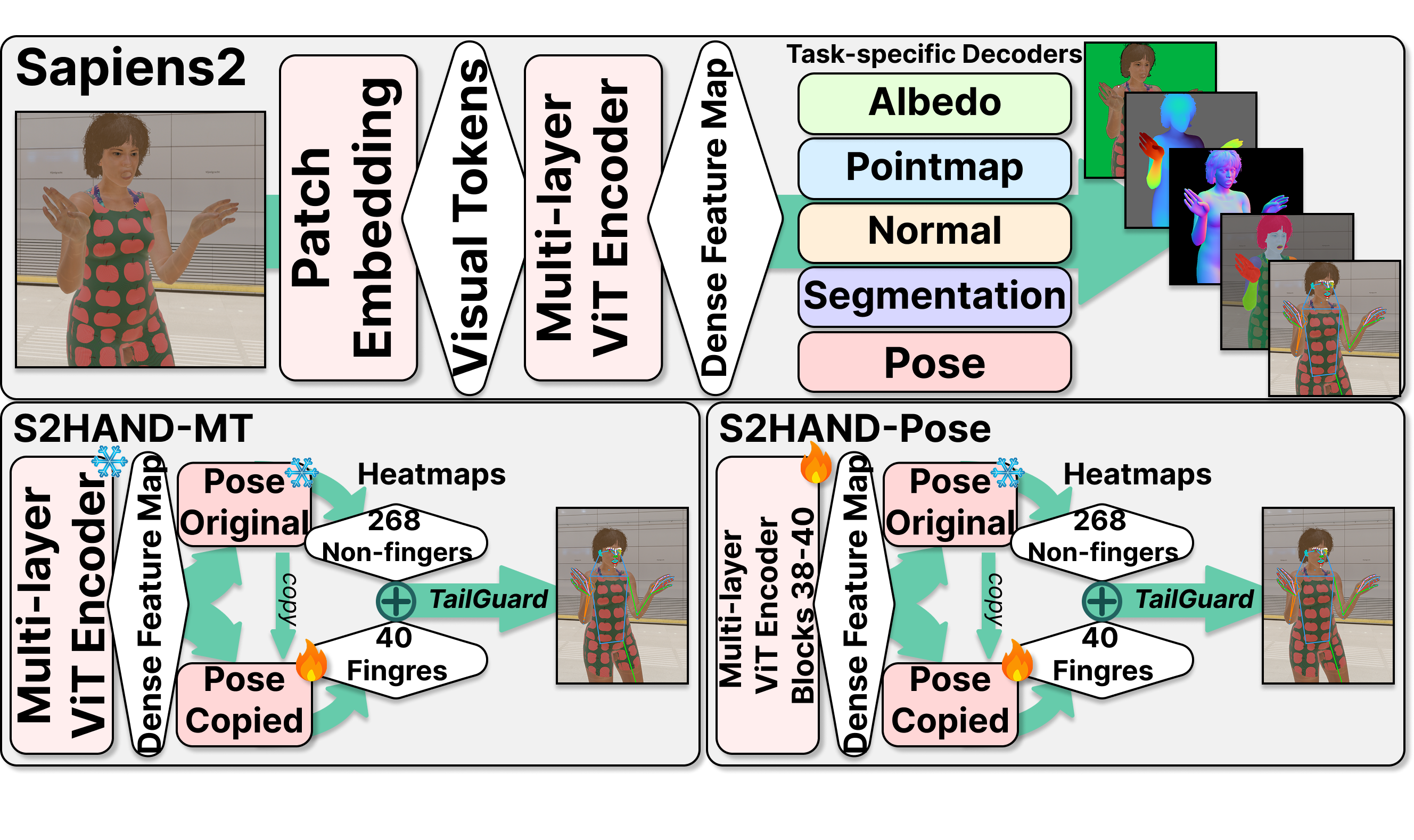}
\caption{\textbf{S2HAND system overview.} S2HAND-MT duplicates the lightweight pose decoder and projection to route 40 adapted finger heatmaps into the frozen public output. S2HAND-Pose additionally fine-tunes Transformer blocks 38--40. TailGuard optionally re-crops low-confidence hands. Snowflakes and flames denote frozen and trainable components.}
\label{fig:s2hand-system}
\end{figure}

We build S2HAND on the public Sapiens2-5B model~\cite{khirodkar2026sapiens2}. Given a person crop \(x\), its patch embedding and 56-block Transformer extract shared features \(h=B(E(x))\), which feed into task-specific heads. The pose head outputs 308 heatmaps covering the body, feet, face, and hands. We denote the 40 non-wrist finger channels by \(\mathcal H\) and the remaining 268 channels, including both wrists, by \(\mathcal P\). Figure~\ref{fig:s2hand-system} summarizes the architecture.

\subsection{S2HAND-MT: Protected Hand Adaptation}
\label{sec:s2hand-mt}

S2HAND-MT freezes the patch embedding, Transformer, public pose head, and pose deconvolution stack \(U_0\). From the dense feature map \(f=U_0(B(E(x)))\), the public path produces \(Z^0=P_0(C_0(f))\). We duplicate only the three-layer \(1\times1\) decoder \(C_0\) and its 308-channel projection \(P_0\) to form an adapted branch \(Z^a=P_a(C_a(f))\). Initialized from the public weights, this branch adds only 0.867M trainable parameters without duplicating the backbone or deconvolution stack.

While both branches output 308 heatmaps, the final prediction routes only the adapted finger channels:

\begin{equation}
Z^{\mathrm{MT}}_c =
\begin{cases}
Z^a_c, & c\in\mathcal H,\\
Z^0_c, & c\in\mathcal P.
\end{cases}
\label{eq:mt-routing}
\end{equation}
Consequently, the 40 finger heatmaps are updated while all 268 protected heatmaps come directly from the public branch in a single forward pass. For \(c\in\mathcal P\), \(Z^{\mathrm{MT}}_c=Z^0_c\) holds by construction prior to coordinate decoding, without relying on a preservation loss. Moreover, the frozen Transformer supplies identical shared features to all other task heads.

Training supervises the 40 adapted finger channels alongside the two detached public wrist channels, which receive no gradient but anchor the five wrist-to-MCP edges. We optimize:

\begin{equation}
\mathcal L_{\mathrm{hand}}
=\mathcal L_{\mathrm{hm}}
+\lambda_{\mathrm{coord}}\mathcal L_{\mathrm{coord}}
+\lambda_{\mathrm{topo}}\mathcal L_{\mathrm{topo}},
\label{eq:hand-loss}
\end{equation}
where \(\mathcal L_{\mathrm{hm}}\) is a weighted heatmap MSE. A differentiable soft coordinate decoder is applied only during training: \(\mathcal L_{\mathrm{coord}}\) computes Smooth-L1 loss on scale-normalized joint coordinates, and \(\mathcal L_{\mathrm{topo}}\) applies Smooth-L1 loss to normalized 2D bone vectors. The auxiliary weights are calibrated via gradient norms rather than a validation sweep. Appendix~\ref{app:architecture} details the validity masks, decoder temperature, thresholds, and exact weights.

At inference, S2HAND-MT discards the training-only soft decoder and applies the standard UDP coordinate transform and DARK heatmap decoder~\cite{huang2020udp,zhang2020dark}, preserving the native 308-keypoint ordering.

\input{tables/tab_p2a_native_pipeline}

\subsection{S2HAND-Pose: Fisher-Guided Shared Adaptation}
\label{sec:fisher-selection}
\label{sec:s2hand-pose}

Because a frozen Transformer may lack fine-grained features for small or occluded fingers, we score each Transformer block via a diagonal-Fisher proxy~\cite{sung2021fishmask,he2023spt}. For block parameters \(\theta_b\),

\begin{equation}
F_b=\frac{1}{N}\sum_{i=1}^{N}
\left\|\nabla_{\theta_b}\mathcal L_{\mathrm{hand}}(\mathcal B_i)\right\|_2^2,
\qquad N=8,
\label{eq:fisher}
\end{equation}
where \(\mathcal B_i\) denotes fixed training batches. All blocks are evaluated on the exact same batches without optimizer steps or validation/test feedback. Blocks 38, 40, and 39 rank highest and form the contiguous tuning window 38--40. Terminal-block and non-contiguous controls are compared in Sec.~\ref{sec:adaptation-results}.

S2HAND-Pose retains the MT routing interface (Eq.~\ref{eq:mt-routing}) and hand loss (Eq.~\ref{eq:hand-loss}), while jointly tuning blocks 38--40 alongside the adapted decoder and projection (267.13M trainable parameters in total). The patch embedding, remaining 53 Transformer blocks, public deconvolution stack, and public decoder stay frozen.

Although protected channels still pass through the frozen public decoder, their upstream Transformer features shift. Thus, S2HAND-Pose preserves the 308-keypoint interface but relaxes the exact invariance guarantee of S2HAND-MT. Similarly, auxiliary task heads retain their frozen weights but receive updated backbone features; we audit their empirical preservation in Sec.~\ref{sec:preservation}.

\subsection{TailGuard: Conditional Hand Re-Cropping}
\label{sec:tailguard}

TailGuard is an optional inference policy for S2HAND-MT that begins with standard whole-person prediction. For each hand \(s\in\{\text{left},\text{right}\}\), let \(r_{s,j}\) be the confidence of joint \(j\) and \(b_s\) be the same-side hand box from a frozen WiLoR-YOLO detector~\cite{potamias2025wilorendtoend3dhand}. Re-cropping triggers via:

\begin{equation}
\begin{aligned}
c_s&=\frac{1}{21}\sum_{j=1}^{21}r_{s,j},\\
g_s&=\mathbf1[c_s\leq\tau\ \land\ b_s\text{ exists}],
\qquad \tau=0.510756,
\end{aligned}
\label{eq:tailguard}
\end{equation}
where \(\tau\) is selected on HInt-EPICK VAL~\cite{pavlakos2024hamer} and frozen. When \(g_s=1\), \(b_s\) is expanded by \(2\times\), the same S2HAND-MT checkpoint is re-applied to that crop, and all 21 predicted hand joints are mapped back to replace the native hand output; otherwise, the native prediction is retained. TailGuard requires neither GT boxes nor a separate keypoint model.

Because TailGuard modifies final coordinates via a conditional second crop rather than altering model weights, the exact native-route guarantee strictly applies to S2HAND-MT with TailGuard disabled. We report performance with TailGuard ON and OFF, detector overhead, and conditional second-pass latency separately.

%% file: tables/tab_p2a_native_pipeline.tex
\begin{table*}[t]
\centering
\scriptsize
\setlength{\tabcolsep}{3pt}
\caption{\textbf{Native Pipeline evaluation.} Each cell reports AUC@.20 $\uparrow$ / severe-tail rate $\downarrow$ (\%). All reruns include their declared detector and crop; missed hands count as failures. The two HInt-FT controls use 12,000 balanced HInt TRAIN hands; their short adaptation budget is detailed in Appendix~\ref{app:matched-training}. Bottom rows report paired S2HAND-Pose AUC gains with 95\% cluster-bootstrap intervals (10,000 replicates), computed before rounding. EPICK is VAL; H2SMV-FPV is synthetic TEST. TailGuard is off. Full PCK values, sample counts, and published HandOS results are in Appendix~\ref{app:protocols}.}
\label{tab:p2a-native}
\begin{tabular*}{\textwidth}{@{\extracolsep{\fill}}llllll@{}}
\toprule
\shortstack{\textbf{System}} & \shortstack{\textbf{NewDays (TEST)}} & \shortstack{\textbf{VISOR (TEST)}} & \shortstack{\textbf{H2O (TEST)}} & \shortstack{\textbf{EPICK (VAL)}} & \shortstack{\textbf{H2SMV-FPV (synthetic TEST)}}\\
\midrule
Public Sapiens2-5B~\cite{khirodkar2026sapiens2} & 61.48 / 14.14 & 56.12 / 17.68 & 69.02 / 4.38 & 55.16 / 18.56 & 47.93 / 29.86\\
RTMPose-Hand5 (public)~\cite{jiang2023rtmpose} & 59.72 / 16.59 & 63.51 / 10.86 & 67.43 / 0.43 & 64.10 / 8.48 & 40.57 / 39.00\\
RTMPose-Hand5 + HInt FT~\cite{jiang2023rtmpose} & 61.57 / 17.33 & 69.85 / 10.02 & 72.57 / 0.30 & 71.01 / 6.40 & 39.78 / 39.64\\
FreiHAND-HRNet reproduction~\cite{sun2019hrnet} & 17.62 / 79.19 & 28.32 / 54.09 & 60.19 / 7.29 & 27.84 / 55.20 & 26.66 / 61.59\\
FreiHAND-HRNet + HInt FT~\cite{sun2019hrnet} & 39.42 / 40.31 & 56.57 / 18.57 & 68.74 / 1.80 & 57.14 / 15.20 & 35.27 / 47.49\\
EffHandEgoNet-H2O~\cite{mucha2024effhand} & 5.20 / 92.30 & 32.25 / 42.55 & \textbf{75.25} / \textbf{0.06} & 29.11 / 45.60 & 8.37 / 84.97\\
S2HAND-MT (Ours) & 62.57 / 13.45 & 59.42 / 16.21 & 70.79 / 3.63 & 58.54 / 16.96 & 49.01 / 29.09\\
S2HAND-Pose (Ours) & \textbf{70.04} / \textbf{6.61} & \textbf{74.88} / \textbf{4.30} & 73.96 / 0.49 & \textbf{74.35} / \textbf{4.00} & \textbf{56.14} / \textbf{17.14}\\
\midrule
$\Delta$ vs Public [95\% CI] & $+8.56\ [7.88,9.29]$ & $+18.75\ [17.16,20.39]$ & $+4.94\ [4.11,5.84]$ & $+19.19\ [17.32,21.04]$ & $+8.21\ [7.87,8.55]$\\
$\Delta$ vs Hand5 + HInt FT [95\% CI] & $+8.48\ [7.27,9.72]$ & $+5.02\ [3.66,6.25]$ & $+1.40\ [0.91,1.85]$ & $+3.35\ [1.92,5.04]$ & $+16.36\ [15.68,16.98]$\\
\bottomrule
\end{tabular*}
\end{table*}

%% file: sec/4_experiments.tex
\section{Experimental Setup}
\label{sec:experimental-setup}

\noindent\textbf{Data and input protocols.}
We train S2HAND on a $1{:}1$ balanced mixture of HInt NewDays TRAIN and EPICK TRAIN~\cite{pavlakos2024hamer}. Checkpoints are selected on HInt VAL, while TailGuard is tuned on EPICK VAL. NewDays TEST and VISOR TEST serve as held-out subsets within the HInt family rather than strictly out-of-domain tests. H2O~\cite{kwon2021h2o}, FreiHAND~\cite{zimmermann2019freihand}, AssemblyHands~\cite{ohkawa2023assemblyhands}, DexYCB~\cite{chao2021dexycb}, and synthetic H2SMV-FPV are never seen during S2HAND training or selection (see Appendix~\ref{app:protocols} for dataset roles).

\emph{Native Pipeline} evaluates each model with its deployable detector and crop, counting missed hands as failures. \emph{Shared-GT Crop} supplies the same side-specific $2\times$ GT-hand ROI to all models, whereas \emph{Shared-Detector Crop} shares WiLoR-YOLO $2\times$ boxes~\cite{potamias2025wilorendtoend3dhand}. These three protocols are never pooled, and TailGuard remains disabled unless stated otherwise. Stitched baselines pair Public Sapiens2 with external Hand5 joints under three wrist policies: replacing all 21 joints, retaining the public wrist, or retaining the public wrist while translating the external fingers. Specialist misses fall back to Public Sapiens2, whereas public person-detection misses remain failures.

\noindent\textbf{Baselines and adaptation.}
We compare against Public Sapiens2-5B, public RTMPose-Hand5~\cite{jiang2023rtmpose}, EffHandEgoNet-H2O~\cite{mucha2024effhand}, and HRNet~\cite{sun2019hrnet}. We further adapt public Hand5 and our reproduced FreiHAND-HRNet on 12,000 balanced HInt TRAIN hands (12k effective sample updates at batch size 32, FP16, seed 42), selecting the best checkpoint on HInt VAL prior to early stopping. These serve as matched-data controls for training-domain access rather than equal-compute counterparts to S2HAND's multi-stage pretraining. Because official HandOS code and weights are unavailable, we list its published PCK~\cite{chen2025handos} separately from our controlled reruns. Released Pose2Mesh HRNet coordinates~\cite{choi2020pose2mesh} are included solely as a FreiHAND reference.

Both S2HAND variants train for 18k updates from a shared HInt-adapted initialization using AdamW, BF16, and a $2\times10^{-5}$ decoder learning rate; S2HAND-Pose additionally updates blocks 38--40 at $2\times10^{-6}$. We evaluate three MT seeds and two Pose seeds. All Sapiens2 variants share identical normalization, joint mapping, UDP/DARK decoding, and protocol-specific crop transforms (schedules and learning curves in Appendix~\ref{app:matched-training}).

\noindent\textbf{Metrics and preservation.}
Joint errors are normalized by the maximum side length of the tight GT-hand box. PCK measures the fraction of valid joints within a threshold, and AUC@.20 integrates PCK across 201 thresholds over $[0,.20]$. Severe-tail rate records the proportion of hands with valid-joint mean NME $>.20$, counting missed hands as failures. Paired 95\% confidence intervals use 10,000 cluster-bootstrap replicates, resampling at the video or sequence level where available. Our dense-head audit evaluates 512 fixed COCO images by transferring the learned block 38--40 updates into each released task model with its head frozen, measuring agreement against the public model on a $256\times192$ grid. This quantifies teacher agreement rather than GT accuracy (full definitions in Appendix~\ref{app:preservation}).

%% file: sec/5_results.tex
\section{Results}
\label{sec:results}

\subsection{Hand Pose Accuracy}
\label{sec:hand-pose-results}

Table~\ref{tab:p2a-native} reports Native Pipeline AUC and severe-tail rates, including both HInt-adapted specialists. S2HAND-Pose improves over Public Sapiens2 across all five sets: AUC rises from 61.48 to 70.04 on NewDays and from 56.12 to 74.88 on VISOR, while severe tails drop from 14.14\% to 6.61\% and from 17.68\% to 4.30\%. Paired AUC intervals exclude zero against both Public Sapiens2 and matched Hand5 on every set; these intervals reflect test-sample uncertainty for the evaluated checkpoints rather than training-seed variance.

\noindent\textbf{Training-data fairness.}
Corrected HInt adaptation raises Hand5 AUC from 59.72 to 61.57 on NewDays and from 63.51 to 69.85 on VISOR, while HRNet improves from 17.62 to 39.42 and from 28.32 to 56.57. We replace the earlier collapsed Hand5 control rather than attributing its failure to domain shift (validation curves and the corrected protocol appear in Appendix~\ref{app:matched-training}). Using unrounded predictions, S2HAND-Pose remains ahead of matched Hand5 by 8.48 AUC points on NewDays and 5.02 on VISOR. Thus, the accuracy gains hold even after fine-tuning both external specialists on HInt to validation convergence.
\input{tables/tab_shared_gt_matched}

Shared-GT results in Table~\ref{tab:shared-gt-matched} isolate keypoint regression from detector differences. While HInt adaptation lifts Hand5 AUC from 57.30 to 60.21 on NewDays and from 59.19 to 65.86 on VISOR, S2HAND-Pose reaches 74.27 and 78.34. Its regression advantage therefore persists under identical crops and full coverage (see Appendix~\ref{app:protocols} for full Native PCK, published HandOS scores, and Shared-Detector results).

\begin{figure*}[t]
\centering
\includegraphics[width=\textwidth]{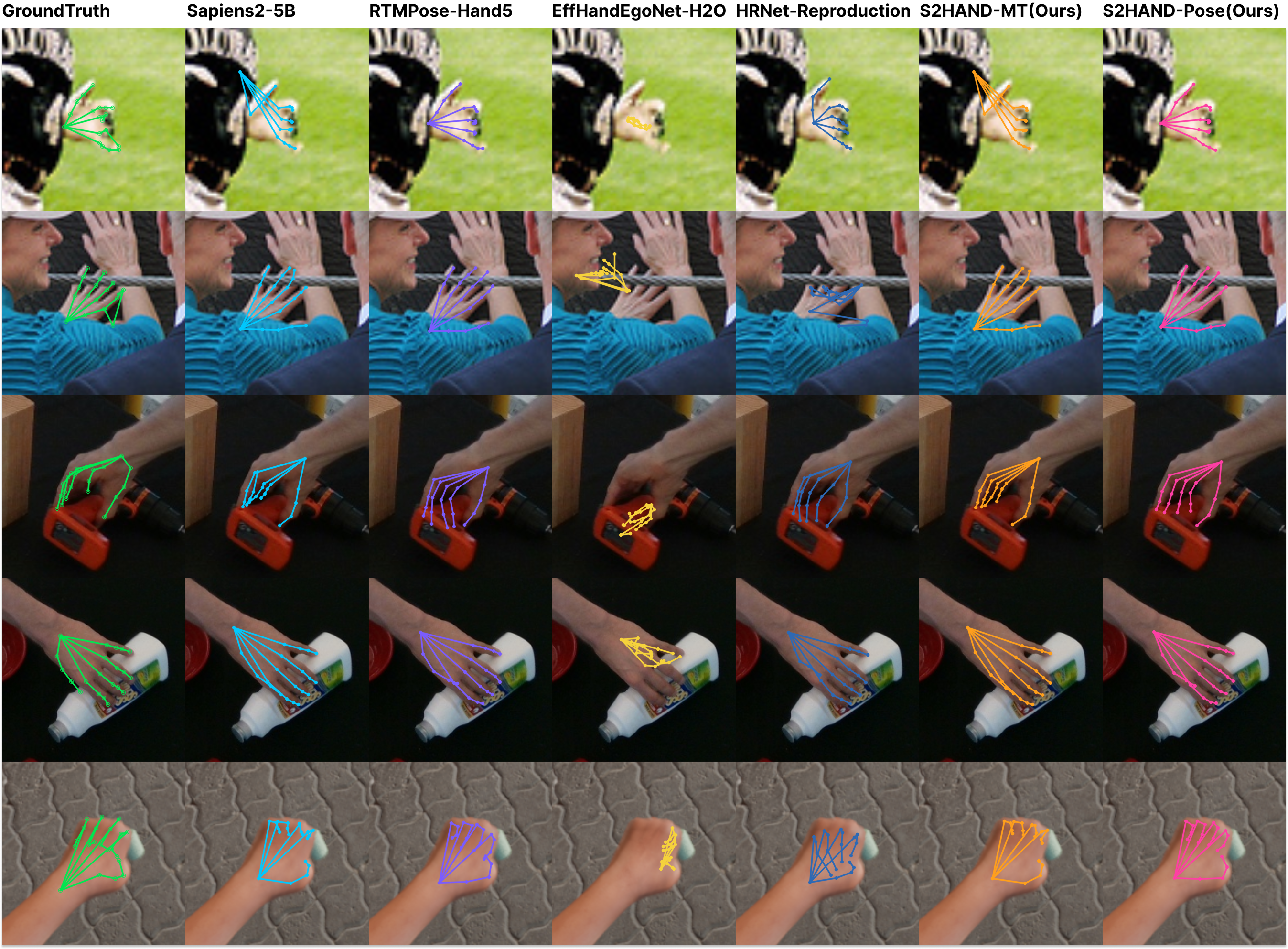}
\caption{\textbf{Qualitative hand-pose comparison.} We show two COCO WholeBody, two DexYCB, and one held-out synthetic H2SMV-FPV example. Each row uses the same image and annotated hand across models. Further datasets and balanced success/failure cases appear in Appendix~\ref{app:hand-results}.}
\label{fig:hand-qual-main}
\end{figure*}

On H2O, S2HAND-Pose attains 73.96 AUC, trailing the H2O-dedicated EffHandEgoNet (75.25). It achieves 74.35 on EPICK VAL and 56.14 on synthetic H2SMV-FPV TEST; neither split is used to claim held-out real-data SOTA. Notably, matched Hand5 dips slightly on H2SMV (40.57 to 39.78) despite gains on the real sets. Published PCK numbers serve as external references rather than controlled reruns: ExtPose reports higher scores for its ViTPose baseline~\cite{chen2025extpose,xu2022vitpose}, where matching detector, crop, and averaging rules remain unverified. Appendix~\ref{app:protocols} separates these reference scores and quantifies the impact of PCK aggregation. Figure~\ref{fig:hand-qual-main} provides qualitative comparisons on COCO WholeBody, DexYCB, and held-out synthetic H2SMV-FPV examples.

\subsection{Integrated and Stitched Prediction}
\label{sec:stitched-results}

\input{tables/tab_stitched_baseline}

Table~\ref{tab:stitched-baseline} directly evaluates the external-specialist stitching alternative across all three wrist policies for both public and HInt-adapted Hand5. While the strongest matched stitch narrows the gap to 1.43 AUC points on VISOR, 1.21 on EPICK, and 1.37 on H2O, S2HAND-Pose remains higher across all five sets (e.g., 70.04 vs.\ 64.96 on NewDays). These comparisons use point estimates; the Native Hand5 confidence intervals in Table~\ref{tab:p2a-native} are not reused for the stitched systems.

\begin{figure*}[t]
\centering
\includegraphics[width=.98\textwidth]{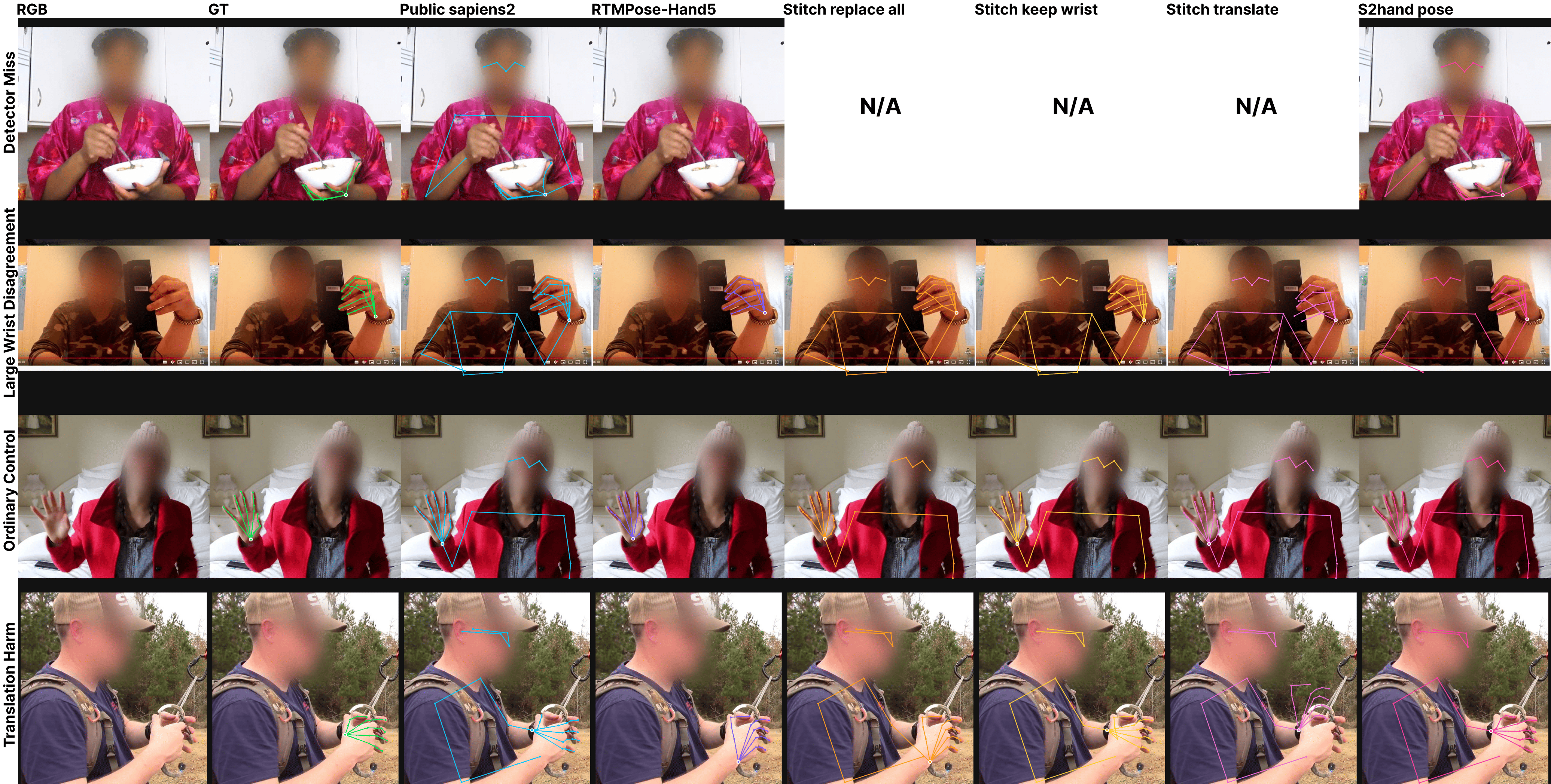}
\caption{\textbf{Hand-specialist stitching choices.} Columns: input image, hand GT, Public Sapiens2-5B, public RTMPose-Hand5~\cite{jiang2023rtmpose}, replace all 21 joints, keep the public wrist, translate external fingers to the public wrist, and S2HAND-Pose (Ours). Rows illustrate a detector miss, wrist disagreement, an ordinary control, and translation drift. Body skeletons are overlaid for Public, Pose, and stitched variants, whereas GT and standalone Hand5 show hands only. N/A indicates unavailable stitching after a specialist miss (quantitative evaluation falls back to Public Sapiens2). TailGuard is disabled.}
\label{fig:hand5-stitching}
\end{figure*}

Figure~\ref{fig:hand5-stitching} illustrates why wrist policy matters: replacing all joints displaces the body wrist; keeping the public wrist distorts wrist-to-finger bone connections; and translating the external fingers preserves internal hand geometry but shifts their image alignment. Although the ordinary control row confirms that stitching can succeed on clean inputs, the failure rows isolate specific failure mechanisms rather than their overall prevalence (see Appendix~\ref{app:stitched-details} for WiLoR coverage and wrist-gap diagnostics).

\subsection{Adaptation Scope and Fisher Selection}
\label{sec:adaptation-results}

\input{tables/tab_adaptation_ablation}

The decoder controls share identical initialization and training budgets. Ordinary head tuning and S2HAND-MT reach 68.47 and 68.45 HInt VAL AUC, indicating that routing provides exact output protection rather than an accuracy gain. Across MT seeds 42--44, HInt VAL AUC is $68.422\pm0.063$ and HInt TEST is $67.446\pm0.055$. Because the 0.02-point gap from ordinary tuning falls within this seed variance, we claim neither an accuracy advantage nor formal statistical equivalence; S2HAND-MT simply serves deployments that require strictly unchanged protected outputs.

\begin{figure}[t]
\centering
\includegraphics[width=\linewidth]{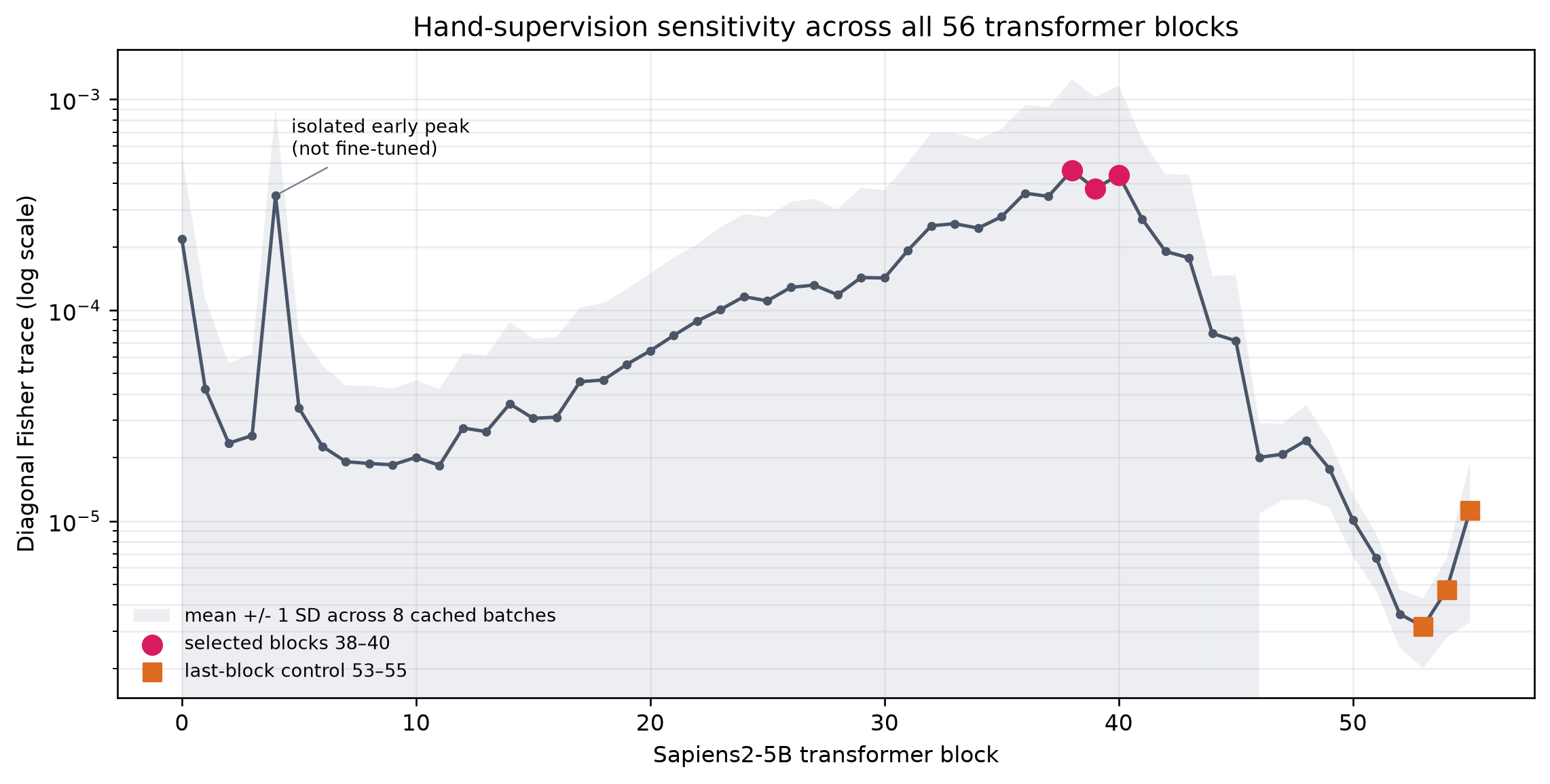}
\caption{\textbf{Transformer-block sensitivity.} Mean diagonal-Fisher trace across eight fixed training batches (shaded region: $\pm1$ SD). Blocks 38, 40, and 39 rank highest and form the contiguous window 38--40; block 4 and blocks 53--55 serve as controls.}
\label{fig:fisher-profile}
\end{figure}

Figure~\ref{fig:fisher-profile} shows the block sensitivity profile. Tuning block 38 alone yields 75.76 HInt VAL AUC, whereas blocks 38--40 reach 76.56. In contrast, the terminal blocks 53--55 attain only 72.15 despite more trainable parameters (284.89M vs.\ 267.13M for Pose). While the non-contiguous $\{4,38,40\}$ control slightly edges out blocks 38--40 in three-set hand AUC (65.86 vs.\ 65.73), it degrades face NME (0.1543 vs.\ 0.1510) and PCK@.05 (50.73\% vs.\ 53.70\%). We therefore select blocks 38--40 for their superior accuracy--preservation balance rather than raw hand AUC alone. A second Pose seed achieves 76.50 HInt VAL AUC and a 65.55 three-set mean, closely matching the main run (76.56 and 65.73; full rankings, seed runs, and face controls in Appendix~\ref{app:adaptation}). 

\subsection{Whole-Body and Multi-task Preservation}
\label{sec:preservation}

\input{tables/tab_preservation}
\input{tables/tab_pose2mesh_downstream}
\input{tables/tab_h2o_action}
\input{tables/tab_model_cost}
Because the original task-specific training sets and native annotations are unavailable, Table~\ref{tab:preservation} separates compatible human-GT accuracy from public-model agreement. S2HAND-MT routes 268 heatmaps, including both wrists and all face channels, through the frozen public branch while leaving the shared encoder and other task heads untouched, guaranteeing exact non-hand preservation by construction.

On COCO, S2HAND-Pose changes body+foot AP from 79.09 to 79.28 while lifting hand AP from 51.92 to 64.81. S2HAND-MT achieves 54.75/60.41 hand AP/AR while preserving the public body, foot, and face outputs by construction (left/right hand breakdowns in Appendix~\ref{app:preservation}).

Our expanded multi-task audit evaluates segmentation, surface normals, and pointmaps on the same 512 COCO images. S2HAND-Pose retains 99.5956\% foreground segmentation agreement and 99.3246\% hand-mask IoU, with $0.6118^\circ$ mean normal angular drift and 0.4791\% mean normalized pointmap endpoint drift. While average agreement is high, pointmap drift exhibits a heavy tail: five images exceed 5\% mean drift, peaking at 11.8552\%. These metrics quantify teacher-relative drift rather than GT accuracy. Because the official release provides matting only at 1B and no albedo checkpoint, a matching 5B audit is unavailable for those two tasks.

On 2,004 COCO instances across 16 shared semantic face anchors, results remain mixed: mean NME shifts from 0.1428 to 0.1510, whereas PCK@.05 improves from 49.25\% to 53.70\% and the face-failure rate drops from 12.23\% to 5.99\%. We do not claim human-GT accuracy across all 238 dense face landmarks (full distributions and visualizations in Appendix~\ref{app:preservation}).

\subsection{Downstream Transfer}
\label{sec:downstream}

\noindent\textbf{Frozen 2D-to-3D transfer.}
All drivers feed into the same released FreiHAND-trained Pose2Mesh checkpoint~\cite{choi2020pose2mesh} under identical joint ordering, handedness, normalization, and common-valid samples, varying only the 21 input hand joints. S2HAND-Pose lowers PA-MPJPE relative to Public Sapiens2 across all five sets, with paired 95\% confidence intervals excluding zero. It also outperforms public Hand5 on four sets, whereas Hand5 remains stronger on DexYCB (paired difference $-0.851$ mm, 95\% CI $[-1.087,-0.617]$). Thus, 2D-to-3D transfer is effective across most benchmarks but not universal across all drivers and domains.

This experiment swaps the 2D frontend rather than reproducing the full Pose2Mesh pipeline, and does not claim 3D SOTA. Moreover, GT 2D joints serve as a reference rather than a strict lower bound for a frozen consumer: on AssemblyHands, S2HAND-Pose yields 16.025 mm vs.\ 17.080 mm for GT. Detection coverage is factored out of these common-valid comparisons (sample counts, clustering units, and full intervals in Appendix~\ref{app:downstream}).

\noindent\textbf{Recognition.}
We evaluate the frozen ActionTransformer~\cite{mucha2024effhand} on 122 H2O Subject-3 clips (20 frames per clip) with fixed GT hand-presence and object features. S2HAND-Pose achieves 95.08\% Top-1 accuracy, compared with 94.26\% for Public Sapiens2, 92.62\% for S2HAND-MT, and 95.90\% for EffHandEgoNet. Zeroing hand inputs drops Top-1 to 72.95\%, and shuffling hands within the same object class lowers it to 40.98\%, confirming that the model actively relies on hand trajectories even though paired intervals for S2HAND-Pose include zero.

Similarly, we test frozen Uni-Sign~\cite{li2025unisign} on the 84-video WLASL-2000 overlap~\cite{li2020wlasl} (68 classes), updating only the 2D hand coordinates while keeping confidences and non-hand keypoints fixed. S2HAND-Pose and Public Sapiens2 both reach 71.43\% instance Top-1 (76.72\% vs.\ 75.98\% macro Top-1), with paired intervals again spanning zero. Both recognition evaluations therefore remain inconclusive rather than demonstrating statistically significant downstream gains (full protocols in Appendix~\ref{app:downstream}).

\subsection{System Cost and Optional Recovery}
\label{sec:efficiency-limitations}
\label{sec:tailguard-results}

Table~\ref{tab:model-cost} compares whole-body systems alongside standalone hand estimators. Both S2HAND-MT and S2HAND-Pose add under 1\% to Public Sapiens2's L40S pose-forward latency, whereas stitched systems incur the full whole-body cost plus external detection and hand inference. Serial component latency is at least 331.05 ms with Hand5 (excluding crop, decode, and stitching overhead) and 356.07 ms with WiLoR; neither represents a jointly timed end-to-end deployment measurement (stage boundaries and timing details in Appendix~\ref{app:system-cost}).

As an optional recovery module, TailGuard raises S2HAND-MT AUC on NewDays from 62.57 to 64.55 and cuts severe tails from 13.45\% to 10.43\%, with smaller gains on H2O. In a separate A100 timing audit, mean latency increases from 737.76 to 975.21 ms (+32.2\%) at a 46\% trigger rate, including occasional system stalls (threshold/crop ablations and stage-wise costs in Appendix~\ref{app:tailguard}). The exact MT native-route guarantee holds with TailGuard disabled.

\noindent\textbf{Limitations.}
The 5B backbone is computationally heavier than lightweight standalone hand estimators. Moreover, because the proprietary pretraining and multi-task datasets of Sapiens2 are not publicly released, we cannot co-train on the original data distribution or evaluate ground-truth accuracy on unannotated 238-point face and dense outputs; consequently, S2HAND-Pose relies on teacher-relative agreement and cross-dataset anchor audits, where face metrics remain mixed and pointmap drift shows a minor tail. Finally, frozen recognition transfer remains statistically inconclusive on the evaluated subsets.

%% file: tables/tab_shared_gt_matched.tex
\begin{table}[t]
\centering
\scriptsize
\setlength{\tabcolsep}{3pt}
\caption{\textbf{Matched-data Shared-GT Crop control.} AUC@.20 $\uparrow$ (\%) on the same 1,754 NewDays and 1,906 VISOR TEST hands. All methods receive a side-specific $2\times$ GT-hand crop with 100\% coverage. HRNet + HInt FT starts from the reproduced FreiHAND-HRNet checkpoint. This table isolates landmark regression; it is not a Native Pipeline comparison. TailGuard is off.}
\label{tab:shared-gt-matched}
\begin{tabular*}{\linewidth}{@{\extracolsep{\fill}}lll@{}}
\toprule
\shortstack{\textbf{System}} & \shortstack{\textbf{NewDays (TEST)}} &\shortstack{\textbf{VISOR (TEST)}}\\
\midrule
RTMPose-Hand5 (public)~\cite{jiang2023rtmpose} & 57.30 & 59.19\\
RTMPose-Hand5 + HInt FT~\cite{jiang2023rtmpose} & 60.21 & 65.86\\
FreiHAND-HRNet + HInt FT~\cite{sun2019hrnet} & 40.18 & 49.43\\
S2HAND-Pose (Ours) & \textbf{74.27} & \textbf{78.34}\\
\bottomrule
\end{tabular*}
\end{table}

%% file: tables/tab_stitched_baseline.tex
\begin{table*}[t]
\centering
\scriptsize
\setlength{\tabcolsep}{3pt}
\caption{\textbf{Integrated versus stitched prediction.} AUC@.20 $\uparrow$ (\%). Every stitched row retains Public Sapiens2-5B and attaches the stated public or HInt-adapted RTMPose-Hand5~\cite{jiang2023rtmpose}. All three wrist policies are reported without selecting a different deployment policy using each test set. An external miss falls back to Public; a public-person miss remains a system failure. EPICK is VAL and H2SMV-FPV is synthetic TEST. TailGuard is off.}
\label{tab:stitched-baseline}
\begin{tabular*}{\textwidth}{@{\extracolsep{\fill}}llllll@{}}
\toprule
\shortstack{\textbf{External Hand/Wrist Policy}} & \textbf{NewDays} & \textbf{VISOR} & \textbf{H2O} & \textbf{EPICK} & \textbf{H2SMV-FPV}\\
\midrule
RTMPose-Hand5 (public) / replace all & 63.11 & 65.88 & 67.45 & 66.23 & 47.85\\
RTMPose-Hand5 (public) / keep wrist & 62.82 & 65.73 & 67.04 & 66.10 & 47.86\\
RTMPose-Hand5 (public) / translate fingers & 36.72 & 44.50 & 56.00 & 44.08 & 35.49\\
RTMPose-Hand5 + HInt FT / replace all & 64.96 & 73.45 & 72.59 & 73.14 & 47.06\\
RTMPose-Hand5 + HInt FT / keep wrist & 64.41 & 73.14 & 71.05 & 72.00 & 47.22\\
RTMPose-Hand5 + HInt FT / translate fingers & 34.97 & 56.18 & 40.39 & 40.38 & 29.52\\
\midrule
S2HAND-Pose (Ours) & \textbf{70.04} & \textbf{74.88} & \textbf{73.96} & \textbf{74.35} & \textbf{56.14}\\
\bottomrule
\end{tabular*}
\end{table*}

%% file: tables/tab_adaptation_ablation.tex
\begin{table*}[t]
\centering
\scriptsize
\setlength{\tabcolsep}{3pt}
\caption{\textbf{Adaptation-scope and preservation trade-off.} AUC/PCK are percentages; higher is better. Lower NME is better. Bold/underline denote the best/second-best distinct values per accuracy column, including ties. Mean (3 sets) averages HInt TEST, COCO WholeBody VAL, and FreiHAND EVAL. Face metrics use the same 2,004-person, 16-anchor protocol; NME is normalized by face-box diagonal. A dash means not evaluated. The non-contiguous control has slightly higher hand AUC, but the selected blocks 38--40 give better face NME and PCK; their paired face-NME difference is $-0.00334$ (95\% CI $[-0.00383,-0.00288]$). This supports an accuracy--preservation trade-off, not a claim of maximum hand AUC.}
\label{tab:adaptation-ablation}
\resizebox{\textwidth}{!}{\begin{tabular}{llllllll}
\toprule
\textbf{Adaptation Scope} & \shortstack{\textbf{Trainable Params (M)}} & \shortstack{\textbf{Model Updated (\%)}} & \shortstack{\textbf{HInt VAL}\\\textbf{AUC@.20 $\uparrow$}} & \shortstack{\textbf{Mean (3 sets)}\\\textbf{AUC@.20 $\uparrow$}} & \shortstack{\textbf{Face mean}\\\textbf{NME $\downarrow$}} & \shortstack{\textbf{Face}\\\textbf{PCK@.05 (\%) $\uparrow$}} & \textbf{Preservation}\\
\midrule
Public Sapiens2-5B & 0 & 0 & -- & 57.83 & \textbf{0.1428} & 49.25 & public reference\\
Rank-64 adapter & 0.033 & 0.0006 & 67.49 & 58.07 & -- & -- & empirical\\
Finger-output rows & 0.079 & 0.0015 & 67.49 & 58.08 & -- & -- & channel-local\\
Hand-output rows & 0.079 & 0.0015 & 67.49 & 58.08 & -- & -- & channel-local\\
Decoder tail & 0.210 & 0.0041 & 67.64 & 58.16 & -- & -- & empirical\\
Ordinary pose head & 0.867 & 0.0169 & 68.47 & 58.92 & -- & -- & empirical\\
S2HAND-MT (Ours) & 0.867 & 0.0169 & 68.45 & 58.82 & public route & public route & exact 268 heatmaps\\
Transformer block 38 & 89.620 & 1.7501 & 75.76 & -- & -- & -- & empirical\\
S2HAND-Pose (Ours; blocks 38--40) & 267.130 & 5.2164 & \underline{76.56} & \underline{65.73} & \underline{0.1510} & \textbf{53.70} & empirical audits\\
S2HAND-Pose (seed 43; blocks 38--40) & 267.130 & 5.2164 & 76.50 & 65.55 & -- & -- & independent seed\\
Transformer blocks $\{4,38,40\}$ & 273.052 & 5.3320 & \textbf{76.60} & \textbf{65.86} & 0.1543 & \underline{50.73} & non-contiguous control\\
Transformer blocks 53--55 & 284.890 & 5.5632 & 72.15 & -- & -- & -- & control\\
\bottomrule
\end{tabular}}
\end{table*}

%% file: tables/tab_preservation.tex
\begin{table*}[t]
\centering
\scriptsize
\setlength{\tabcolsep}{3pt}
\caption{\textbf{Whole-body and multi-task preservation.} AP/AR (\%) and face NME use human GT; dense rows report teacher agreement on 512 COCO images. S2HAND-MT preserves all 268 non-finger pose heatmaps and auxiliary task paths by construction (definitions in Appendix~\ref{app:preservation}). TailGuard is disabled.}
\label{tab:preservation}
\begin{tabular*}{\textwidth}{@{\extracolsep{\fill}}llll@{}}
\toprule
\textbf{Endpoint} & \textbf{Public Sapiens2-5B} & \textbf{S2HAND-MT (Ours)} & \textbf{S2HAND-Pose (Ours)}\\
\midrule
Protected pose heatmaps & reference & exact 268-channel route & no exact guarantee\\
COCO body+foot AP / AR $\uparrow$ & 79.09 / 82.83 & public route & 79.28 / 83.07\\
COCO hand AP / AR $\uparrow$ & 51.92 / 57.18 & 54.75 / 60.41 & 64.81 / 70.38\\
Face 16-anchor mean NME $\downarrow$ & 0.1428 & public route & 0.1510\\
Face 16-anchor PCK@.05 $\uparrow$ & 49.25\% & public route & 53.70\%\\
Segmentation foreground agreement $\uparrow$ & reference & unchanged path & 99.5956\%\\
Segmentation hand-mask IoU $\uparrow$ & reference & unchanged path & 99.3246\%\\
Normal mean angular drift $\downarrow$ & reference & unchanged path & $0.6118^\circ$\\
Pointmap mean normalized drift $\downarrow$ & reference & unchanged path & 0.4791\%\\
Pointmap image-mean drift $>5\%$ $\downarrow$ & reference & unchanged path & 5/512 images\\
Matting / albedo matching 5B audit & unavailable & unavailable & unavailable\\
\bottomrule
\end{tabular*}
\end{table*}

%% file: tables/tab_pose2mesh_downstream.tex
\begin{table*}[!t]
\scriptsize
\setlength{\tabcolsep}{1.5pt}
\centering
\caption{\textbf{Frozen Pose2Mesh transfer.} We report PA-MPJPE in mm (lower is better). Only the 2D hand driver changes; all consumer mechanics are fixed. The last rows give baseline-minus-Pose PA-MPJPE differences with paired 95\% cluster-bootstrap intervals; positive favors Pose. GT is a reference input, not a guaranteed lower bound. TailGuard is off.}
\label{tab:pose2mesh}
\begin{tabular*}{\textwidth}{@{\extracolsep{\fill}}llllll@{}}
\toprule
\textbf{2D Driver} & \textbf{FreiHAND} & \textbf{H2O} & \textbf{DexYCB} & \textbf{AssemblyHands} & \textbf{H2SMV-FPV}\\
\midrule
Public Sapiens2-5B~\cite{khirodkar2026sapiens2} & 11.746 & 9.923 & 13.400 & 18.470 & 14.597\\
RTMPose-Hand5 (public)~\cite{jiang2023rtmpose} & 8.768 & 9.705 & \textbf{9.931} & 18.043 & 16.360\\
S2HAND-MT (Ours) & 11.306 & 9.917 & 13.151 & 18.104 & 14.388\\
\textbf{S2HAND-Pose (Ours)} & \textbf{8.424} & \textbf{8.598} & 10.783 & \textbf{16.025} & \textbf{12.521}\\
\midrule
GT 2D (reference) & 4.772 & 5.901 & 5.522 & 17.080 & 7.888\\
\midrule
Public $-$ Pose [95\% CI] & $+3.322\ [3.155,3.488]$ & $+1.325\ [1.056,1.609]$ & $+2.617\ [2.332,2.908]$ & $+2.444\ [2.172,2.747]$ & $+2.077\ [1.946,2.213]$\\
Hand5 $-$ Pose [95\% CI] & $+0.344\ [0.231,0.461]$ & $+1.107\ [0.793,1.452]$ & $-0.851\ [-1.087,-0.617]$ & $+2.017\ [1.718,2.329]$ & $+3.840\ [3.630,4.051]$\\
\bottomrule
\end{tabular*}
\end{table*}

%% file: tables/tab_h2o_action.tex
\begin{table}[!tb]
\scriptsize
\centering
\caption{\textbf{Frozen H2O action-recognition transfer.} 122 clips; GT presence/object features. Both metrics are percentages; Macro denotes macro Top-1. Bold excludes GT/controls. S2HAND-Pose comparison intervals touch zero. Top-5 is omitted because every learned driver reaches 100\%.}
\label{tab:h2o-action}
\begin{tabular*}{\columnwidth}{@{\extracolsep{\fill}}lll@{}}
\toprule
\textbf{Hand XY Source} & \textbf{Top-1 $\uparrow$} & \textbf{Macro Top-1 $\uparrow$}\\
\midrule
EffHandEgoNet-H2O & \textbf{95.90} & \textbf{95.60}\\
Public Sapiens2-5B & 94.26 & 93.75\\
S2HAND-MT (Ours) & 92.62 & 91.90\\
S2HAND-Pose (Ours) & 95.08 & 94.68\\
\midrule
GT 2D (reference) & 96.72 & 96.30\\
Zero hands (control) & 72.95 & 70.37\\
Shuffled hands (control) & 40.98 & 39.35\\
\bottomrule
\end{tabular*}
\end{table}

%% file: tables/tab_model_cost.tex
\begin{table*}[!t]
\centering
\scriptsize
\setlength{\tabcolsep}{3pt}
\caption{\textbf{Model and system cost.} Public, MT, Pose, Hand5, and EffHand rows use L40S, batch-1, native input sizes and pose-forward timing; preprocessing and decoding are excluded. Stitched rows sum existing component medians and are not jointly timed end-to-end: Hand5 estimates omit crop/decode/stitch cost and are lower bounds. Stitched-system parameter totals include the external detector; standalone rows list estimator parameters. Native inputs differ: Sapiens2 $1024\times768$, Hand5 $256\times256$, EffHand $512\times512$.}
\label{tab:model-cost}
\begin{tabular*}{\textwidth}{@{\extracolsep{\fill}}lllll@{}}
\toprule
\textbf{System} & \textbf{Total Params} & \textbf{Adapted Params} & \textbf{Latency (ms) $\downarrow$} & \textbf{Boundary}\\
\midrule
Public Sapiens2-5B~\cite{khirodkar2026sapiens2} & 5.121B & 0 & 320.50 & forward\\
RTMPose-Hand5 (public)~\cite{jiang2023rtmpose} & 13.775M & 0 & 4.61 & hand forward\\
EffHandEgoNet-H2O~\cite{mucha2024effhand} & 25.242M & 0 & 10.10 & native forward\\
Public + RTMPose-Hand5 (public) & $\sim$5.161B & 0 & $\geq331.05$ & component estimate\\
Public + RTMPose-Hand5 + HInt FT & $\sim$5.161B & 13.775M & $\geq331.05$ & component estimate\\
Public + WiLoR~\cite{potamias2025wilorendtoend3dhand} & 5.787B & 0 & 356.07 & component estimate\\
S2HAND-MT (Ours) & 5.121B & 0.867M & 322.27 & forward\\
S2HAND-Pose (Ours) & 5.121B & 267.130M & 323.36 & forward\\
\bottomrule
\end{tabular*}
\end{table*}

%% file: sec/6_conclusion.tex
\section{Conclusion}

We presented S2HAND, a hand-specialized adaptation of Sapiens2-5B that retains its unified whole-body interface. For existing Sapiens2 deployments requiring strictly unchanged non-hand outputs, S2HAND-MT routes 40 adapted finger heatmaps while preserving the remaining 268 pose heatmaps and auxiliary task paths by construction. S2HAND-Pose additionally tunes three Fisher-selected blocks for larger hand gains with empirical preservation, surpassing HInt-adapted Hand5/HRNet specialists and stitched Hand5 systems in Native AUC across all five sets (while EffHandEgoNet remains stronger on H2O). Our multi-task audit quantifies segmentation, normal, and pointmap drift alongside mixed face preservation, and optional TailGuard reduces severe S2HAND-MT failures. Moreover, the adapted 2D hand predictions reduce reconstruction error for frozen Pose2Mesh, whereas action and sign recognition gains remain inconclusive. Overall, our results support specializing hands directly within a unified whole-body model without claiming universal superiority over standalone estimators. We will release our code, both checkpoints, and the H2SMV evaluation benchmark.

%% file: sec/7_appendix.tex
\section{Architecture and Release Details}
\label{app:architecture}
This appendix specifies the trainable parameter groups, checkpoint
initialization, mixed-precision settings, and routed output interfaces for
S2HAND-MT and S2HAND-Pose. S2HAND-MT freezes the public 308-channel
branch and routes only the 40 finger channels from its adapted branch, leaving
both wrist channels on the public route. S2HAND-Pose uses the same
routed hand branch while additionally updating Transformer blocks 38--40.

Crucially, the adapted S2HAND-MT branch is not a narrow 40-channel output head;
it duplicates the full three-layer $1\times1$ pose decoder and its 308-channel
projection. A deterministic router then selects only the 40 non-wrist finger
heatmaps from this adapted branch while drawing the remaining 268 heatmaps
(including both wrists) from the frozen public branch. Thus, ``40-channel
adaptation'' refers to the routed output interface rather than the internal
width of the copied decoder.

\subsection{Training-Loss Implementation}
\label{app:loss-details}
Let \(\mathcal J=\mathcal H\cup\{w_L,w_R\}\) denote the 40 adapted finger
channels and two detached public-route wrists. Let \(Z_j^a\) be an
adapted-branch heatmap and \(Z_j^{\rm pr}\) be the corresponding
heatmap from the frozen public decoder route. The routed training heatmap is
\[
\widetilde Z_j=
\begin{cases}
Z_j^a, & j\in\mathcal H,\\
\operatorname{stopgrad}(Z_j^{\rm pr}), & j\in\{w_L,w_R\}.
\end{cases}
\]
Given target heatmap \(G_j\) and visibility/validity weight \(w_j\), the
heatmap loss is
\begin{equation}
\mathcal L_{\rm hm}=\operatorname{mean}_{j\in\mathcal J,u,v}
\left[w_j\bigl(\widetilde Z_j(u,v)-G_j(u,v)\bigr)^2\right].
\end{equation}
Here, the tensor mean is computed over all selected channels and spatial locations
without renormalizing by the sum of valid weights. Because the public wrist
heatmaps are detached, their heatmap errors act as constants with respect to
the adapted branch.

During training, a differentiable coordinate decoder applies a spatial softmax with inverse
temperature \(\beta=64\):
\begin{equation}
\begin{aligned}
p_j(u,v)&=\frac{\exp(\beta\widetilde Z_j(u,v))}
{\sum_{u',v'}\exp(\beta\widetilde Z_j(u',v'))},\\
\mathbf r(u,v)&=\left[\frac{768u}{W-1},
\frac{1024v}{H-1}\right]^\top,\\
\hat{\mathbf q}_j&=\sum_{u,v}p_j(u,v)\mathbf r(u,v),
\end{aligned}
\end{equation}
where \(W\times H\) is the heatmap grid size, and \(\mathbf r(u,v)\) maps
grid coordinates to the \(768\times1024\) network-input space. Let
\(\mathbf q_j\in\mathbb R^2\) be the GT input-space coordinate of joint
\(j\), and let \(\mathcal V\) be the set of joints with finite \(\mathbf q_j\) and
positive visibility. The normalization scale
\(s\) is the maximum horizontal or vertical span of valid GT hand joints,
clamped to at least 16 input pixels; hands with fewer than two valid joints
are excluded from the coordinate and topology terms. We compute
\begin{align}
\mathcal L_{\rm coord}
&=\frac{1}{|\mathcal V|}\sum_{j\in\mathcal V}
\operatorname{SmoothL1}_{0.05}
\!\left(\frac{\hat{\mathbf q}_j-\mathbf q_j}{s}\right),\\
\mathcal L_{\rm topo}
&=\frac{1}{|\mathcal E_v|}\sum_{(k,j)\in\mathcal E_v}
\notag\\[-2pt]
&\quad
\operatorname{SmoothL1}_{0.025}
\!\left(\frac{\hat{\mathbf q}_j-\hat{\mathbf q}_k
-(\mathbf q_j-\mathbf q_k)}{s}\right),
\end{align}
where \(\mathcal E_v\) is the union of valid directed hand edges (including the
five wrist-to-MCP edges per hand). The Smooth-L1 subscripts denote the PyTorch
transition thresholds \(0.05\) and \(0.025\) applied after scale normalization by
\(s\), averaging across the two coordinate axes. The loss weights are fixed at
\(\lambda_{\rm coord}=0.0030615822269171302\) and
\(\lambda_{\rm topo}=0.0016362079765548576\), calibrated so that each
auxiliary term contributes \(0.1\times\) the heatmap-loss gradient norm rather than
tuned via a validation sweep. This soft coordinate decoder is used strictly during
training, whereas inference relies on the standard public UDP and DARK decoding path.

\section{Datasets and Evaluation Protocols}
\label{app:protocols}
We report three evaluation protocols separately because they address distinct questions:
Native Pipeline captures each system's end-to-end detector and cropping behavior;
Shared-GT Crop isolates keypoint regression using a $2\times$ GT-hand ROI; and
Shared-Detector Crop shares identical WiLoR-YOLO $2\times$ detected ROIs while
counting missed detections as failures. All coordinates are evaluated in native
image pixels. NME normalizes each joint error by the maximum side length of the tight GT
hand-keypoint box, and both PCK and AUC follow this same normalization.

H2SMV is our synthetic dataset, where H2SMV-FPV denotes its held-out
first-person evaluation split with paired 2D pose
and 3D mesh annotations. Excluded from all training, validation, Fisher
probing, checkpoint selection, and TailGuard tuning, it serves as a
model-agnostic cross-domain audit independent of any single real
dataset's annotation convention.

\subsection{Published HInt Scores and PCK Aggregation}
\label{app:published-comparison}
Table~\ref{tab:published-hint-reference} separates published reference numbers from our controlled reruns. HandOS jointly performs detection and 3D reconstruction, RTMPose-Hand5 is an efficient 2D hand estimator, and EffHandEgoNet targets egocentric H2O/FPHA benchmarks~\cite{chen2025handos,jiang2023rtmpose,mucha2024effhand}. A later publication date does not imply higher HInt 2D accuracy, and no official HandOS code or weights were available on its project page for controlled evaluation.\footnote{\url{https://idea-research.github.io/HandOSweb/}}

ExtPose predicts 3D hand geometry from images and auxiliary 2D cues, evaluating HInt PCK on projected 3D joints alongside a ViTPose baseline that directly predicts 2D joints~\cite{chen2025extpose,xu2022vitpose}. Because several of these published scores exceed ours, we retain them rather than dismissing them by task category; however, without verified matching detector, crop, handedness, and averaging rules, we do not claim a global HInt ranking across these rows.

Specifically, our canonical evaluator pools all valid joints, whereas the released HaMeR PCK code computes per-joint accuracy across valid samples before averaging the 21 joint scores.\footnote{\url{https://github.com/geopavlakos/hamer/blob/main/hamer/utils/pose_utils.py}} Because valid sample counts vary by joint, the two aggregations differ: applying HaMeR-style averaging to our cached Native predictions shifts NewDays PCK@.05 from 59.83\% to 59.48\% and VISOR from 66.28\% to 66.17\%. This sensitivity check—rather than a reproduction of ExtPose's input protocol—shows why a 0.08-point gap against HandOS cannot establish a reliable ranking. All canonical rerun metrics, cached predictions, and AUC intervals remain unchanged.

\subsection{Training and Inference Details}
\label{app:training-details}
Final S2HAND models train exclusively on HInt NewDays TRAIN and EPICK TRAIN, sampled at a \(1{:}1\) ratio. Checkpoints are selected by the unweighted mean AUC across NewDays VAL and EPICK VAL under our fixed severe-error guard, while TailGuard is tuned separately on EPICK VAL alone; no TEST split enters either decision. During training, crop context scale is sampled from \([1.875,3.125]\), center jitter is bounded at \(0.08\) of crop size, rotation is sampled from \([-30^\circ,30^\circ]\), and random hand occlusion is applied with probability \(0.20\).

Both S2HAND variants branch from the same source-balanced HInt-adapted pose checkpoint (selected at 12k sample steps). From this shared initialization, S2HAND-MT trains its isolated branch for 18k sample steps, while S2HAND-Pose is trained as an independent 18k-sample-step run rather than continuing from the final MT checkpoint. All stages draw from the same two HInt TRAIN sources. S2HAND-MT employs AdamW with a \(2\times10^{-5}\) decoder learning rate, zero weight decay, BF16 autocast, gradient clipping at \(1.0\), and validation every 3k sample steps.

Training and inference operate on the standard $1024\times768$ network input and $256\times192$ UDP heatmaps. Our formal 5B runs use a per-device batch size of one (so one optimizer step equals one sample step), four to eight dataloader workers depending on the stage, pinned host memory, and no gradient accumulation. Samples are shuffled with a fixed seed, and batch size one precludes within-batch temporal grouping. Our code release will include environment-resolved configs, package locks, random seeds, split manifests, and launch scripts.

In its 18k-sample-step run, S2HAND-Pose sets the hand-branch learning rate to \(2\times10^{-5}\) and updates Transformer blocks 38--40 at \(2\times10^{-6}\) using BF16 and activation checkpointing (seed 42 for the main run, seed 43 for the stability run). At inference, all Sapiens2 variants share identical image normalization, joint ordering, handedness conversion, UDP/DARK decoding, sample IDs, and protocol-specific crop transforms.

Under Shared-Detector Crop, H2O TEST and H2SMV-FPV TEST are evaluated on prospectively frozen 2,048-image subsets selected by hash prior to model comparison, whereas NewDays TEST
and EPICK VAL use full splits. All models within a dataset are evaluated on identical hand IDs, counting missing detector boxes as failures and reporting detector coverage alongside accuracy.

\subsection{Matched-Data Specialist Adaptation}
\label{app:matched-training}
We replace the earlier collapsed Hand5 control with a reproducible native-geometry adaptation. Both public RTMPose-Hand5 and our reproduced FreiHAND-HRNet-W18 are fine-tuned on the exact same manifest of 12,000 balanced HInt TRAIN hands (6,000 NewDays and 6,000 EPICK, shuffled with seed 42) for 12,000 sample steps at batch size 32, matching the 12k-sample HInt adaptation scale used to initialize S2HAND.

Both specialists take $256\times256$ inputs with MMPose's native $1.25\times$ box padding, omitting Sapiens-style $2\times$ context expansion during Native inference. All parameters remain trainable. Hand5 uses AdamW (weight decay 0.01) and HRNet uses Adam; both warm up over 800 sample steps from $10^{-5}$ to a peak LR of $10^{-4}$, followed by cosine decay to $10^{-6}$, with gradient clipping at 1.0 and dynamic FP16 loss scaling. Geometric augmentations are listed in Table~\ref{tab:matched-training}: both apply horizontal flipping and photometric distortion, while HRNet adds a 0.15 center-shift factor. Flip testing is enabled for both, and HRNet additionally uses shifted heatmaps. Checkpoints are evaluated on HInt VAL every 3,000 sample steps (at 3k, 6k, 9k, and 12k steps); as validation AUC plateaus by 12k steps (e.g., shifting by only $+0.09$ points from 9k to 12k on Hand5), we select the converged 12k peak checkpoint for both models.

Because the earlier collapsed Hand5 run is not a credible negative baseline, it is omitted from the main tables. Restoring the native crop convention and verifying checkpoint identity, sample IDs, and coordinate transforms improves AUC on both real HInt test splits without reproducing the prior collapse. As this fix was not designed as a factorial isolation of the historical bug, we neither attribute the earlier collapse to a single cause nor frame it as ordinary domain shift. Native and Shared-GT results are reported separately, TEST performance never informs checkpoint selection, and the main-paper Shared-GT table evaluates all models on the exact same 3,660 hand IDs and $2\times$ GT crops.

\subsection{Stitched Systems and Wrist Policies}
\label{app:stitched-details}
In stitched baselines, the public whole-body model supplies the body-connected wrist while Hand5 predicts a wrist and 20 finger joints. Under \emph{replace-all}, all 21 external joints overwrite the public hand, including the body wrist; \emph{keep-wrist} retains the public wrist and copies the 20 external fingers unchanged; and \emph{translate} retains the public wrist while shifting all external fingers by the wrist-offset vector (preserving internal finger geometry at the cost of image alignment). All policies operate in native image coordinates under a unified anatomical ordering. Specialist misses fall back to Public Sapiens2, whereas public person-detection misses remain system failures. Rather than cherry-picking policies per test set, the main-paper stitching table reports all three policies across every benchmark.

Wrist disagreement is defined on mutually detected hands as the Euclidean distance between the public and external wrists, normalized by the maximum side length of the tight GT-hand box; it serves as a geometric diagnostic rather than a standalone accuracy metric. Coverage is computed over all canonical hands, counting misses as failures. On NewDays, WiLoR achieves $1,599/1,754=91.16\%$ coverage and reaches 61.31 (replace-all), 60.94 (keep-wrist), and 33.43 (translate) AUC, compared with 70.04 for S2HAND-Pose. We include this single-dataset WiLoR test as a targeted diagnostic rather than a full multi-dataset benchmark.

\subsection{System-Cost Boundaries}
\label{app:system-cost}

Forward-only latency is measured on a single L40S GPU (batch size 1) using each model's native input resolution and BF16 autocast where supported, excluding crop preprocessing and coordinate decoding. External-stage timing is evaluated over 100 fixed unique NewDays images with three repeats after warm-up, starting from decoded RGB frames. Combining WiLoR-YOLO (26.631M parameters) and Hand5 (13.775M) adds 40.406M external parameters. By comparison, the full WiLoR detector-plus-regressor pipeline totals 665.912M parameters with a median runtime of 35.57 ms, scaling with the number of detected hands.

The serial Hand5 component estimate sums Public Sapiens2 forward, hand detection, and single-hand inference ($320.50+5.94+4.61=331.05$ ms). This represents a lower-bound accounting convention: two detected hands require a second hand pass, and cropping, decoding, and stitching overhead are omitted. Similarly, the WiLoR estimate ($320.50+35.57=356.07$ ms) adds Public Sapiens2 forward latency to the full external-stage median under a separate timing boundary. Neither figure represents a jointly timed end-to-end deployment latency. Public and HInt-adapted Hand5 share identical architectures and runtime accounting, whereas TailGuard is audited separately on an A100 GPU and is not mixed into these L40S measurements.

\subsection{H2SMV Dataset and Release}
\label{app:h2smv}
H2SMV comprises 3,000 rendered motion sequences and 318,082 first-person RGB
frames at $768\times1024$ resolution. While source motions and reference imagery originate
from How2Sign~\cite{duarte2021how2sign}, 3D body/hand reconstruction, virtual camera placement,
mesh-derived keypoints, and visibility labels are generated by our
standalone annotation pipeline. Primary egocentric cameras are
mounted at the eyes, head, and left/right upper arms, supplemented by a sparse
chest-mounted view when passing visibility checks. Each frame
provides COCO WholeBody-133 2D image coordinates, camera-space 3D joints (in
meters), camera intrinsics and extrinsics, paired 3D meshes, and tri-state visibility labels
($0$: out of frame, $1$: occluded, $2$: visible).

Splits are partitioned by motion sequence rather than individual frames, keeping all camera views and
temporally adjacent frames from a given sequence within the same split. Final
S2HAND models never access H2SMV for training, validation, checkpoint
selection, Fisher probing, or TailGuard tuning; only its held-out test split is
evaluated as an independent synthetic audit. We will release the S2HAND
training and inference code, model weights, H2SMV split manifests, camera
conventions, COCO-133 mappings, integrity metadata, evaluation scripts, and all
redistributable derived annotations. Source RGB redistribution follows
the How2Sign license without overriding third-party terms, and
repository URLs are withheld during anonymous review.

Supporting material is collected after the explanatory sections: Tables~\ref{tab:dataset-protocols}, \ref{tab:native-full}, \ref{tab:p1-shared-roi}, \ref{tab:p2b-shared}, \ref{tab:matched-training}, \ref{tab:stitch-diagnostics}, \ref{tab:stitched-cost}.

\section{Extended 2D Hand Results}
\label{app:hand-results}
Table~\ref{tab:s2hand-multiscale-native} separates adaptation gains from model
capacity by pairing each S2HAND-Pose scale with its same-size public counterpart.
Adaptation improves Native Pipeline AUC across all reported datasets at all
three smaller scales. Notably, the 1B variant reaches a five-set mean within 0.59
AUC points of the 5B specialist, even though the public baselines do not scale monotonically
with model size on every cross-domain benchmark.

The multi-scale qualitative plates display eight deterministic matched hands per
dataset, fixing the source image, annotation, and visualization crop across each row.
Columns pair the 0.4B, 0.8B, 1B, and 5B public models with their adapted
counterparts (including S2HAND-MT at 5B); these plates
supplement Table~\ref{tab:s2hand-multiscale-native} and never inform model selection.
Figures~\ref{fig:app-main-assemblyhands}--\ref{fig:app-main-hint} then present
the main-model comparisons across eight deterministic samples per page under
fixed columns: GT, Public Sapiens2-5B, RTMPose-Hand5 (public),
EffHandEgoNet-H2O, FreiHAND-HRNet reproduction, S2HAND-MT, and S2HAND-Pose,
rendering all coordinates within the fixed per-row crop.

Supporting material is collected after the explanatory sections: Tables~\ref{tab:s2hand-multiscale-native}; Figures~\ref{fig:app-scale-newdays}, \ref{fig:app-scale-visor}, \ref{fig:app-scale-h2o}, \ref{fig:app-scale-epick}, \ref{fig:app-scale-h2smv}, \ref{fig:app-main-assemblyhands}, \ref{fig:app-main-coco}, \ref{fig:app-main-dexycb}, \ref{fig:app-main-freihand}, \ref{fig:app-main-h2smv}, \ref{fig:app-main-hint}.

\section{Adaptation and Layer-Selection Controls}
\label{app:adaptation}
Our diagonal-Fisher probe evaluates all blocks on the same eight cached training batches
without taking optimizer steps. Ablation controls compare channel-local adaptation,
the pose decoder tail, the full pose head, block 38 alone, our contiguous
triplet 38--40, a non-contiguous \{4,38,40\} control, and terminal
blocks 53--55. Seed replication and loss-gradient calibration are evaluated on
the same HInt validation split.

Across scales, the selected contiguous windows are blocks 16--18 (0.4B), 17--19 (0.8B),
28--30 (1B), and 38--40 (5B), each fixed from its training-only Fisher
profile prior to adaptation without validation or test feedback.
We do not claim that these windows are globally optimal across all scales:
for instance, the 0.8B window spans individually ranked
blocks 2, 4, and 5 rather than the top three isolated blocks. Split-half
probe Spearman correlations reach 0.963, 0.804, 0.963, and 0.951 across the four scales,
and the 5B release is unique in that its top three individual blocks (38, 40, and
39) directly form the contiguous window 38--40.

Compared with blocks 38--40, the non-contiguous \(\{4,38,40\}\) control shifts HInt TEST, COCO WholeBody
VAL, and FreiHAND AUC by only \(+0.20\), \(+0.14\), and \(+0.05\) points, with mixed effects on other hand metrics. Evaluating both windows on the full 2,004-person face protocol from
Table~\ref{tab:preservation} (rather than a smaller subset),
S2HAND-Pose and \(\{4,38,40\}\) achieve mean NME of 0.1510 vs.\ 0.1543,
PCK@.05 of 53.70\% vs.\ 50.73\%, PCK@.10 of 79.69\% vs.\ 78.23\%, and point-wise
P95 of 0.1547 vs.\ 0.1580. The paired per-face NME difference
(S2HAND-Pose minus \(\{4,38,40\}\)) is \(-0.00334\) (95\% bootstrap CI \([-0.00383,-0.00288]\)), confirming that adding an early
block degrades full-protocol face preservation without justifying the broader
adaptation scope. Finally, for training stability, the seed-43 Pose run reaches 76.50 HInt VAL AUC and a
65.55 three-set mean, closely matching the main seed-42 run (76.56 and 65.73).

\subsection{MT Seed Repeatability and Paired Uncertainty}
\label{app:repeatability}

Evaluating S2HAND-MT across seeds 42, 43, and 44 under identical routing and checkpoint selection yields a sample SD of 0.063 AUC points on HInt VAL and 0.055 on HInt TEST (controlled-crop metrics that complement rather than replace the Native table). While the 0.02-point single-run gap between S2HAND-MT and ordinary head tuning falls within this seed variance, ordinary head tuning was not repeated across all three seeds, and no formal equivalence test is claimed.

Our Native paired bootstrap runs 10,000 replicates (seed 20261017) on matched sample IDs across 850 NewDays videos, 41 VISOR videos, 46 H2O sequences, 91 EPICK videos, and 20 H2SMV video groups. Each replicate resamples clusters with replacement, retains all constituent hands, and recomputes valid-joint-weighted AUC and paired differences over the exact 201-point threshold grid of our canonical evaluator. The resulting 2.5th and 97.5th percentiles quantify evaluation-sample uncertainty conditional on fixed checkpoints rather than multi-seed training variance, without multiple-comparison adjustments. Because external published scores lack paired predictions and verified aggregation rules, we do not conduct significance tests against them.

Supporting material is collected after the explanatory sections: Tables~\ref{tab:fisher-probe}, \ref{tab:fisher-full-0.4b}, \ref{tab:mt-repeatability}; Figures~\ref{fig:fisher-model-family}, \ref{fig:accuracy-budget}.

\section{Whole-Body and Multi-task Preservation}
\label{app:preservation}
While S2HAND-MT provides an exact routing guarantee for all 268 protected pose heatmaps,
S2HAND-Pose updates shared Transformer blocks and is therefore audited empirically
across body, face, and auxiliary tasks. Because public human GT annotations are
unavailable for the Sapiens2 238-point dense face layout, dense-face metrics reflect
teacher-relative agreement; similarly, segmentation metrics quantify agreement with
Public Sapiens2-5B rather than GT accuracy.

Body-and-foot accuracy evaluates 17 body and 6 foot keypoints via COCO OKS (0.50--0.95), while combined hand AP/AR averages separately evaluated left and right hands. Table~\ref{tab:coco-hand-complete} reports the complete left, right, and combined hand scores across all 5B pose variants, whereas non-hand outputs for S2HAND-MT remain identical to Public Sapiens2-5B by construction.

Our expanded dense audit uses 512 unique COCO WholeBody VAL full images, selected deterministically from 698 eligible images via round-robin stratification over scale, visibility, and crowding. In each eligible image, the primary non-crowd person has at least 14 labeled body joints, a bounding-box area fraction in $[0.04,0.8]$, a height of at least 35\% of the image, and a 1\% in-frame margin. Selection is strictly independent of model scores. All task heads use the exact same image manifest and ordering, with nonempty public foreground masks, and are evaluated at $256\times192$ via the public full-image pipeline.

For each dense task, we inject the learned S2HAND-Pose block 38--40 weight deltas into the released task checkpoint while keeping its task head and remaining weights frozen (transferring 51 tensors with an $\ell_2$ update norm of 5.30657 to isolate the shared-backbone shift). All-pixel segmentation agreement includes background, foreground agreement uses the union of public and adapted non-background masks, and hand-mask IoU compares the union of left and right hand classes, weighting all images equally.

Surface normals are scaled to unit length, and angular drift is measured on the fixed public-segmentation foreground mask. Mean angular drift and the fraction of pixels within $5^\circ$ are macro-averaged across images, while image-P95 averages each image's 95th-percentile foreground drift.

Following the released pipeline, pointmap predictions are divided by their estimated scale. At each public-foreground pixel, we compute the 3D endpoint distance between public and adapted pointmaps, normalized by that image's 95th-percentile public foreground point magnitude (floored at $10^{-6}$). Macro drift averages per-image normalized mean distances, whereas the pixel-weighted mean weights each image by its foreground pixel count. We retain all 512 images—including 40 exceeding 1\% mean drift, five exceeding 5\%, and a worst-case mean of 11.8552\%—noting that these thresholds measure teacher-relative drift rather than GT reconstruction error, and a low global mean does not imply uniform per-image preservation.

Because the official Sapiens2 model zoo provides a 1B matting checkpoint but neither 5B matting nor 5B albedo weights, release availability—rather than missing GT—precludes a matching 5B audit; both tasks are marked unavailable rather than preserved or failed.\footnote{\url{https://github.com/facebookresearch/sapiens2/blob/main/docs/MODEL_ZOO.md}} This dense audit is conducted at 5B, whereas multi-scale pose and face audits are reported separately below.

Face evaluation covers 2,004 valid COCO WholeBody VAL instances across the 16 semantic anchors shared by COCO-68 and the Sapiens2 238-point layout. Joint errors are normalized by the annotated face-box diagonal, averaged first across valid anchors per face and then across faces. For S2HAND-Pose, mean anchor NME worsens slightly while threshold accuracy and tail rates improve; dense 238-point comparisons remain teacher-relative rather than human-GT accuracy.

Finally, Table~\ref{tab:multiscale-preservation} reveals scale-dependent preservation trade-offs: the 0.4B and 0.8B adaptations trade 1.62 and 0.98 body+foot AP points for their hand gains, whereas the 1B and 5B shifts stay within 0.19 points. Conversely, the 16-anchor face NME improves at 0.4B and 0.8B but slightly degrades at 1B and 5B (reflecting sparse anchor accuracy while the final column reports teacher-relative dense drift).

Supporting material is collected after the explanatory sections: Tables~\ref{tab:coco-hand-complete}, \ref{tab:dense-audit}, \ref{tab:multiscale-preservation}; Figures~\ref{fig:app-body-neutral}, \ref{fig:app-body-tail}, \ref{fig:app-face-mixed}, \ref{fig:app-face-tail}, \ref{fig:app-scale-body}, \ref{fig:app-scale-face}, \ref{fig:body-resolution-app}, \ref{fig:face-resolution-app}.

\section{TailGuard Details}
\label{app:tailguard}

TailGuard is a post-hoc inference router that re-applies the same S2HAND model to a
WiLoR-YOLO $2\times$ same-side hand crop whenever native hand confidence falls below a frozen
threshold, retaining the native prediction if no same-side box is detected. While our primary accuracy ablation applies TailGuard to S2HAND-MT, we also transfer the exact same threshold and crop policy to S2HAND-Pose without retuning. Threshold and context-scale sweeps, trigger rates, harmful re-crops, detector overhead, and conditional second-pass latency are reported separately.

The deployed threshold represents a conservative accuracy--compute operating point
rather than the peak-AUC setting from our validation sweep. On EPICK VAL, raising the
threshold from 0.510756 to 0.65 lifts AUC from 65.02 to 67.35 while increasing
the trigger rate from 38.56\% to 54.08\%; unconditionally re-cropping reaches 68.71 AUC
at a 95.04\% trigger rate. At $\tau=0.510756$, a $2\times$ crop context
outperforms $1.5\times$ and trails $2.5\times$ by only 0.09 AUC points, so we freeze the
smaller of the two near-tied contexts for TEST evaluation.

Transferred directly to S2HAND-Pose, this frozen router improves AUC and all three PCK thresholds across
NewDays, H2O, and H2SMV-FPV
(Table~\ref{tab:tailguard-pose-transfer}), with the largest gains on NewDays and a
0.39\% trigger rate on H2O. This confirms conservative cross-variant transferability,
though individual triggered re-crops are not guaranteed to improve every sample.

Our repeated A100 timing audit spans 1,000 inputs, of which 460 trigger the second
S2HAND-MT pass. Timing begins from a decoded BGR frame and cached person box
and ends at decoded keypoints, excluding disk I/O and person detection.
Because fewer than half of the samples trigger re-cropping, the conditional stage has a median latency of zero.
WiLoR-YOLO achieves median and P95 latencies of 10.15 and 10.62 ms, closely matching
the main audit (10.20 and 10.81 ms). However, a single 5.48-s system stall
inflates the repeat mean to 16.21 ms (10.74 ms after excluding this single outlier).
We therefore report the mean, median, P95, P99, and maximum in
Table~\ref{tab:tailguard-cost} rather than treating the arithmetic mean as
steady-state detector throughput, and retain all raw per-sample stage timings.

Supporting material is collected after the explanatory sections: Tables~\ref{tab:tailguard-ablation}, \ref{tab:tailguard-val-sweep}, \ref{tab:tailguard-pose-transfer}, \ref{tab:tailguard-cost}; Figures~\ref{fig:tailguard-audit-app}.

\section{Downstream Protocols and Limitations}
\label{app:downstream}
Using one frozen FreiHAND-trained checkpoint across all drivers and sets, Pose2Mesh receives substituted 2D hand coordinates after standardized joint-order, handedness, crop, and normalization transforms. Common-valid sets contain 3,960 FreiHAND, 4,096 H2O, 997 DexYCB, 7,982 AssemblyHands, and 9,379 H2SMV-FPV hands (full splits rather than 500-hand diagnostic subsets). Native detection and 3D transfer use fixed separate manifests.

Our 10,000-replicate paired bootstrap resamples instances on FreiHAND (3,960) and sequence/capture groups on H2O (46), DexYCB (197), AssemblyHands (293), and H2SMV-FPV (306), recomputing sample-weighted PA-MPJPE and baseline-minus-Pose differences (positive favors Pose). Gains over Public Sapiens2 exclude zero on all five sets, whereas public Hand5 leads on DexYCB by 0.851 mm (CI excluding zero). This test uses the public Hand5 driver without substituting our HInt-adapted 2D Hand5 control.

Frozen H2O ActionTransformer (122 clips, 20 frames/clip, fixed GT hand-presence and object features) and frozen Uni-Sign are evaluated similarly by varying only 2D hand trajectories. Because these tests measure zero-shot plug-in compatibility rather than consumer retraining—and their paired recognition CIs span zero—recognition transfer remains a limitation.

The Uni-Sign overlap spans 84 videos across 68 WLASL-2000 classes, replacing only the 42 hand keypoints while fixing native RTMPose confidences and non-hand joints. S2HAND-Pose matches native RTMPose at 71.43\% instance Top-1 and reaches 76.72\% macro Top-1, though paired CIs include zero and this subset is not the full WLASL-2000 benchmark.

Supporting material is collected after the explanatory sections: Tables~\ref{tab:pose2mesh-ci}, \ref{tab:unisign}.

%% file: sec/8_appendix_tables.tex
\input{tables/tab_dataset_protocols_appendix}
\input{tables/tab_native_full_appendix}
\input{tables/tab_published_hint_reference_appendix}
\input{tables/tab_p1_shared_oracle_roi_appendix}
\input{tables/tab_p2b_shared_detector_appendix}
\input{tables/tab_matched_training_appendix}
\input{tables/tab_stitch_diagnostics_appendix}
\input{tables/tab_stitched_cost_appendix}
\input{tables/tab_s2hand_multiscale_native_appendix}
\input{tables/tab_fisher_probe_appendix}
\input{tables/tab_fisher_full_ranks_appendix}
\input{tables/tab_mt_repeatability_appendix}
\input{tables/tab_coco_hand_complete_appendix}
\input{tables/tab_dense_audit_appendix}
\input{tables/tab_multiscale_preservation_appendix}
\input{tables/tab_tailguard_ablation}
\input{tables/tab_tailguard_val_sweep_appendix}
\input{tables/tab_tailguard_pose_transfer_appendix}
\input{tables/tab_tailguard_cost_appendix}
\input{tables/tab_pose2mesh_ci_appendix}
\input{tables/tab_unisign_wlasl_appendix}
\clearpage

%% file: tables/tab_dataset_protocols_appendix.tex
\begin{table*}[t]
\small
\centering
\caption{\textbf{Dataset roles and protocols.} We list annotations, frozen roles, and resampling units. EPICK is a development split and H2SMV-FPV is synthetic; neither is used as a held-out headline endpoint.}
\label{tab:dataset-protocols}
\resizebox{\textwidth}{!}{\begin{tabular}{llll}
\toprule
\textbf{Dataset} & \textbf{Domain/Annotation} & \textbf{Role} & \textbf{Resampling Unit}\\
\midrule
HInt-NewDays TRAIN~\cite{pavlakos2024hamer} & difficult visibility-aware real hands & training; balanced 1:1 with EPICK TRAIN & video/sequence\\
HInt-EPICK TRAIN~\cite{pavlakos2024hamer} & difficult visibility-aware real hands & training; balanced 1:1 with NewDays TRAIN & video/sequence\\
HInt-NewDays VAL~\cite{pavlakos2024hamer} & difficult visibility-aware real hands & checkpoint selection; macro-averaged with EPICK VAL & video/sequence\\
HInt-EPICK VAL~\cite{pavlakos2024hamer} & difficult visibility-aware real hands & checkpoint selection and TailGuard selection & video/sequence\\
HInt-NewDays TEST~\cite{pavlakos2024hamer} & difficult visibility-aware real hands & final HInt test endpoint & video/sequence\\
HInt-VISOR TEST~\cite{pavlakos2024hamer} & difficult visibility-aware real hands & final HInt test endpoint & video/sequence\\
COCO WholeBody VAL~\cite{jin2020wholebody} & third-person 133-point GT & external pose/preservation & image/person\\
H2O TEST~\cite{kwon2021h2o} & FPV hand-object 2D/3D & external 2D and 3D & sequence\\
H2O subject-3 action~\cite{kwon2021h2o} & temporal action/object/hand labels & frozen recognition & clip\\
H2SMV-FPV TEST & synthetic paired 2D/3D/mesh & held-out audit only; never model/tuning input & synthetic sequence\\
FreiHAND EVAL~\cite{zimmermann2019freihand} & real hand 2D/3D/mesh & external 2D and 3D & capture/image\\
DexYCB subjects 01/02~\cite{chao2021dexycb} & hand-object 2D/3D/masks & occlusion and 3D & sequence\\
AssemblyHands VAL~\cite{ohkawa2023assemblyhands} & FPV/multiview 2D/3D & external 2D and 3D & sequence/capture\\
WLASL-2000 overlap~\cite{li2020wlasl} & isolated sign video & frozen pose-only recognition & video\\
\bottomrule
\end{tabular}}
\end{table*}

%% file: tables/tab_native_full_appendix.tex
\begin{table*}[t]
\scriptsize
\renewcommand{\arraystretch}{0.90}
\setlength{\tabcolsep}{4pt}
\centering
\caption{\textbf{Complete Native Pipeline evaluation.} AUC, PCK, and severe tails are percentages. Missed hands count as failures. Sapiens2 uses person crops, Hand5/HRNet use WiLoR-YOLO hand boxes, and EffHand uses full images. Published literature scores are separated in Table~\ref{tab:published-hint-reference} and are not included in this controlled rerun table. HInt-FT rows use the corrected training protocol. EPICK is VAL; H2SMV-FPV is synthetic TEST. TailGuard is off.}
\label{tab:native-full}
\resizebox{\textwidth}{!}{\begin{tabular}{llllllll}
\toprule
\textbf{Dataset} & \textbf{System} & \textbf{Hands} & \textbf{AUC@.20 $\uparrow$} & \textbf{PCK@.05 $\uparrow$} & \textbf{PCK@.10 $\uparrow$} & \textbf{PCK@.15 $\uparrow$} & \textbf{Tail$>$.20 $\downarrow$}\\
\midrule
NewDays TEST~\cite{pavlakos2024hamer} & Public Sapiens2-5B~\cite{khirodkar2026sapiens2} & 1754 & 61.48 & 47.72 & 70.69 & 81.40 & 14.14\\
 & RTMPose-Hand5 (public)~\cite{jiang2023rtmpose} & 1754 & 59.72 & 48.63 & 69.04 & 77.46 & 16.59\\
 & RTMPose-Hand5 + HInt FT~\cite{jiang2023rtmpose} & 1754 & 61.57 & 52.82 & 70.49 & 77.62 & 17.33\\
 & FreiHAND-HRNet reproduction~\cite{sun2019hrnet,potamias2025wilorendtoend3dhand} & 1754 & 17.62 & 8.66 & 17.55 & 26.68 & 79.19\\
 & FreiHAND-HRNet + HInt FT~\cite{sun2019hrnet} & 1754 & 39.42 & 27.47 & 43.68 & 54.26 & 40.31\\
 & EffHandEgoNet-H2O~\cite{mucha2024effhand} & 1754 & 5.20 & 1.57 & 4.94 & 8.57 & 92.30\\
 & S2HAND-MT (Ours) & 1754 & 62.57 & 49.42 & 72.48 & 82.36 & 13.45\\
 & \textbf{S2HAND-Pose (Ours)} & 1754 & \textbf{70.04} & \textbf{59.83} & \textbf{81.14} & \textbf{88.69} & \textbf{6.61}\\
\midrule
VISOR TEST~\cite{pavlakos2024hamer} & Public Sapiens2-5B~\cite{khirodkar2026sapiens2} & 1906 & 56.12 & 41.10 & 65.12 & 75.89 & 17.68\\
 & RTMPose-Hand5 (public)~\cite{jiang2023rtmpose} & 1906 & 63.51 & 49.26 & 74.24 & 83.98 & 10.86\\
 & RTMPose-Hand5 + HInt FT~\cite{jiang2023rtmpose} & 1906 & 69.85 & 61.49 & 81.34 & 86.98 & 10.02\\
 & FreiHAND-HRNet reproduction~\cite{sun2019hrnet,potamias2025wilorendtoend3dhand} & 1906 & 28.32 & 14.14 & 29.89 & 42.92 & 54.09\\
 & FreiHAND-HRNet + HInt FT~\cite{sun2019hrnet} & 1906 & 56.57 & 43.76 & 65.30 & 75.20 & 18.57\\
 & EffHandEgoNet-H2O~\cite{mucha2024effhand} & 1906 & 32.25 & 15.06 & 36.34 & 49.39 & 42.55\\
 & S2HAND-MT (Ours) & 1906 & 59.42 & 45.50 & 69.31 & 79.10 & 16.21\\
 & \textbf{S2HAND-Pose (Ours)} & 1906 & \textbf{74.88} & \textbf{66.28} & \textbf{87.04} & \textbf{92.99} & \textbf{4.30}\\
\midrule
H2O TEST~\cite{kwon2021h2o} & Public Sapiens2-5B~\cite{khirodkar2026sapiens2} & 46742 & 69.02 & 51.54 & 83.96 & 93.03 & 4.38\\
 & RTMPose-Hand5 (public)~\cite{jiang2023rtmpose} & 46742 & 67.43 & 45.56 & 80.79 & 94.06 & 0.43\\
 & RTMPose-Hand5 + HInt FT~\cite{jiang2023rtmpose} & 46742 & 72.57 & 56.17 & 88.46 & 96.48 & 0.30\\
 & FreiHAND-HRNet reproduction~\cite{sun2019hrnet,potamias2025wilorendtoend3dhand} & 46742 & 60.19 & 42.24 & 70.66 & 82.83 & 7.29\\
 & FreiHAND-HRNet + HInt FT~\cite{sun2019hrnet} & 46742 & 68.74 & 51.94 & 82.82 & 92.15 & 1.80\\
 & \textbf{EffHandEgoNet-H2O}~\cite{mucha2024effhand} & 46742 & \textbf{75.25} & \textbf{61.33} & \textbf{92.34} & \textbf{97.98} & \textbf{0.06}\\
 & S2HAND-MT (Ours) & 46742 & 70.79 & 53.39 & 87.39 & 94.45 & 3.63\\
 & S2HAND-Pose (Ours) & 46742 & 73.96 & 58.08 & 90.79 & 97.62 & 0.49\\
\midrule
EPICK VAL~\cite{pavlakos2024hamer} & Public Sapiens2-5B~\cite{khirodkar2026sapiens2} & 625 & 55.16 & 40.00 & 63.99 & 75.20 & 18.56\\
 & RTMPose-Hand5 (public)~\cite{jiang2023rtmpose} & 625 & 64.10 & 48.69 & 74.87 & 85.52 & 8.48\\
 & RTMPose-Hand5 + HInt FT~\cite{jiang2023rtmpose} & 625 & 71.01 & 61.54 & 82.31 & 89.31 & 6.40\\
 & FreiHAND-HRNet reproduction~\cite{sun2019hrnet,potamias2025wilorendtoend3dhand} & 625 & 27.84 & 13.65 & 29.02 & 42.48 & 55.20\\
 & FreiHAND-HRNet + HInt FT~\cite{sun2019hrnet} & 625 & 57.14 & 43.60 & 65.97 & 76.01 & 15.20\\
 & EffHandEgoNet-H2O~\cite{mucha2024effhand} & 625 & 29.11 & 12.47 & 32.15 & 45.53 & 45.60\\
 & S2HAND-MT (Ours) & 625 & 58.54 & 43.98 & 68.37 & 78.65 & 16.96\\
 & \textbf{S2HAND-Pose (Ours)} & 625 & \textbf{74.35} & \textbf{64.83} & \textbf{86.61} & \textbf{93.11} & \textbf{4.00}\\
\midrule
H2SMV-FPV TEST & Public Sapiens2-5B~\cite{khirodkar2026sapiens2} & 9661 & 47.93 & 32.85 & 54.98 & 66.83 & 29.86\\
 & RTMPose-Hand5 (public)~\cite{jiang2023rtmpose} & 9661 & 40.57 & 28.13 & 47.65 & 56.29 & 39.00\\
 & RTMPose-Hand5 + HInt FT~\cite{jiang2023rtmpose} & 9661 & 39.78 & 28.15 & 46.47 & 54.67 & 39.64\\
 & FreiHAND-HRNet reproduction~\cite{sun2019hrnet,potamias2025wilorendtoend3dhand} & 9661 & 26.66 & 17.37 & 29.92 & 37.61 & 61.59\\
 & FreiHAND-HRNet + HInt FT~\cite{sun2019hrnet} & 9661 & 35.27 & 24.66 & 40.88 & 48.65 & 47.49\\
 & EffHandEgoNet-H2O~\cite{mucha2024effhand} & 9661 & 8.37 & 3.20 & 8.62 & 13.35 & 84.97\\
 & S2HAND-MT (Ours) & 9661 & 49.01 & 33.52 & 57.00 & 68.26 & 29.09\\
 & \textbf{S2HAND-Pose (Ours)} & 9661 & \textbf{56.14} & \textbf{40.59} & \textbf{65.79} & \textbf{76.61} & \textbf{17.14}\\
\bottomrule
\end{tabular}}
\end{table*}

%% file: tables/tab_published_hint_reference_appendix.tex
\begin{table*}[t]
\centering
\scriptsize
\setlength{\tabcolsep}{4pt}
\caption{\textbf{Published HInt references and evaluation boundaries.} PCK percentages at thresholds .05/.10/.15. Literature rows are transcribed, not rerun, and are not pooled with our Native AUC or ranked as a common-protocol benchmark. HandOS DINO-X reports only PCK@.10; dashes mean unreported, not zero. ExtPose uses predicted 2D evidence; its GT-evidence oracle is excluded. The exact ExtPose/ViTPose detection, crop, and aggregation match to our evaluator is unverified. The last two rows use identical cached S2HAND-Pose predictions and differ only in PCK averaging. TailGuard is off.}
\label{tab:published-hint-reference}
\begin{tabular*}{\textwidth}{@{\extracolsep{\fill}}lllllll@{}}
\toprule
 & \multicolumn{3}{c}{\textbf{NewDays (TEST)}} & \multicolumn{3}{c}{\textbf{VISOR (TEST)}}\\
\cmidrule(lr){2-4}\cmidrule(lr){5-7}
\textbf{System/Source} & \textbf{PCK@.05 $\uparrow$} & \textbf{PCK@.10 $\uparrow$} & \textbf{PCK@.15 $\uparrow$} & \textbf{PCK@.05 $\uparrow$} & \textbf{PCK@.10 $\uparrow$} & \textbf{PCK@.15 $\uparrow$}\\
\midrule
HaMeR + HInt (published)~\cite{pavlakos2024hamer,chen2025handos} & 51.6 & 81.9 & 91.9 & 56.5 & 88.1 & 95.6\\
HandOS-2D (published)~\cite{chen2025handos} & 55.8 & 75.8 & 84.5 & 66.2 & 85.3 & 91.8\\
HandOS-proj (published)~\cite{chen2025handos} & 53.7 & 75.9 & 85.1 & 64.8 & 85.4 & 92.0\\
HandOS DINO-X, 2D (published)~\cite{chen2025handos} & -- & 76.3 & -- & -- & 84.8 & --\\
HandOS DINO-X, proj (published)~\cite{chen2025handos} & -- & 76.5 & -- & -- & 84.6 & --\\
ViTPose reference in ExtPose~\cite{chen2025extpose,xu2022vitpose} & 66.5 & 86.5 & 93.1 & 70.8 & 90.6 & 96.2\\
ExtPose (published, predicted evidence)~\cite{chen2025extpose} & 59.6 & 84.8 & 92.7 & 61.1 & 88.5 & 95.6\\
\midrule
S2HAND-Pose (Ours), Native joint-micro & 59.83 & 81.14 & 88.69 & 66.28 & 87.04 & 92.99\\
S2HAND-Pose (Ours), Native joint-macro & 59.48 & 81.02 & 88.64 & 66.17 & 87.07 & 93.02\\
\bottomrule
\end{tabular*}
\end{table*}

%% file: tables/tab_p1_shared_oracle_roi_appendix.tex
\begin{table*}[t]
\centering
\caption{\textbf{Shared-GT Crop evaluation.} AUC@.20 $\uparrow$ (\%) uses identical $2\times$ GT-hand crops and tight-GT-box maximum-side normalization. Mean (4 sets) averages the first four dataset columns; Mean (5 sets) adds FreiHAND. EPICK is VAL and H2SMV-FPV is synthetic. Corrected HInt-FT rows contain only true Shared-GT reruns; a dash means not evaluated, never a substituted detected-box result.}
\label{tab:p1-shared-roi}
\small
\resizebox{\textwidth}{!}{\begin{tabular}{llllllll}
\toprule
\textbf{Model} & \textbf{H2O (TEST)}~\cite{kwon2021h2o} & \textbf{NewDays (TEST)}~\cite{pavlakos2024hamer} & \textbf{EPICK (VAL)}~\cite{pavlakos2024hamer} & \textbf{H2SMV-FPV (TEST)} & \textbf{FreiHAND (EVAL)}~\cite{zimmermann2019freihand} & \textbf{Mean (4 sets)} & \textbf{Mean (5 sets)}\\
\midrule
Public Sapiens2-5B~\cite{khirodkar2026sapiens2} & 70.53 & 65.69 & 65.73 & 51.96 & 60.62 & 63.48 & 62.91\\
RTMPose-Hand5 (public)~\cite{jiang2023rtmpose} & 66.64 & 57.30 & 61.98 & 42.97 & 65.59 & 57.22 & 58.90\\
RTMPose-Hand5 + HInt FT~\cite{jiang2023rtmpose} & -- & 60.21 & -- & -- & -- & -- & --\\
FreiHAND-HRNet + HInt FT~\cite{sun2019hrnet} & -- & 40.18 & -- & -- & -- & -- & --\\
EffHandEgoNet-H2O~\cite{mucha2024effhand} & 38.90 & 13.04 & 23.64 & 11.35 & 12.10 & 21.73 & 19.81\\
COCO-Hand HRNet (official)~\cite{sun2019hrnet} & 62.41 & 56.44 & 60.37 & 38.99 & 57.40 & 54.55 & 55.12\\
FreiHAND-HRNet reproduction~\cite{sun2019hrnet} & 63.18 & 32.76 & 39.19 & 33.32 & \textbf{71.94} & 42.11 & 48.08\\
\midrule
S2HAND-MT (Ours) & 70.93 & 66.92 & 68.40 & 53.02 & 61.76 & 64.82 & 64.21\\
\textbf{S2HAND-Pose (Ours)} & \textbf{74.55} & \textbf{74.27} & \textbf{78.10} & \textbf{58.36} & 66.14 & \textbf{71.32} & \textbf{70.28}\\
\bottomrule
\end{tabular}}
\end{table*}

%% file: tables/tab_p2b_shared_detector_appendix.tex
\begin{table*}[t]
\centering
\caption{\textbf{Shared-Detector Crop evaluation.} All models use identical WiLoR-YOLO $2\times$ detected ROIs. H2O TEST and H2SMV-FPV TEST use prospectively frozen 2,048-image subsets; NewDays TEST and EPICK VAL are complete. Coverage, AUC, PCK, side-swap rate, and tail rate are percentages; missing detections count as failures and are shared within each dataset. Side swap is computed only over detected hands in frames containing a labeled hand of the opposite side, and denotes cases closer on average to that opposite hand than to the assigned hand.}
\label{tab:p2b-shared}
\scriptsize
\resizebox{\textwidth}{!}{\begin{tabular}{llllllll}
\toprule
\textbf{Dataset} & \textbf{Model} & \textbf{GT Hands} & \textbf{Coverage (\%) $\uparrow$} & \textbf{AUC@.20 (\%) $\uparrow$} & \textbf{PCK@.05/.10/.15 (\%) $\uparrow$} & \textbf{Side Swap (\%) $\downarrow$} & \textbf{Tail$>$.20 (\%) $\downarrow$}\\
\midrule
H2O TEST subset~\cite{kwon2021h2o} & Public Sapiens2-5B~\cite{khirodkar2026sapiens2} & 4093 & 99.93 & 71.20 & 52.70/86.48/96.46 & 0.00 & 0.61\\
 & RTMPose-Hand5 (public)~\cite{jiang2023rtmpose} & 4093 & 99.93 & 52.49 & 33.35/63.62/74.14 & 17.81 & 26.46\\
 & FreiHAND-HRNet reproduction~\cite{sun2019hrnet} & 4093 & 99.93 & 57.17 & 40.90/68.71/77.87 & 10.67 & 20.35\\
 & EffHandEgoNet-H2O~\cite{mucha2024effhand} & 4093 & 99.93 & 73.94 & \textbf{61.58}/88.39/95.83 & 0.00 & 0.42\\
 & S2HAND-MT (Ours) & 4093 & 99.93 & 72.16 & 53.77/88.63/97.06 & 0.00 & 0.59\\
 & \textbf{S2HAND-Pose (Ours)} & 4093 & 99.93 & \textbf{74.80} & 60.11/\textbf{91.70}/\textbf{98.09} & 0.00 & \textbf{0.10}\\
\midrule
NewDays TEST~\cite{pavlakos2024hamer} & Public Sapiens2-5B~\cite{khirodkar2026sapiens2} & 1754 & 91.16 & 60.49 & 48.55/69.83/78.89 & -- & 16.19\\
 & RTMPose-Hand5 (public)~\cite{jiang2023rtmpose} & 1754 & 91.16 & 54.30 & 41.25/63.46/72.53 & -- & 22.81\\
 & FreiHAND-HRNet reproduction~\cite{sun2019hrnet} & 1754 & 91.16 & 30.35 & 16.68/33.18/44.97 & -- & 53.59\\
 & EffHandEgoNet-H2O~\cite{mucha2024effhand} & 1754 & 91.16 & 13.41 & 4.14/12.40/22.01 & -- & 80.22\\
 & S2HAND-MT (Ours) & 1754 & 91.16 & 61.63 & 50.57/71.08/79.74 & -- & 15.91\\
 & \textbf{S2HAND-Pose (Ours)} & 1754 & 91.16 & \textbf{67.91} & \textbf{60.52/78.09/84.11} & -- & \textbf{12.66}\\
\midrule
EPICK VAL~\cite{pavlakos2024hamer} & Public Sapiens2-5B~\cite{khirodkar2026sapiens2} & 625 & 95.04 & 63.84 & 48.11/74.59/85.58 & -- & 7.52\\
 & RTMPose-Hand5 (public)~\cite{jiang2023rtmpose} & 625 & 95.04 & 53.17 & 36.63/62.53/73.83 & -- & 22.88\\
 & FreiHAND-HRNet reproduction~\cite{sun2019hrnet} & 625 & 95.04 & 34.86 & 16.91/38.88/52.99 & -- & 40.48\\
 & EffHandEgoNet-H2O~\cite{mucha2024effhand} & 625 & 95.04 & 32.90 & 13.88/35.26/52.08 & -- & 38.88\\
 & S2HAND-MT (Ours) & 625 & 95.04 & 66.38 & 52.56/77.69/87.55 & -- & 7.20\\
 & \textbf{S2HAND-Pose (Ours)} & 625 & 95.04 & \textbf{74.42} & \textbf{66.57/86.55/92.06} & -- & \textbf{5.44}\\
\midrule
H2SMV-FPV TEST subset & Public Sapiens2-5B~\cite{khirodkar2026sapiens2} & 4095 & 71.67 & 43.39 & 31.29/50.07/59.55 & 4.46 & 35.48\\
 & RTMPose-Hand5 (public)~\cite{jiang2023rtmpose} & 4095 & 71.67 & 22.70 & 13.48/26.10/32.93 & 37.49 & 67.16\\
 & FreiHAND-HRNet reproduction~\cite{sun2019hrnet} & 4095 & 71.67 & 23.69 & 12.80/26.54/35.23 & 19.60 & 63.98\\
 & EffHandEgoNet-H2O~\cite{mucha2024effhand} & 4095 & 71.67 & 16.67 & 6.20/17.10/26.83 & 12.20 & 69.96\\
 & S2HAND-MT (Ours) & 4095 & 71.67 & 44.63 & 32.90/52.29/60.62 & 4.19 & 34.85\\
 & \textbf{S2HAND-Pose (Ours)} & 4095 & 71.67 & \textbf{47.77} & \textbf{36.86/56.55/63.54} & \textbf{1.81} & \textbf{31.65}\\
\bottomrule
\end{tabular}}
\end{table*}

%% file: tables/tab_matched_training_appendix.tex
\begin{table*}[t]
\centering
\scriptsize
\setlength{\tabcolsep}{3pt}
\caption{\textbf{Matched-data specialist adaptation.} Both specialists are fine-tuned on 12,000 balanced HInt TRAIN hands for 12,000 sample steps (batch size 32, FP16, seed 42) until HInt VAL convergence, selecting the peak validation checkpoint. Validation AUC below is reported by the MMPose training evaluator rather than the canonical Native or Shared-GT metric.}
\label{tab:matched-training}
\begin{tabular*}{\textwidth}{@{\extracolsep{\fill}}lll@{}}
\toprule
\textbf{Setting} & \textbf{RTMPose-Hand5 + HInt FT} & \textbf{FreiHAND-HRNet + HInt FT}\\
\midrule
Initialization & public Hand5 & reproduced FreiHAND-HRNet-W18\\
Optimizer & AdamW, decay 0.01 & Adam\\
Peak LR / final LR & $10^{-4}$ / $10^{-6}$ & $10^{-4}$ / $10^{-6}$\\
Schedule & 800-sample warm-up, cosine & 800-sample warm-up, cosine\\
Input / bbox padding & $256\times256$ / 1.25 & $256\times256$ / 1.25\\
BBox scale / rotate-factor & $[0.5,1.5]$ / $180^\circ$ & $[0.75,1.25]$ / $90^\circ$\\
Horizontal flip / photometric & yes / yes & yes / yes\\
VAL AUC at 3k sample steps & 49.83 & 23.74\\
VAL AUC at 6k sample steps & 53.95 & 33.09\\
VAL AUC at 9k sample steps & 55.29 & 36.38\\
VAL AUC at 12k sample steps & 55.38 & 37.07\\
Selected checkpoint (HInt VAL) & 12k (converged peak) & 12k (converged peak)\\
\bottomrule
\end{tabular*}
\end{table*}

%% file: tables/tab_stitch_diagnostics_appendix.tex
\begin{table*}[t]
\centering
\scriptsize
\setlength{\tabcolsep}{3pt}
\caption{\textbf{External detection and wrist disagreement.} Coverage is the percentage of canonical hands detected by the external stage. Wrist distances are divided by the larger side of the tight GT-hand box and summarized only on hands detected by both systems. These distances quantify disagreement, not which wrist is more accurate. TailGuard is off.}
\label{tab:stitch-diagnostics}
\begin{tabular*}{\textwidth}{@{\extracolsep{\fill}}llllll@{}}
\toprule
\textbf{External Model} & \textbf{Dataset} & \textbf{Coverage (\%) $\uparrow$} & \textbf{Gap Median} & \textbf{Gap P95} & \textbf{Gap P99}\\
\midrule
RTMPose-Hand5 (public) & NewDays TEST & 91.16 & 0.102 & 1.169 & 2.991\\
RTMPose-Hand5 (public) & VISOR TEST & 92.50 & 0.069 & 2.508 & 4.193\\
RTMPose-Hand5 (public) & H2O TEST & 99.96 & 0.047 & 0.152 & 3.029\\
RTMPose-Hand5 (public) & EPICK VAL & 95.04 & 0.073 & 2.912 & 3.862\\
RTMPose-Hand5 (public) & H2SMV-FPV TEST & 71.16 & 0.072 & 1.439 & 2.701\\
RTMPose-Hand5 + HInt FT & NewDays TEST & 91.16 & 0.114 & 1.135 & 2.995\\
RTMPose-Hand5 + HInt FT & VISOR TEST & 92.50 & 0.059 & 0.443 & 2.885\\
RTMPose-Hand5 + HInt FT & H2O TEST & 99.96 & 0.118 & 0.216 & 2.970\\
RTMPose-Hand5 + HInt FT & EPICK VAL & 95.04 & 0.100 & 2.839 & 3.818\\
RTMPose-Hand5 + HInt FT & H2SMV-FPV TEST & 71.16 & 0.105 & 1.445 & 2.740\\
WiLoR~\cite{potamias2025wilorendtoend3dhand} & NewDays TEST & 91.16 & 0.115 & 1.215 & 3.031\\
\bottomrule
\end{tabular*}
\end{table*}

%% file: tables/tab_stitched_cost_appendix.tex
\begin{table*}[t]
\centering
\scriptsize
\setlength{\tabcolsep}{3pt}
\caption{\textbf{External-stage cost accounting.} Measured on one L40S, 100 fixed NewDays images, three repeats, after warm-up. The input is an already decoded RGB image. WiLoR runtime varies with hand count. Component sums omit public preprocessing/decoding and final stitching.}
\label{tab:stitched-cost}
\begin{tabular*}{\textwidth}{@{\extracolsep{\fill}}llll@{}}
\toprule
\textbf{Stage} & \textbf{Parameters} & \textbf{Median (ms)} & \textbf{Mean (ms)}\\
\midrule
WiLoR-YOLO detector & 26.631M & 5.94 & 6.63\\
RTMPose-Hand5 forward (one hand) & 13.775M & 4.61 & --\\
WiLoR detector + regressor & 665.912M & 35.57 & 50.70\\
Public + Hand5 component sum & $\sim$5.161B & $\geq331.05$ & --\\
Public + WiLoR component sum & 5.787B & 356.07 & --\\
\bottomrule
\end{tabular*}
\end{table*}

%% file: tables/tab_s2hand_multiscale_native_appendix.tex
\begin{table*}[t]
\centering
\scriptsize
\setlength{\tabcolsep}{3.5pt}
\caption{\textbf{S2HAND scaling under Native Pipeline evaluation.} AUC@.20 is reported in percent; higher is better. Each adapted model is compared with the same-size Public Sapiens2 model using identical detector caches, crops, sample IDs, and missed-hand accounting. ``Native Mean (5 sets)'' is the unweighted mean over the five datasets, and \(\Delta\) is relative to the same-size public model. Unqualified S2HAND names denote 5B models.}
\label{tab:s2hand-multiscale-native}
\resizebox{\textwidth}{!}{%
\begin{tabular}{llllllll}
\toprule
\textbf{Model} & \textbf{NewDays (TEST)} & \textbf{VISOR (TEST)} & \textbf{H2O (TEST)} & \textbf{EPICK (VAL)} & \textbf{H2SMV-FPV (TEST)} & \textbf{Native Mean (5 sets)} & \textbf{\(\Delta\)} \\
\midrule
Public Sapiens2-0.4B~\cite{khirodkar2026sapiens2} & 57.94 & 55.70 & 69.69 & 54.84 & 40.90 & 55.81 & -- \\
S2HAND-Pose-0.4B (Ours) & 64.73 & 70.43 & 74.40 & 70.66 & 49.48 & 65.94 & +10.13 \\
\midrule
Public Sapiens2-0.8B~\cite{khirodkar2026sapiens2} & 59.06 & 54.78 & 68.03 & 54.21 & 41.78 & 55.57 & -- \\
S2HAND-Pose-0.8B (Ours) & 64.60 & 70.12 & 73.66 & 69.68 & 50.01 & 65.62 & +10.04 \\
\midrule
Public Sapiens2-1B~\cite{khirodkar2026sapiens2} & 61.71 & 52.53 & 57.17 & 52.26 & 46.76 & 54.09 & -- \\
S2HAND-Pose-1B (Ours) & 69.30 & 73.68 & 75.08 & 73.46 & 54.89 & 69.28 & +15.20 \\
\midrule
Public Sapiens2-5B~\cite{khirodkar2026sapiens2} & 61.48 & 56.12 & 69.02 & 55.16 & 47.93 & 57.94 & -- \\
S2HAND-MT (Ours) & 62.57 & 59.42 & 70.79 & 58.54 & 49.01 & 60.07 & +2.12 \\
S2HAND-Pose (Ours) & 70.04 & 74.88 & 73.96 & 74.35 & 56.14 & 69.87 & +11.93 \\
\bottomrule
\end{tabular}}
\end{table*}

%% file: tables/tab_fisher_probe_appendix.tex
\begin{table*}[t]
\centering
\caption{\textbf{Fisher-guided block selection.} Eight identical cached batches are used per Transformer block; no optimizer step is taken during probing. Per parameter denotes \(F_b/|\theta_b|\). HInt VAL AUC@.20 and trainable-parameter units match Table~\ref{tab:adaptation-ablation}.}
\label{tab:fisher-probe}
\small
\setlength{\tabcolsep}{4pt}
\begin{tabular*}{\textwidth}{@{\extracolsep{\fill}}llllll@{}}
\toprule
\textbf{Rank} & \textbf{Adaptation Scope} & \textbf{Fisher Trace} & \textbf{Per Parameter} & \textbf{Trainable Params (M)} & \textbf{HInt (VAL) AUC@.20 (\%)}\\
\midrule
1 & Transformer block 38 & $4.6094\times10^{-4}$ & $5.1934\times10^{-12}$ & 89.620 & 75.76\\
2 & Transformer block 40 & $4.3889\times10^{-4}$ & $4.9449\times10^{-12}$ & -- & --\\
3 & Transformer block 39 & $3.7660\times10^{-4}$ & $4.2431\times10^{-12}$ & -- & --\\
4 & Transformer block 36 & $3.5900\times10^{-4}$ & $4.0448\times10^{-12}$ & -- & --\\
5 & Transformer block 4 & $3.5005\times10^{-4}$ & $3.6974\times10^{-12}$ & -- & --\\
6 & Transformer block 37 & $3.4698\times10^{-4}$ & $3.9094\times10^{-12}$ & -- & --\\
-- & Transformer blocks 38--40 & top contiguous triplet & -- & 267.130 & 76.56\\
-- & Transformer blocks \{4,38,40\} & non-contiguous control & -- & 273.052 & \textbf{76.60}\\
-- & Transformer blocks 53--55 & ranks 56/54/51 & -- & 284.890 & 72.15\\
\bottomrule
\end{tabular*}
\end{table*}

%% file: tables/tab_fisher_full_ranks_appendix.tex
\begin{table*}[t]
\centering
\scriptsize
\setlength{\tabcolsep}{3pt}
\caption{\textbf{Complete 0.4B Fisher ranking.} All Transformer blocks are ranked from the same eight cached training batches without an optimizer step. Fisher trace is the mean sum of squared hand-loss gradients.}
\label{tab:fisher-full-0.4b}
\begin{tabular*}{\textwidth}{@{\extracolsep{\fill}}llllllllllll@{}}
\toprule
\textbf{Rank} & \textbf{Block} & \textbf{Fisher Trace} & \textbf{Rank} & \textbf{Block} & \textbf{Fisher Trace} & \textbf{Rank} & \textbf{Block} & \textbf{Fisher Trace} & \textbf{Rank} & \textbf{Block} & \textbf{Fisher Trace} \\
\midrule
1 & 18 & $2.031e-03$ & 7 & 7 & $1.242e-03$ & 13 & 20 & $1.015e-03$ & 19 & 3 & $5.315e-04$ \\
2 & 17 & $1.820e-03$ & 8 & 13 & $1.230e-03$ & 14 & 9 & $9.795e-04$ & 20 & 0 & $5.126e-04$ \\
3 & 16 & $1.815e-03$ & 9 & 12 & $1.118e-03$ & 15 & 11 & $9.745e-04$ & 21 & 22 & $4.242e-04$ \\
4 & 15 & $1.547e-03$ & 10 & 6 & $1.071e-03$ & 16 & 10 & $9.538e-04$ & 22 & 1 & $3.357e-04$ \\
5 & 19 & $1.461e-03$ & 11 & 5 & $1.043e-03$ & 17 & 4 & $7.954e-04$ & 23 & 2 & $3.203e-04$ \\
6 & 14 & $1.441e-03$ & 12 & 8 & $1.028e-03$ & 18 & 21 & $6.877e-04$ & 24 & 23 & $1.147e-04$ \\
\bottomrule
\end{tabular*}
\end{table*}

\begin{table*}[t]
\centering
\scriptsize
\setlength{\tabcolsep}{3pt}
\caption{\textbf{Complete 0.8B Fisher ranking.} All Transformer blocks are ranked from the same eight cached training batches without an optimizer step. Fisher trace is the mean sum of squared hand-loss gradients.}
\label{tab:fisher-full-0.8b}
\begin{tabular*}{\textwidth}{@{\extracolsep{\fill}}llllllllllll@{}}
\toprule
\textbf{Rank} & \textbf{Block} & \textbf{Fisher Trace} & \textbf{Rank} & \textbf{Block} & \textbf{Fisher Trace} & \textbf{Rank} & \textbf{Block} & \textbf{Fisher Trace} & \textbf{Rank} & \textbf{Block} & \textbf{Fisher Trace} \\
\midrule
1 & 22 & $7.380e-04$ & 9 & 15 & $5.727e-04$ & 17 & 10 & $4.200e-04$ & 25 & 28 & $2.069e-04$ \\
2 & 19 & $7.115e-04$ & 10 & 20 & $5.637e-04$ & 18 & 8 & $4.077e-04$ & 26 & 1 & $1.996e-04$ \\
3 & 23 & $7.052e-04$ & 11 & 11 & $5.296e-04$ & 19 & 26 & $4.007e-04$ & 27 & 4 & $1.779e-04$ \\
4 & 17 & $6.747e-04$ & 12 & 14 & $5.217e-04$ & 20 & 7 & $3.256e-04$ & 28 & 3 & $1.277e-04$ \\
5 & 18 & $6.209e-04$ & 13 & 9 & $4.890e-04$ & 21 & 27 & $2.768e-04$ & 29 & 29 & $1.263e-04$ \\
6 & 24 & $6.042e-04$ & 14 & 25 & $4.777e-04$ & 22 & 6 & $2.656e-04$ & 30 & 2 & $9.998e-05$ \\
7 & 16 & $5.828e-04$ & 15 & 13 & $4.692e-04$ & 23 & 0 & $2.574e-04$ & 31 & 30 & $5.598e-05$ \\
8 & 21 & $5.819e-04$ & 16 & 12 & $4.391e-04$ & 24 & 5 & $2.330e-04$ & 32 & 31 & $2.410e-05$ \\
\bottomrule
\end{tabular*}
\end{table*}

\begin{table*}[t]
\centering
\scriptsize
\setlength{\tabcolsep}{3pt}
\caption{\textbf{Complete 1B Fisher ranking.} All Transformer blocks are ranked from the same eight cached training batches without an optimizer step. Fisher trace is the mean sum of squared hand-loss gradients.}
\label{tab:fisher-full-1b}
\begin{tabular*}{\textwidth}{@{\extracolsep{\fill}}llllllllllll@{}}
\toprule
\textbf{Rank} & \textbf{Block} & \textbf{Fisher Trace} & \textbf{Rank} & \textbf{Block} & \textbf{Fisher Trace} & \textbf{Rank} & \textbf{Block} & \textbf{Fisher Trace} & \textbf{Rank} & \textbf{Block} & \textbf{Fisher Trace} \\
\midrule
1 & 28 & $7.042e-04$ & 11 & 23 & $3.010e-04$ & 21 & 17 & $1.679e-04$ & 31 & 4 & $1.037e-04$ \\
2 & 29 & $6.668e-04$ & 12 & 22 & $2.442e-04$ & 22 & 10 & $1.679e-04$ & 32 & 6 & $1.030e-04$ \\
3 & 30 & $6.378e-04$ & 13 & 33 & $2.178e-04$ & 23 & 20 & $1.645e-04$ & 33 & 39 & $7.989e-05$ \\
4 & 31 & $5.522e-04$ & 14 & 21 & $1.994e-04$ & 24 & 11 & $1.619e-04$ & 34 & 2 & $7.307e-05$ \\
5 & 27 & $5.157e-04$ & 15 & 13 & $1.993e-04$ & 25 & 19 & $1.449e-04$ & 35 & 3 & $7.235e-05$ \\
6 & 26 & $4.875e-04$ & 16 & 12 & $1.951e-04$ & 26 & 0 & $1.442e-04$ & 36 & 1 & $6.207e-05$ \\
7 & 32 & $4.537e-04$ & 17 & 15 & $1.750e-04$ & 27 & 7 & $1.360e-04$ & 37 & 38 & $4.785e-05$ \\
8 & 9 & $4.274e-04$ & 18 & 18 & $1.748e-04$ & 28 & 5 & $1.296e-04$ & 38 & 35 & $3.694e-05$ \\
9 & 25 & $4.082e-04$ & 19 & 14 & $1.703e-04$ & 29 & 34 & $1.277e-04$ & 39 & 36 & $3.445e-05$ \\
10 & 24 & $3.854e-04$ & 20 & 16 & $1.683e-04$ & 30 & 8 & $1.151e-04$ & 40 & 37 & $3.138e-05$ \\
\bottomrule
\end{tabular*}
\end{table*}

\begin{table*}[t]
\centering
\scriptsize
\setlength{\tabcolsep}{3pt}
\caption{\textbf{Complete 5B Fisher ranking.} All Transformer blocks are ranked from the same eight cached training batches without an optimizer step. Fisher trace is the mean sum of squared hand-loss gradients.}
\label{tab:fisher-full-5b}
\begin{tabular*}{\textwidth}{@{\extracolsep{\fill}}llllllllllll@{}}
\toprule
\textbf{Rank} & \textbf{Block} & \textbf{Fisher Trace} & \textbf{Rank} & \textbf{Block} & \textbf{Fisher Trace} & \textbf{Rank} & \textbf{Block} & \textbf{Fisher Trace} & \textbf{Rank} & \textbf{Block} & \textbf{Fisher Trace} \\
\midrule
1 & 38 & $4.609e-04$ & 15 & 43 & $1.772e-04$ & 29 & 19 & $5.536e-05$ & 43 & 47 & $2.076e-05$ \\
2 & 40 & $4.389e-04$ & 16 & 29 & $1.429e-04$ & 30 & 18 & $4.667e-05$ & 44 & 10 & $2.005e-05$ \\
3 & 39 & $3.766e-04$ & 17 & 30 & $1.422e-04$ & 31 & 17 & $4.579e-05$ & 45 & 46 & $2.002e-05$ \\
4 & 36 & $3.590e-04$ & 18 & 27 & $1.316e-04$ & 32 & 1 & $4.238e-05$ & 46 & 7 & $1.912e-05$ \\
5 & 4 & $3.500e-04$ & 19 & 26 & $1.283e-04$ & 33 & 14 & $3.594e-05$ & 47 & 8 & $1.874e-05$ \\
6 & 37 & $3.470e-04$ & 20 & 28 & $1.183e-04$ & 34 & 5 & $3.428e-05$ & 48 & 9 & $1.848e-05$ \\
7 & 35 & $2.778e-04$ & 21 & 24 & $1.160e-04$ & 35 & 16 & $3.109e-05$ & 49 & 11 & $1.835e-05$ \\
8 & 41 & $2.710e-04$ & 22 & 25 & $1.108e-04$ & 36 & 15 & $3.069e-05$ & 50 & 49 & $1.769e-05$ \\
9 & 33 & $2.573e-04$ & 23 & 23 & $1.007e-04$ & 37 & 12 & $2.762e-05$ & 51 & 55 & $1.121e-05$ \\
10 & 32 & $2.516e-04$ & 24 & 22 & $8.888e-05$ & 38 & 13 & $2.655e-05$ & 52 & 50 & $1.011e-05$ \\
11 & 34 & $2.451e-04$ & 25 & 44 & $7.755e-05$ & 39 & 3 & $2.542e-05$ & 53 & 51 & $6.659e-06$ \\
12 & 0 & $2.173e-04$ & 26 & 21 & $7.587e-05$ & 40 & 48 & $2.410e-05$ & 54 & 54 & $4.705e-06$ \\
13 & 31 & $1.923e-04$ & 27 & 45 & $7.168e-05$ & 41 & 2 & $2.341e-05$ & 55 & 52 & $3.618e-06$ \\
14 & 42 & $1.901e-04$ & 28 & 20 & $6.419e-05$ & 42 & 6 & $2.260e-05$ & 56 & 53 & $3.154e-06$ \\
\bottomrule
\end{tabular*}
\end{table*}

%% file: tables/tab_mt_repeatability_appendix.tex
\begin{table*}[t]
\centering
\scriptsize
\setlength{\tabcolsep}{3pt}
\caption{\textbf{S2HAND-MT seed repeatability.} AUC@.20 $\uparrow$ (\%) under the fixed controlled-crop evaluation. The same exact-routing architecture is trained with seeds 42, 43, and 44. SD is the sample standard deviation across three runs. These are not Native Pipeline scores. Repeated MT seeds alone do not establish statistical equivalence to ordinary head tuning.}
\label{tab:mt-repeatability}
\begin{tabular*}{\textwidth}{@{\extracolsep{\fill}}lllll@{}}
\toprule
\textbf{Seed} & \textbf{HInt (VAL)} & \textbf{HInt (TEST)} & \textbf{COCO (VAL)} & \textbf{FreiHAND (EVAL)}\\
\midrule
42 & 68.451 & 67.491 & 47.452 & 61.761\\
43 & 68.465 & 67.464 & 47.412 & 61.760\\
44 & 68.350 & 67.384 & 47.336 & 61.740\\
\midrule
Mean $\pm$ SD & $68.422\pm0.063$ & $67.446\pm0.055$ & $47.400\pm0.059$ & $61.753\pm0.012$\\
\bottomrule
\end{tabular*}
\end{table*}

%% file: tables/tab_coco_hand_complete_appendix.tex
\begin{table*}[t]
\centering
\scriptsize
\setlength{\tabcolsep}{3pt}
\caption{\textbf{Complete COCO WholeBody hand export.} AP/AR are percentages using the same 23-point body+foot mapping and 21-point hand mappings. Combined hand AP/AR average the separately evaluated left and right hand scores; \texttt{---} denotes body+foot outputs preserved by construction via the frozen public route in S2HAND-MT.}
\label{tab:coco-hand-complete}
\begin{tabular*}{\textwidth}{@{\extracolsep{\fill}}lllllllll@{}}
\toprule
\textbf{System} & \textbf{L AP} & \textbf{L AR} & \textbf{R AP} & \textbf{R AR} & \textbf{Hand AP} & \textbf{Hand AR} & \textbf{Body AP} & \textbf{Body AR}\\
\midrule
Public Sapiens2-5B & 53.88 & 58.67 & 49.96 & 55.68 & 51.92 & 57.18 & 79.09 & 82.83\\
S2HAND-MT (Ours) & 56.56 & 61.96 & 52.95 & 58.86 & 54.75 & 60.41 & --- & ---\\
S2HAND-Pose (Ours) & 66.11 & 71.25 & 63.51 & 69.51 & 64.81 & 70.38 & 79.28 & 83.07\\
\bottomrule
\end{tabular*}
\end{table*}

%% file: tables/tab_dense_audit_appendix.tex
\begin{table*}[t]
\centering
\scriptsize
\setlength{\tabcolsep}{3pt}
\caption{\textbf{512-image dense-head agreement audit.} All values compare S2HAND-Pose with the corresponding Public Sapiens2-5B head and are not human-GT accuracies. Per-image means and percentiles are macro-averaged unless marked pixel-weighted. Normal/pointmap use a fixed public-segmentation foreground mask. Segmentation foreground agreement uses the union of predicted foreground; hand-mask IoU uses the union of left/right hand labels.}
\label{tab:dense-audit}
\begin{tabular*}{\textwidth}{@{\extracolsep{\fill}}lll@{}}
\toprule
\textbf{Head} & \textbf{Metric} & \textbf{Value}\\
\midrule
Segmentation & all-pixel agreement $\uparrow$ & 99.9612\%\\
 & foreground agreement $\uparrow$ & 99.5956\%\\
 & hand-mask IoU $\uparrow$ & 99.3246\%\\
Normal & mean angular drift $\downarrow$ & $0.6118^\circ$\\
 & mean image-P95 angular drift $\downarrow$ & $1.6569^\circ$\\
 & foreground fraction within $5^\circ$ $\uparrow$ & 99.9755\%\\
Pointmap & mean normalized endpoint drift $\downarrow$ & 0.4791\%\\
 & foreground-pixel-weighted drift $\downarrow$ & 0.3777\%\\
 & mean image-P95 drift $\downarrow$ & 1.1029\%\\
 & image-mean drift $>1\%$ / $>5\%$ $\downarrow$ & 40/512 / 5/512\\
 & worst image-mean drift $\downarrow$ & 11.8552\%\\
\bottomrule
\end{tabular*}
\end{table*}

%% file: tables/tab_multiscale_preservation_appendix.tex
\begin{table*}[t]
\centering
\small
\caption{\textbf{Preservation across Sapiens2 scales.} COCO body+foot AP is
the evaluator's \texttt{nonhand\_AP} over the 17 body and six foot points.
Face NME uses 2,004 people and the 16 semantic anchors shared by COCO-68 and
the Sapiens2 238-point layout. For each person, dense-face drift is the mean
Euclidean displacement of the 238 decoded points from the same-size public
model, divided by the annotated face-box diagonal; the table averages this
quantity over the 2,004 people. It has no human-GT interpretation.}
\label{tab:multiscale-preservation}
\begin{tabular*}{\textwidth}{@{\extracolsep{\fill}}llllllll@{}}
\toprule
& \multicolumn{3}{c}{\textbf{COCO Body+Foot AP (\%)}}
& \multicolumn{3}{c}{\textbf{Face 16-Anchor NME $\downarrow$}}
& \textbf{Dense Drift $\downarrow$} \\
\cmidrule(lr){2-4}\cmidrule(lr){5-7}
\textbf{Scale} & \textbf{Public} & \textbf{S2HAND-Pose} & \textbf{$\Delta$} & \textbf{Public} & \textbf{S2HAND-Pose} & \textbf{$\Delta$} & \textbf{238 Points} \\
\midrule
0.4B & 78.98 & 77.36 & $-1.62$ & 0.1515 & 0.1252 & $-0.0264$ & 0.2082 \\
0.8B & 79.03 & 78.05 & $-0.98$ & 0.1550 & 0.1325 & $-0.0225$ & 0.1057 \\
1B   & 79.23 & 79.34 & $+0.12$ & 0.1297 & 0.1324 & $+0.0027$ & 0.1108 \\
5B   & 79.09 & 79.28 & $+0.19$ & 0.1428 & 0.1510 & $+0.0082$ & 0.1132 \\
\bottomrule
\end{tabular*}
\end{table*}

%% file: tables/tab_tailguard_ablation.tex
\begin{table*}[t]
\scriptsize
\setlength{\tabcolsep}{2pt}
\centering
\caption{\textbf{TailGuard accuracy ablation.} We evaluate the Native Pipeline with S2HAND-MT. The confidence threshold (0.510756) and WiLoR-YOLO $2\times$ detected-hand crop were selected on EPICK VAL and frozen before testing. TailGuard does not change the S2HAND-MT weights. All metrics are percentages.}
\label{tab:tailguard-ablation}
\begin{tabular*}{\textwidth}{@{\extracolsep{\fill}}llllll@{}}
\toprule
\textbf{Dataset} & \textbf{System} & \textbf{AUC@.20 $\uparrow$} & \textbf{PCK@.05 $\uparrow$}
& \textbf{PCK@.10 $\uparrow$} & \textbf{Tail$>$.20 $\downarrow$} \\
\midrule
H2O TEST
& S2HAND-MT
& 70.79 & 53.39 & 87.39 & 3.63 \\
& + TailGuard
& \textbf{71.05} & \textbf{53.55} & \textbf{87.70}
& \textbf{3.33} \\
\midrule
NewDays TEST
& S2HAND-MT
& 62.57 & 49.42 & 72.48 & 13.45 \\
& + TailGuard
& \textbf{64.55} & \textbf{51.07} & \textbf{74.96}
& \textbf{10.43} \\
\bottomrule
\end{tabular*}
\end{table*}

%% file: tables/tab_tailguard_val_sweep_appendix.tex
\begin{table*}[t]
\centering
\scriptsize
\setlength{\tabcolsep}{4pt}
\caption{\textbf{TailGuard validation-only threshold and crop sweep.} All choices use HInt-EPICK VAL only. AUC@.20 uses the same joint-micro aggregation as the main paper; AUC and rates are percentages. The deployed $2.0\times$/0.510756 setting was frozen before every TEST evaluation. The per-hand AUC oracle selects the higher clipped per-hand AUC and is an upper bound for that objective. False re-crop denotes a triggered sample that crosses from native NME $\leq .20$ to re-crop NME $>.20$; because this diagnostic is not the oracle objective, it can be non-zero.}
\label{tab:tailguard-val-sweep}
\begin{tabular*}{\textwidth}{@{\extracolsep{\fill}}clllll@{}}
\toprule
\textbf{Context} & \textbf{Policy} & \textbf{AUC@.20} & \textbf{Tail$>$.20} & \textbf{Trigger (\%)} & \textbf{False Re-Crop} \\
\midrule
1.5 & native & 58.54 & 16.96 & 0.00 & 0.00 \\
1.5 & always if detected & 67.64 & 3.36 & 95.04 & 0.51 \\
1.5 & per-hand AUC oracle & 68.72 & 3.04 & 67.04 & 0.00 \\
1.5 & conf. $\leq$0.350000 & 61.63 & 12.64 & 16.16 & 0.00 \\
1.5 & conf. $\leq$0.400000 & 62.66 & 11.04 & 22.08 & 0.00 \\
1.5 & conf. $\leq$0.450000 & 63.34 & 9.92 & 28.96 & 0.55 \\
1.5 & conf. $\leq$0.510756 & 64.43 & 7.68 & 38.56 & 0.41 \\
1.5 & conf. $\leq$0.550000 & 65.06 & 6.56 & 42.88 & 0.37 \\
1.5 & conf. $\leq$0.600000 & 66.15 & 5.12 & 49.44 & 0.32 \\
1.5 & conf. $\leq$0.650000 & 66.55 & 4.64 & 54.08 & 0.30 \\
\midrule
2.0 & native & 58.54 & 16.96 & 0.00 & 0.00 \\
2.0 & always if detected & 68.71 & 3.20 & 95.04 & 0.51 \\
2.0 & per-hand AUC oracle & 69.54 & 3.04 & 70.56 & 0.23 \\
2.0 & conf. $\leq$0.350000 & 61.92 & 12.64 & 16.16 & 0.99 \\
2.0 & conf. $\leq$0.400000 & 63.04 & 10.88 & 22.08 & 0.72 \\
2.0 & conf. $\leq$0.450000 & 63.78 & 9.92 & 28.96 & 1.10 \\
2.0 & conf. $\leq$0.510756 & 65.02 & 7.68 & 38.56 & 0.83 \\
2.0 & conf. $\leq$0.550000 & 65.75 & 6.56 & 42.88 & 0.75 \\
2.0 & conf. $\leq$0.600000 & 66.86 & 5.12 & 49.44 & 0.65 \\
2.0 & conf. $\leq$0.650000 & 67.35 & 4.64 & 54.08 & 0.59 \\
\midrule
2.5 & native & 58.54 & 16.96 & 0.00 & 0.00 \\
2.5 & always if detected & 68.55 & 3.20 & 95.04 & 0.51 \\
2.5 & per-hand AUC oracle & 69.46 & 2.88 & 68.00 & 0.24 \\
2.5 & conf. $\leq$0.350000 & 61.93 & 12.80 & 16.16 & 0.99 \\
2.5 & conf. $\leq$0.400000 & 63.12 & 11.04 & 22.08 & 0.72 \\
2.5 & conf. $\leq$0.450000 & 63.86 & 9.92 & 28.96 & 1.10 \\
2.5 & conf. $\leq$0.510756 & 65.11 & 7.68 & 38.56 & 0.83 \\
2.5 & conf. $\leq$0.550000 & 65.84 & 6.56 & 42.88 & 0.75 \\
2.5 & conf. $\leq$0.600000 & 66.85 & 5.12 & 49.44 & 0.65 \\
2.5 & conf. $\leq$0.650000 & 67.30 & 4.64 & 54.08 & 0.59 \\
\bottomrule
\end{tabular*}
\end{table*}

%% file: tables/tab_tailguard_pose_transfer_appendix.tex
\begin{table*}[t]
\centering
\scriptsize
\setlength{\tabcolsep}{4pt}
\caption{\textbf{Frozen TailGuard transfer to S2HAND-Pose.} The confidence threshold and \(2\times\) same-side crop selected for S2HAND-MT on EPICK VAL are applied to S2HAND-Pose without retuning. AUC and PCK are percentages; the trigger rate is the percentage of hands receiving the conditional second pass.}
\label{tab:tailguard-pose-transfer}
\begin{tabular*}{\textwidth}{@{\extracolsep{\fill}}lllllll@{}}
\toprule
\textbf{Dataset} & \textbf{System} & \textbf{AUC@.20 \(\uparrow\)} & \textbf{PCK@.05 \(\uparrow\)} & \textbf{PCK@.10 \(\uparrow\)} & \textbf{PCK@.15 \(\uparrow\)} & \textbf{Trigger (\%)} \\
\midrule
NewDays TEST~\cite{pavlakos2024hamer}
& S2HAND-Pose (Ours) & 70.04 & 59.83 & 81.14 & 88.69 & 0.00 \\
& + TailGuard & \textbf{70.69} & \textbf{60.41} & \textbf{82.01} & \textbf{89.48} & 9.29 \\
\midrule
H2O TEST~\cite{kwon2021h2o}
& S2HAND-Pose (Ours) & 73.96 & 58.08 & 90.79 & 97.62 & 0.00 \\
& + TailGuard & \textbf{74.23} & \textbf{58.30} & \textbf{91.11} & \textbf{97.97} & 0.39 \\
\midrule
H2SMV-FPV TEST
& S2HAND-Pose (Ours) & 56.14 & 40.59 & 65.79 & 76.61 & 0.00 \\
& + TailGuard & \textbf{56.57} & \textbf{40.91} & \textbf{66.33} & \textbf{77.18} & 2.33 \\
\bottomrule
\end{tabular*}
\end{table*}

%% file: tables/tab_tailguard_cost_appendix.tex
\begin{table*}[t]
\scriptsize
\setlength{\tabcolsep}{4pt}
\centering
\caption{\textbf{TailGuard repeat timing audit.} Pipeline latency is measured on one NVIDIA A100-SXM4-40GB GPU in an independent repeat of the main timing audit. Timing starts from a decoded BGR image and a cached person box and ends with decoded joint coordinates. The trigger rate is 46\%. Person detection and disk decoding are excluded. Every row reports wall-clock milliseconds over the same 1,000 samples; the triggered-only row contains the 460 samples that invoke the conditional second S2HAND-MT pass. P99 and maximum expose rare system stalls that make the mean less stable than the median and P95.}
\label{tab:tailguard-cost}
\begin{tabular*}{\textwidth}{@{\extracolsep{\fill}}lllllll@{}}
\toprule
\textbf{Stage} & \textbf{Mean (ms)} & \textbf{Median (ms)} & \textbf{P95 (ms)} & \textbf{P99 (ms)} & \textbf{Max (ms)} & \textbf{N} \\
\midrule
Native S2HAND-MT                 & 737.76  & 892.52  & 933.45  & 1804.77 & 6833.52  & 1000 \\
WiLoR-YOLO                       & 16.21   & 10.15   & 10.62   & 11.76   & 5478.96  & 1000 \\
Conditional S2HAND-MT re-crop    & 221.24  & 0.00    & 487.23  & 489.62  & 492.13   & 1000 \\
S2HAND-MT + TailGuard            & 975.21  & 963.38  & 1406.02 & 1815.54 & 12312.48 & 1000 \\
TailGuard-triggered samples only & 1246.91 & 1391.30 & 1408.99 & 1775.58 & 2305.06  & 460  \\
\bottomrule
\end{tabular*}
\end{table*}

%% file: tables/tab_pose2mesh_ci_appendix.tex
\begin{table*}[t]
\centering
\scriptsize
\setlength{\tabcolsep}{3pt}
\caption{\textbf{Frozen Pose2Mesh paired uncertainty.} Differences are baseline minus S2HAND-Pose PA-MPJPE (mm); positive favors S2HAND-Pose. Intervals use 10,000 paired cluster-bootstrap replicates on common finite samples. FreiHAND uses individual examples; other sets use sequence/capture groups where available. The unfavorable DexYCB comparison is retained. Deltas use unrounded errors.}
\label{tab:pose2mesh-ci}
\begin{tabular*}{\textwidth}{@{\extracolsep{\fill}}lllll@{}}
\toprule
\textbf{Dataset} & \textbf{Hands} & \textbf{Clusters} & \textbf{Public $-$ Pose [95\% CI]} & \textbf{Hand5 $-$ Pose [95\% CI]}\\
\midrule
FreiHAND EVAL & 3960 & 3960 & $+3.322\ [3.155,3.488]$ & $+0.344\ [0.231,0.461]$\\
H2O-4096 & 4096 & 46 & $+1.325\ [1.056,1.609]$ & $+1.107\ [0.793,1.452]$\\
DexYCB-997 & 997 & 197 & $+2.617\ [2.332,2.908]$ & $-0.851\ [-1.087,-0.617]$\\
AssemblyHands-8k & 7982 & 293 & $+2.444\ [2.172,2.747]$ & $+2.017\ [1.718,2.329]$\\
H2SMV-FPV & 9379 & 306 & $+2.077\ [1.946,2.213]$ & $+3.840\ [3.630,4.051]$\\
\bottomrule
\end{tabular*}
\end{table*}

%% file: tables/tab_unisign_wlasl_appendix.tex
\begin{table*}[t]
\centering
\caption{\textbf{Frozen sign-recognition transfer.} We evaluate Uni-Sign~\cite{li2025unisign} pose-only recognition on the WLASL-2000~\cite{li2020wlasl} strict local overlap (84 videos, 68 classes). Only the 42 hand XY coordinates change; Uni-Sign native RTMPose confidences and all non-hand coordinates remain fixed. Metrics are percentages. Paired 95\% bootstrap intervals for S2HAND-Pose comparisons include zero.}
\label{tab:unisign}
\small
\begin{tabular*}{\textwidth}{@{\extracolsep{\fill}}lll@{}}
\toprule
\textbf{Hand XY Source} & \textbf{Instance Top-1 (\%) $\uparrow$} & \textbf{Macro Top-1 (\%) $\uparrow$}\\
\midrule
Uni-Sign native RTMPose & \textbf{71.43} & \textbf{76.72}\\
Public Sapiens2-5B & \textbf{71.43} & 75.98\\
S2HAND-MT (Ours) & 69.05 & 73.77\\
S2HAND-Pose (Ours) & \textbf{71.43} & \textbf{76.72}\\
\bottomrule
\end{tabular*}
\end{table*}

%% file: sec/9_appendix_figures.tex
\begin{figure*}[p]
  \centering
  \includegraphics[width=\textwidth]{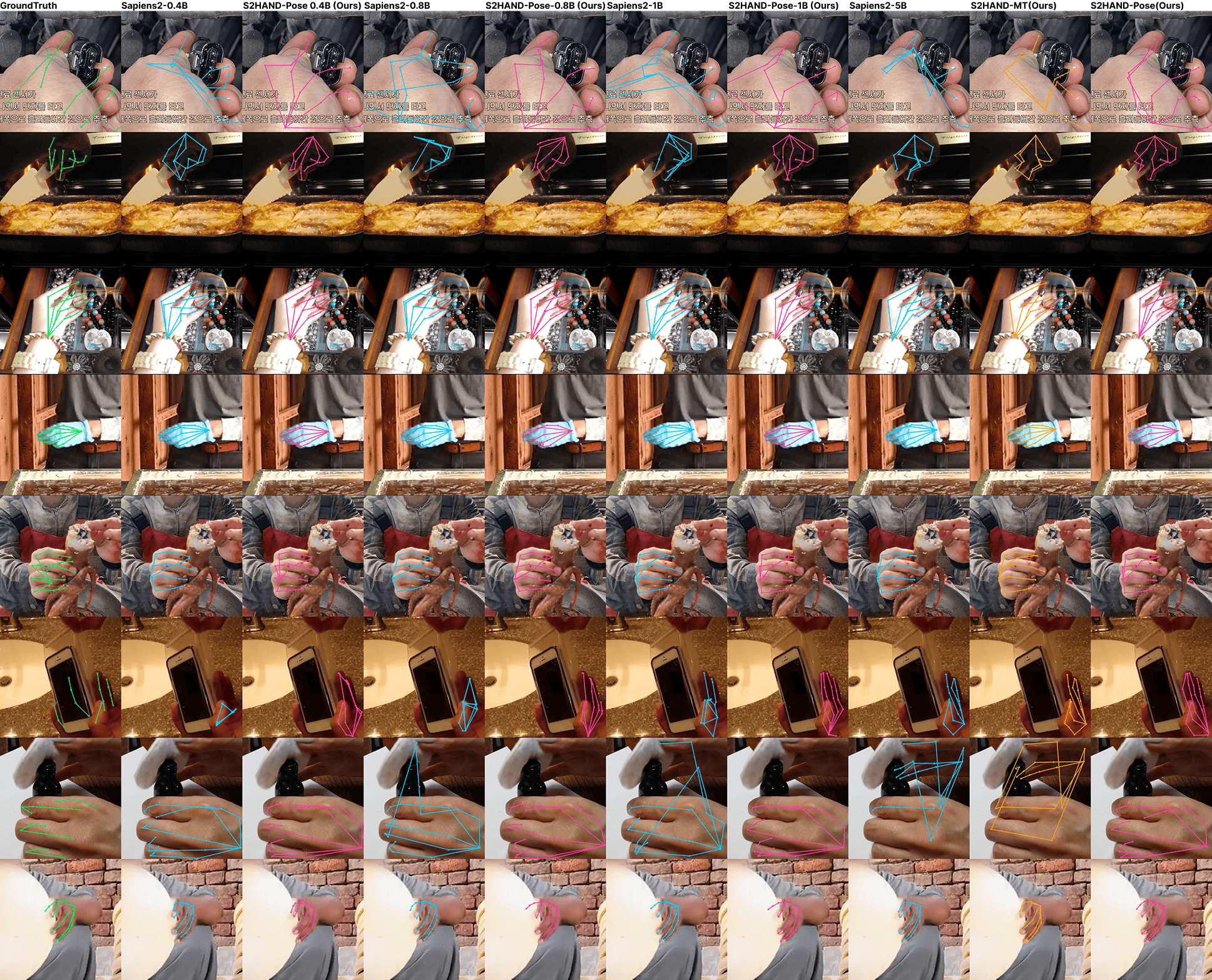}
  \caption{\textbf{Multi-scale NewDays comparison.} Eight deterministic
  NewDays TEST hands rendered under the fixed ten-column scale-control
  ordering described above.}
  \label{fig:app-scale-newdays}
\end{figure*}
\clearpage
\begin{figure*}[p]
  \centering
  \includegraphics[width=\textwidth]{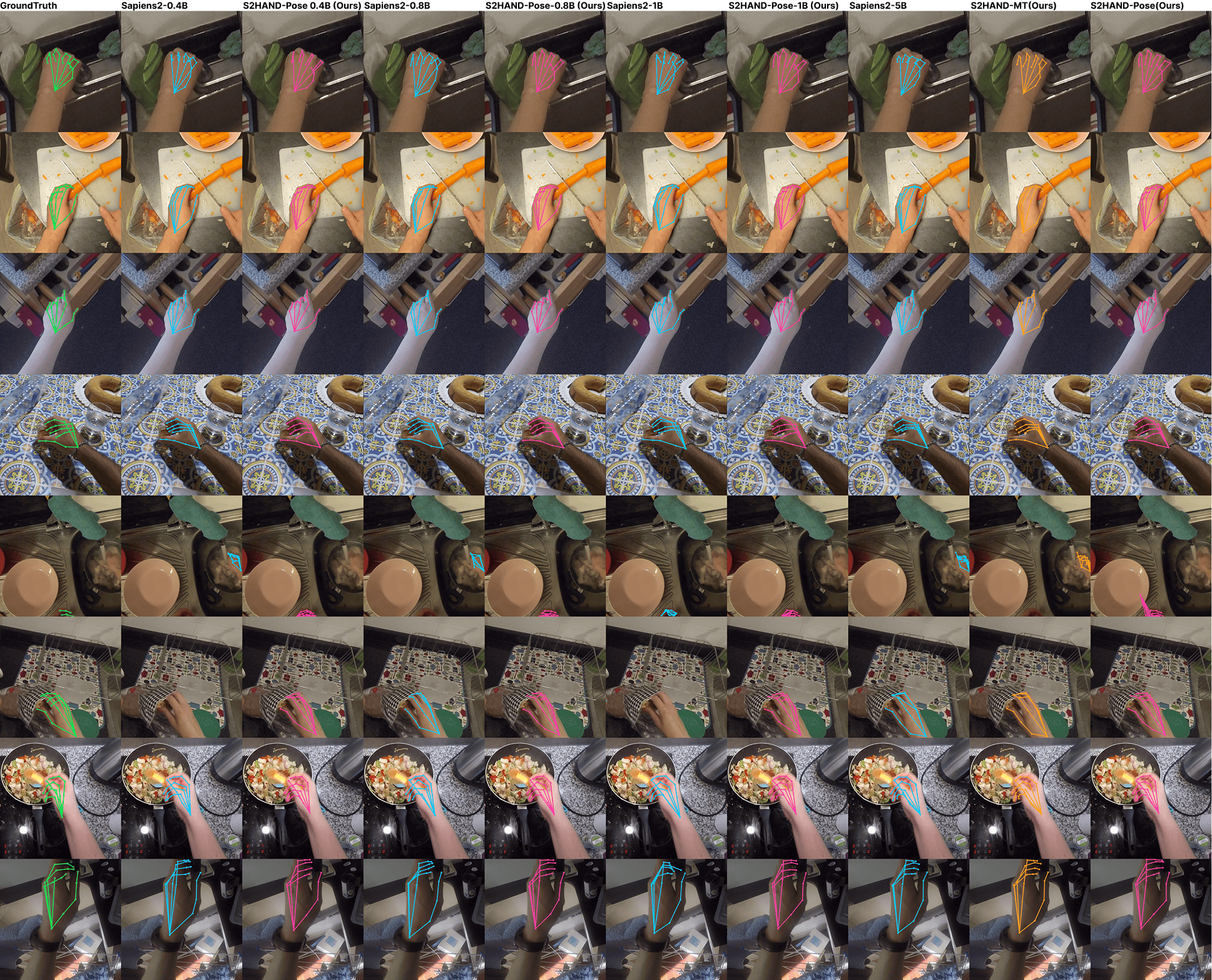}
  \caption{\textbf{Multi-scale VISOR comparison.} Eight deterministic VISOR
  TEST hands following the same scale-control ordering and fixed per-row rendering
  protocol as Fig.~\ref{fig:app-scale-newdays}.}
  \label{fig:app-scale-visor}
\end{figure*}
\clearpage
\begin{figure*}[p]
  \centering
  \includegraphics[width=\textwidth]{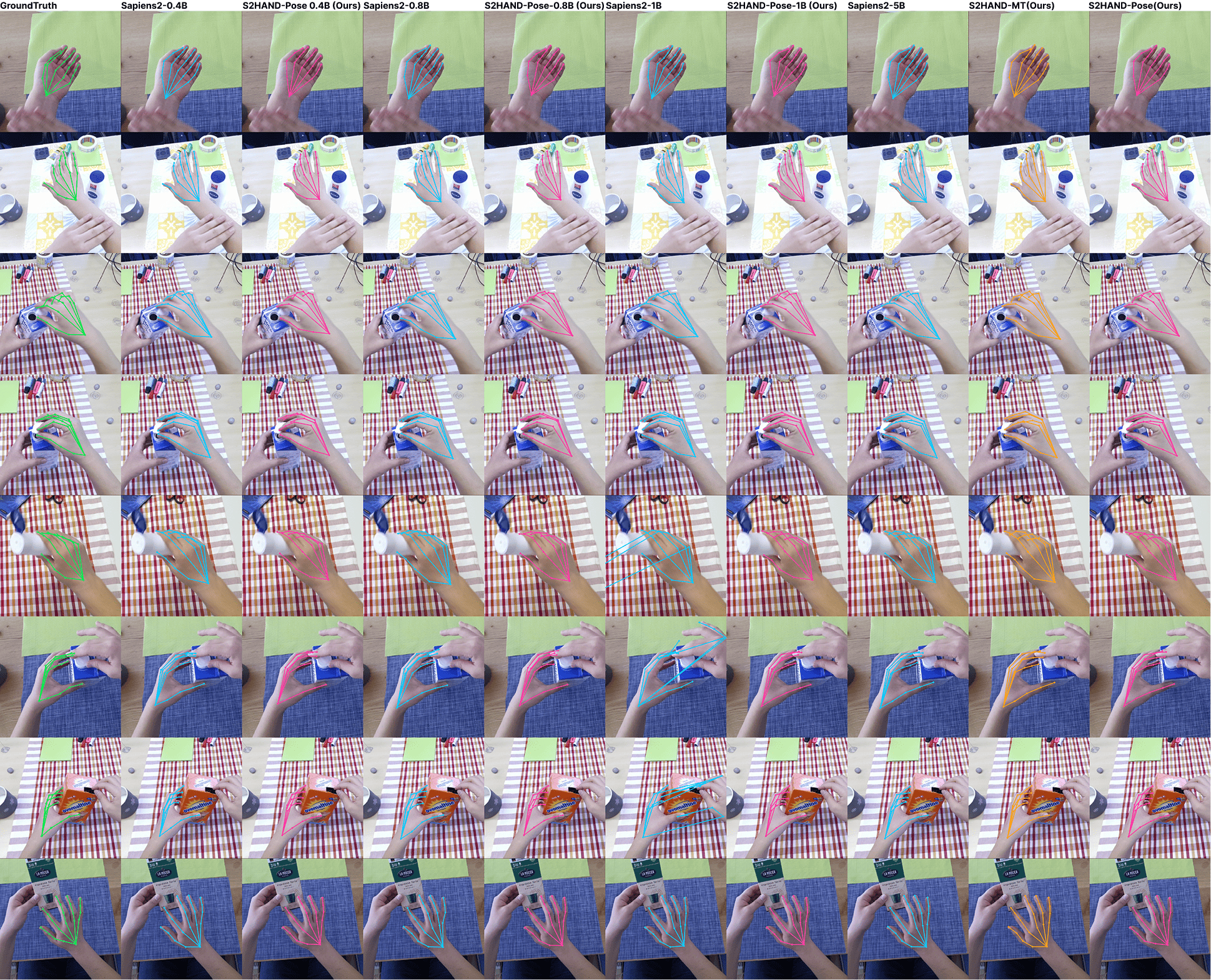}
  \caption{\textbf{Multi-scale H2O comparison.} Eight deterministic H2O TEST
  hands under the fixed ten-column scale-control ordering.}
  \label{fig:app-scale-h2o}
\end{figure*}
\clearpage
\begin{figure*}[p]
  \centering
  \includegraphics[width=\textwidth]{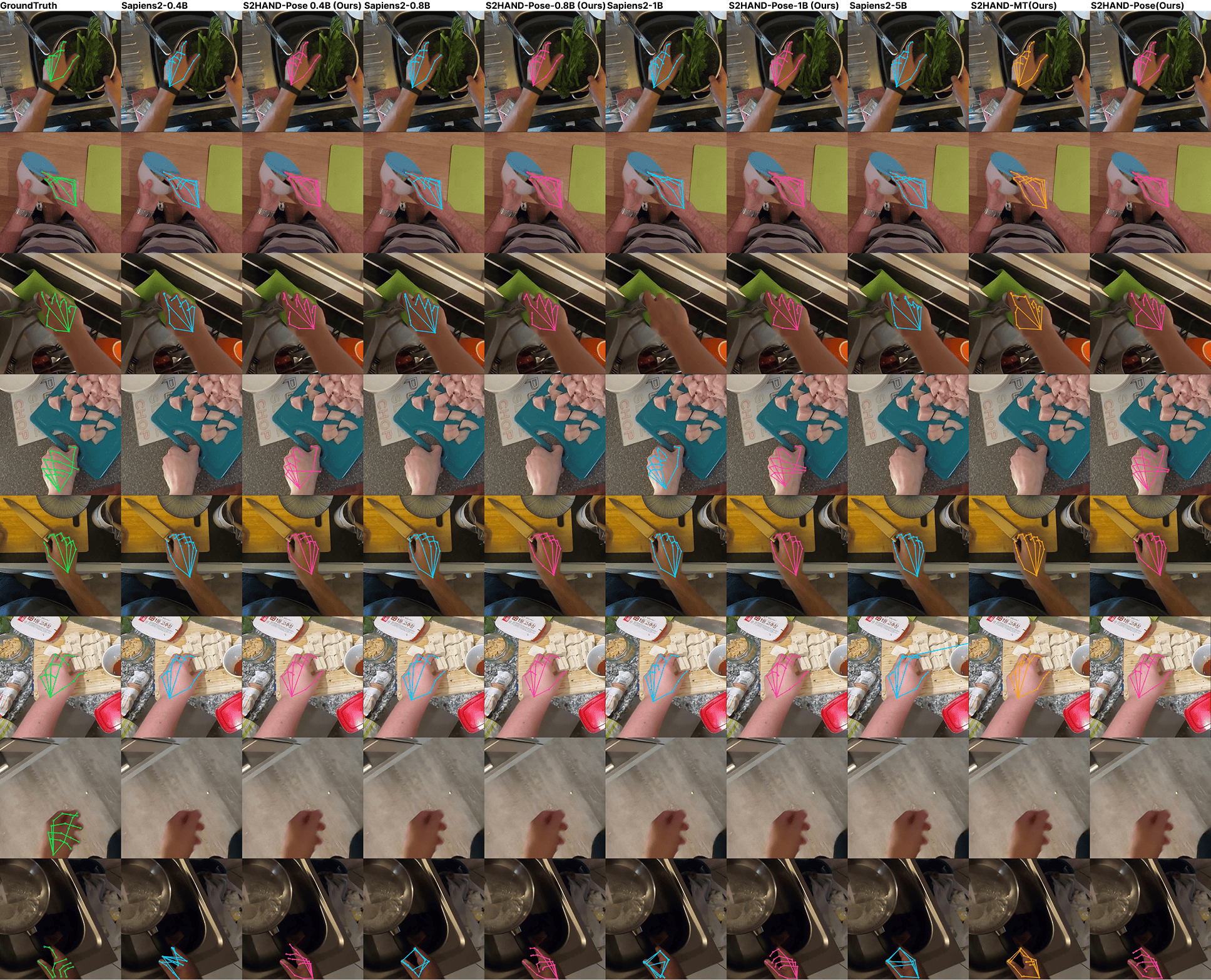}
  \caption{\textbf{Multi-scale EPICK comparison.} Eight deterministic EPICK
  VAL hands under the fixed ten-column scale-control ordering.}
  \label{fig:app-scale-epick}
\end{figure*}
\clearpage
\begin{figure*}[p]
  \centering
  \includegraphics[width=\textwidth]{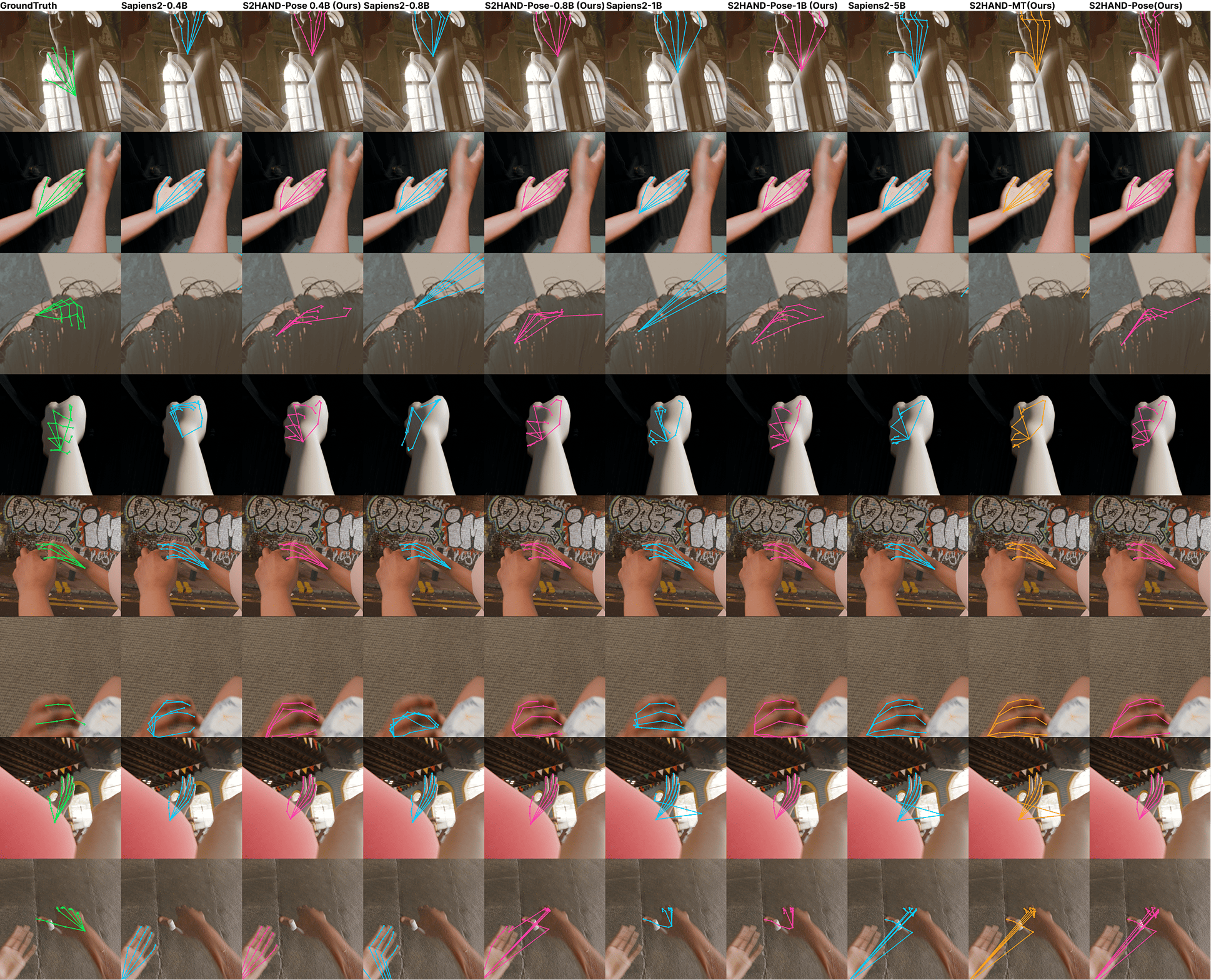}
  \caption{\textbf{Multi-scale H2SMV-FPV comparison.} Eight deterministic
  H2SMV-FPV TEST hands under the fixed ten-column scale-control ordering.}
  \label{fig:app-scale-h2smv}
\end{figure*}
\clearpage
\begin{figure*}[p]
  \centering
  \includegraphics[width=\textwidth]{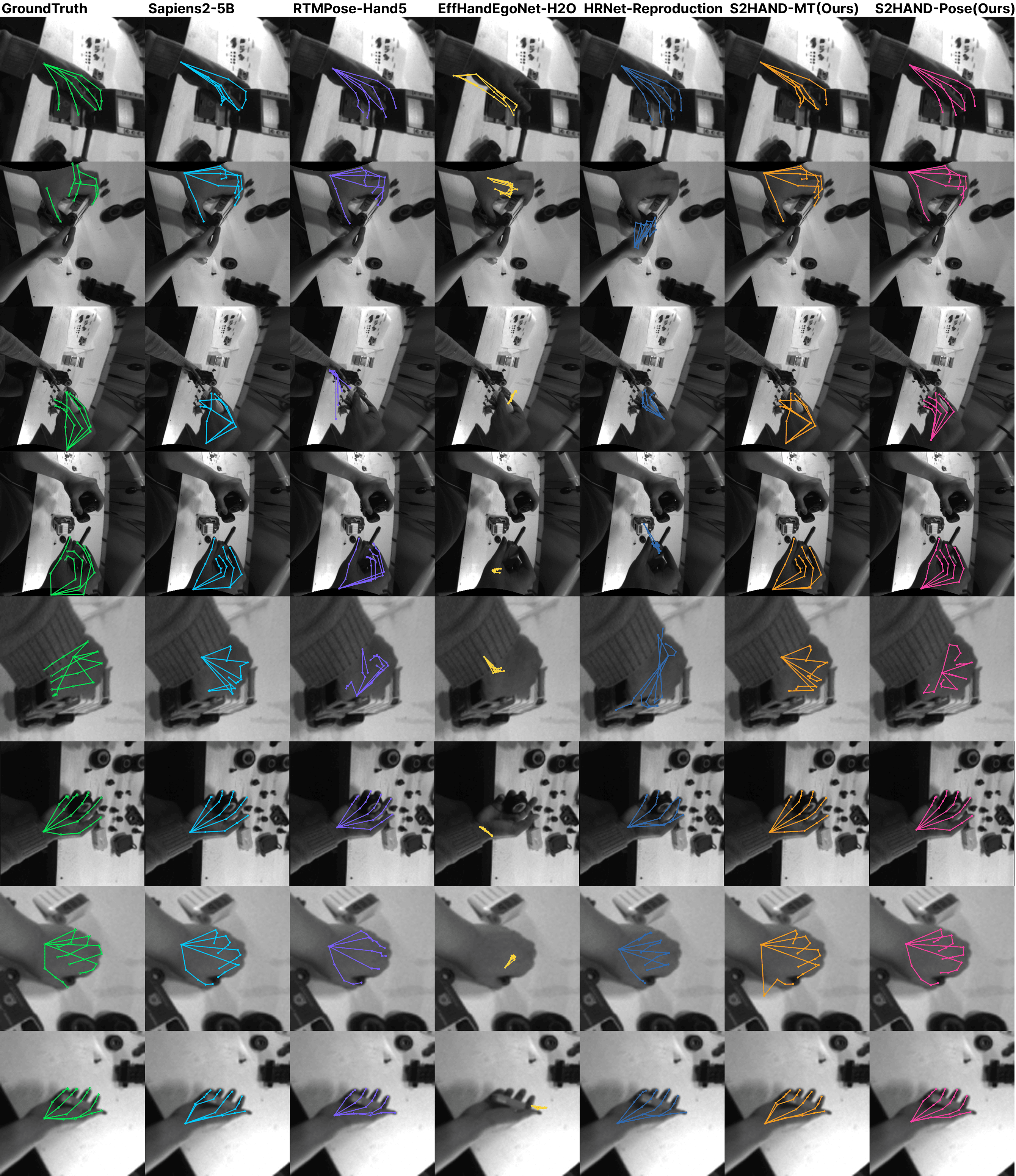}
  \caption{\textbf{AssemblyHands qualitative audit.} Eight deterministic
  AssemblyHands samples under the fixed seven-column main-model
  ordering.}
  \label{fig:app-main-assemblyhands}
\end{figure*}
\clearpage
\begin{figure*}[p]
  \centering
  \includegraphics[width=\textwidth]{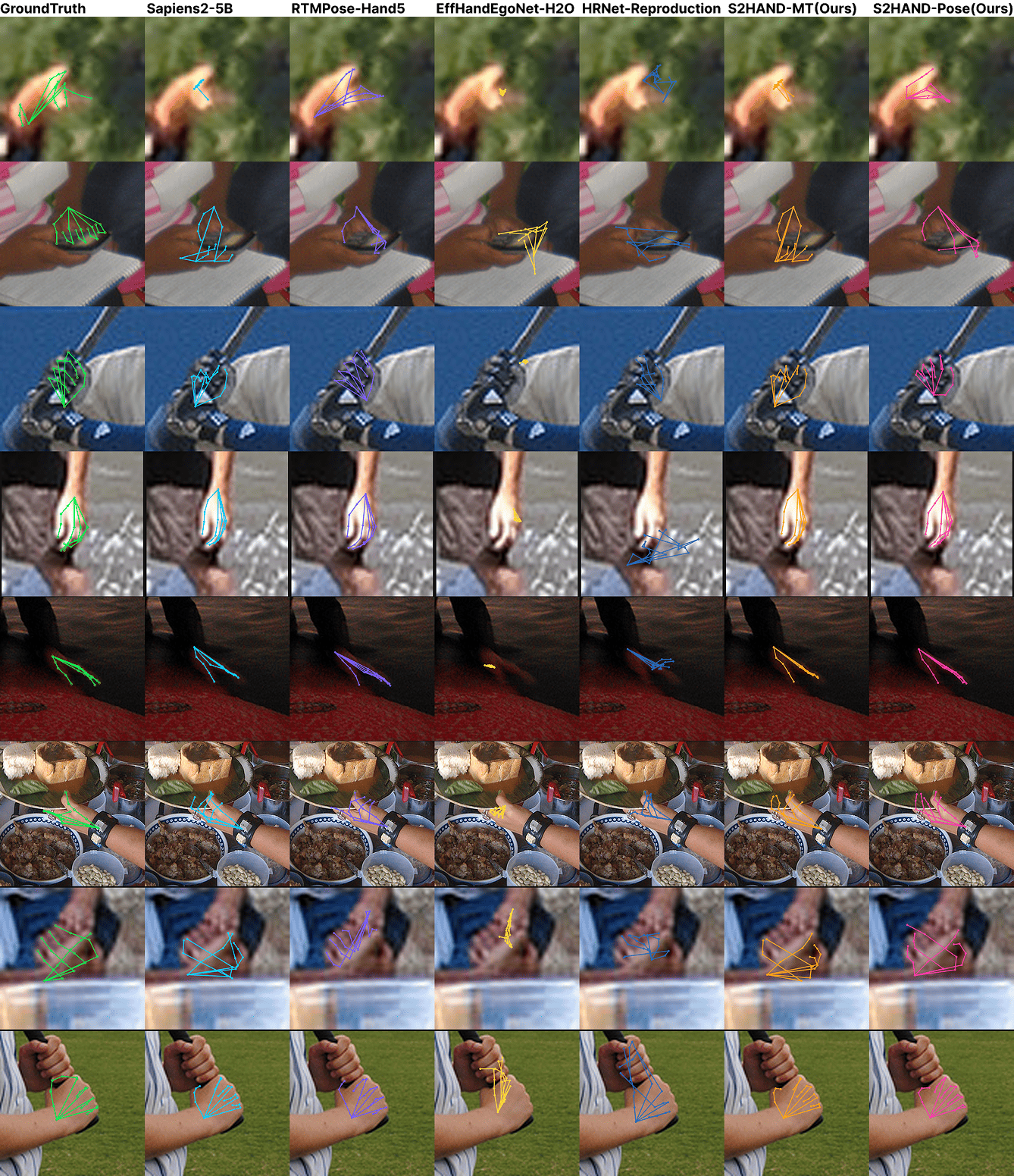}
  \caption{\textbf{COCO qualitative audit.} Eight deterministic COCO
  WholeBody samples under the fixed seven-column main-model ordering.}
  \label{fig:app-main-coco}
\end{figure*}
\clearpage
\begin{figure*}[p]
  \centering
  \includegraphics[width=\textwidth]{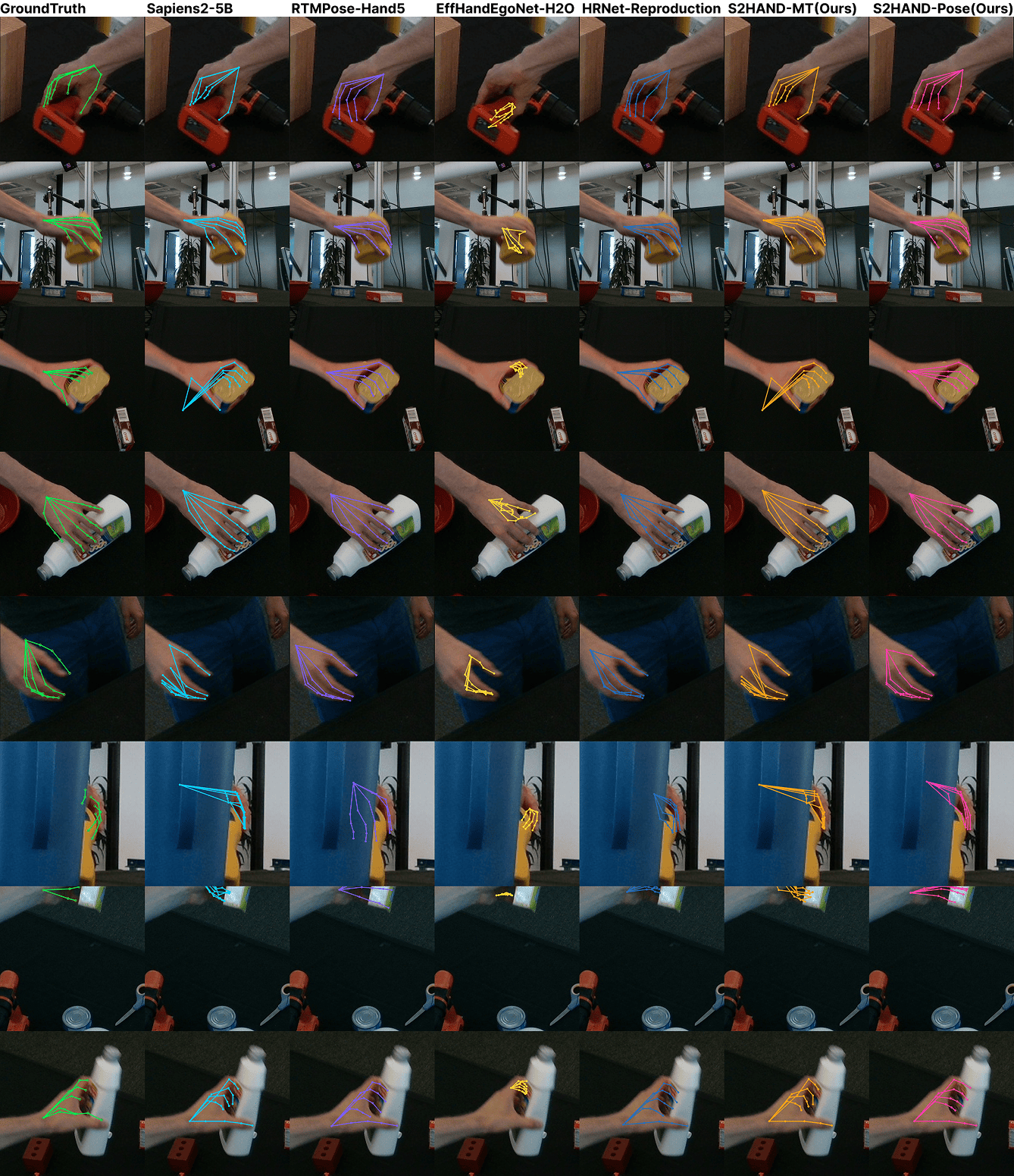}
  \caption{\textbf{DexYCB qualitative audit.} Eight deterministic DexYCB
  samples under the fixed seven-column main-model ordering.}
  \label{fig:app-main-dexycb}
\end{figure*}
\clearpage
\begin{figure*}[p]
  \centering
  \includegraphics[width=\textwidth]{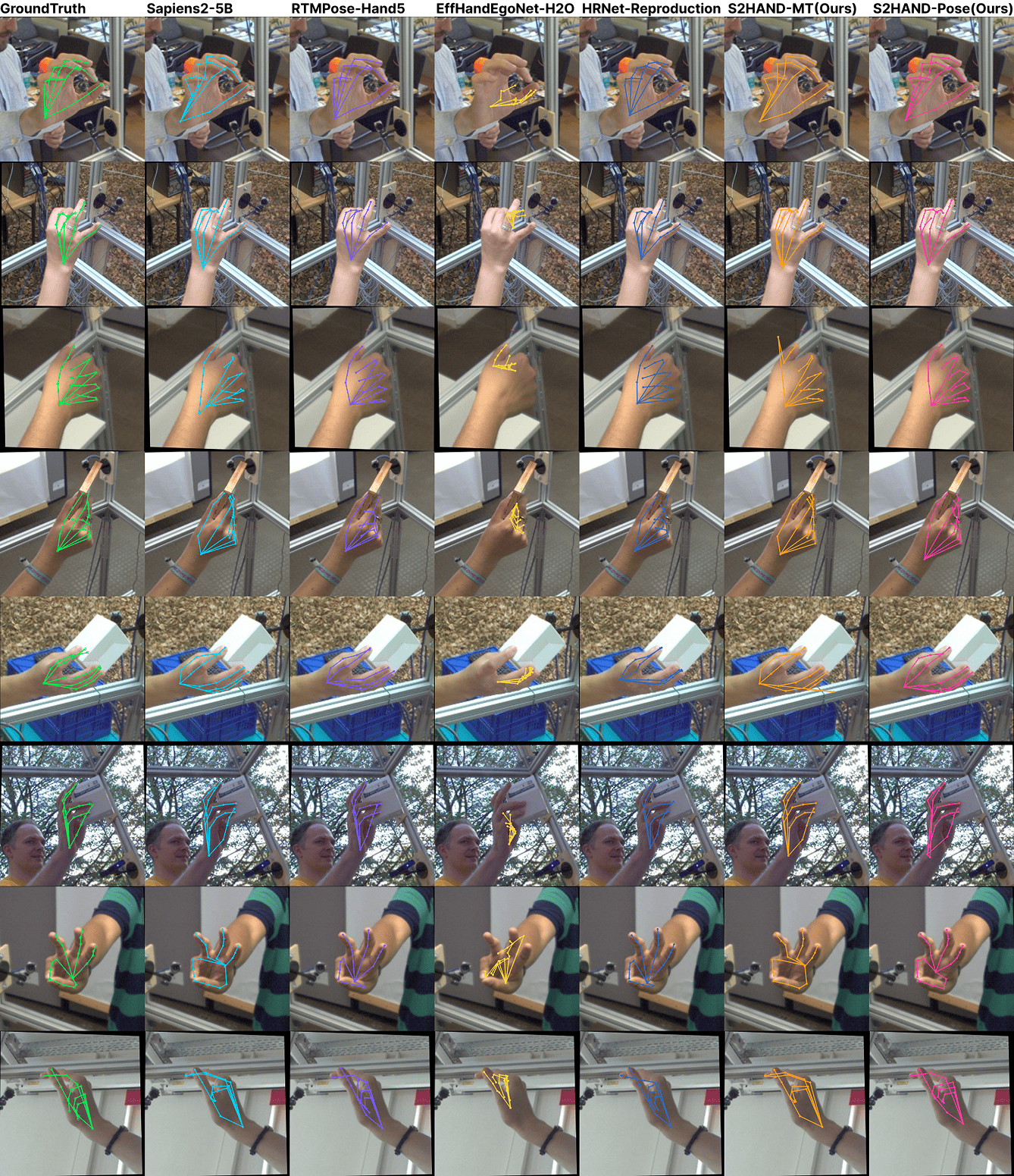}
  \caption{\textbf{FreiHAND qualitative audit.} Eight deterministic FreiHAND
  samples under the fixed seven-column main-model ordering.}
  \label{fig:app-main-freihand}
\end{figure*}
\clearpage
\begin{figure*}[p]
  \centering
  \includegraphics[width=\textwidth]{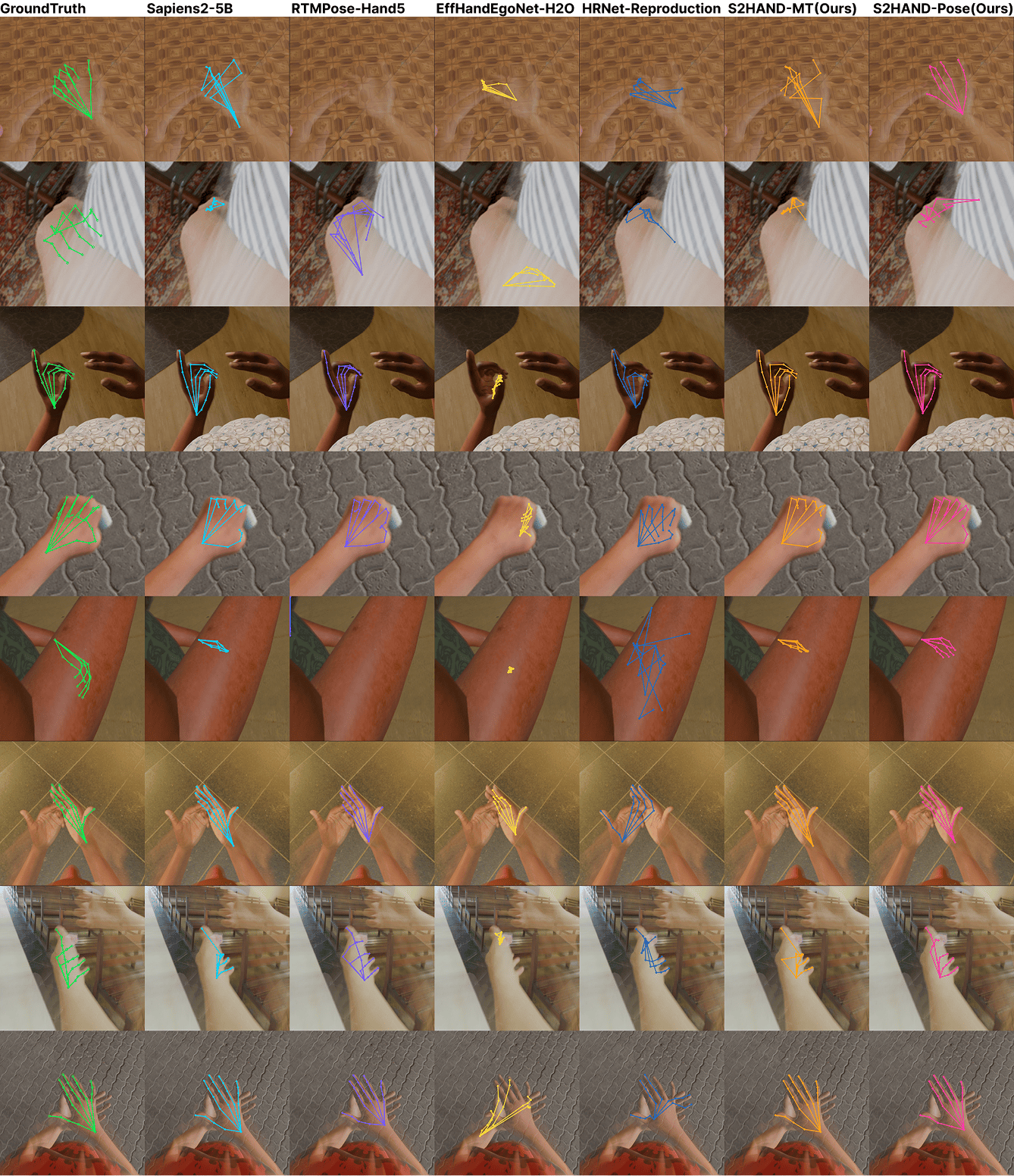}
  \caption{\textbf{H2SMV-FPV qualitative audit.} Eight deterministic
  H2SMV-FPV samples under the fixed seven-column main-model ordering.}
  \label{fig:app-main-h2smv}
\end{figure*}
\clearpage
\begin{figure*}[p]
  \centering
  \includegraphics[width=\textwidth]{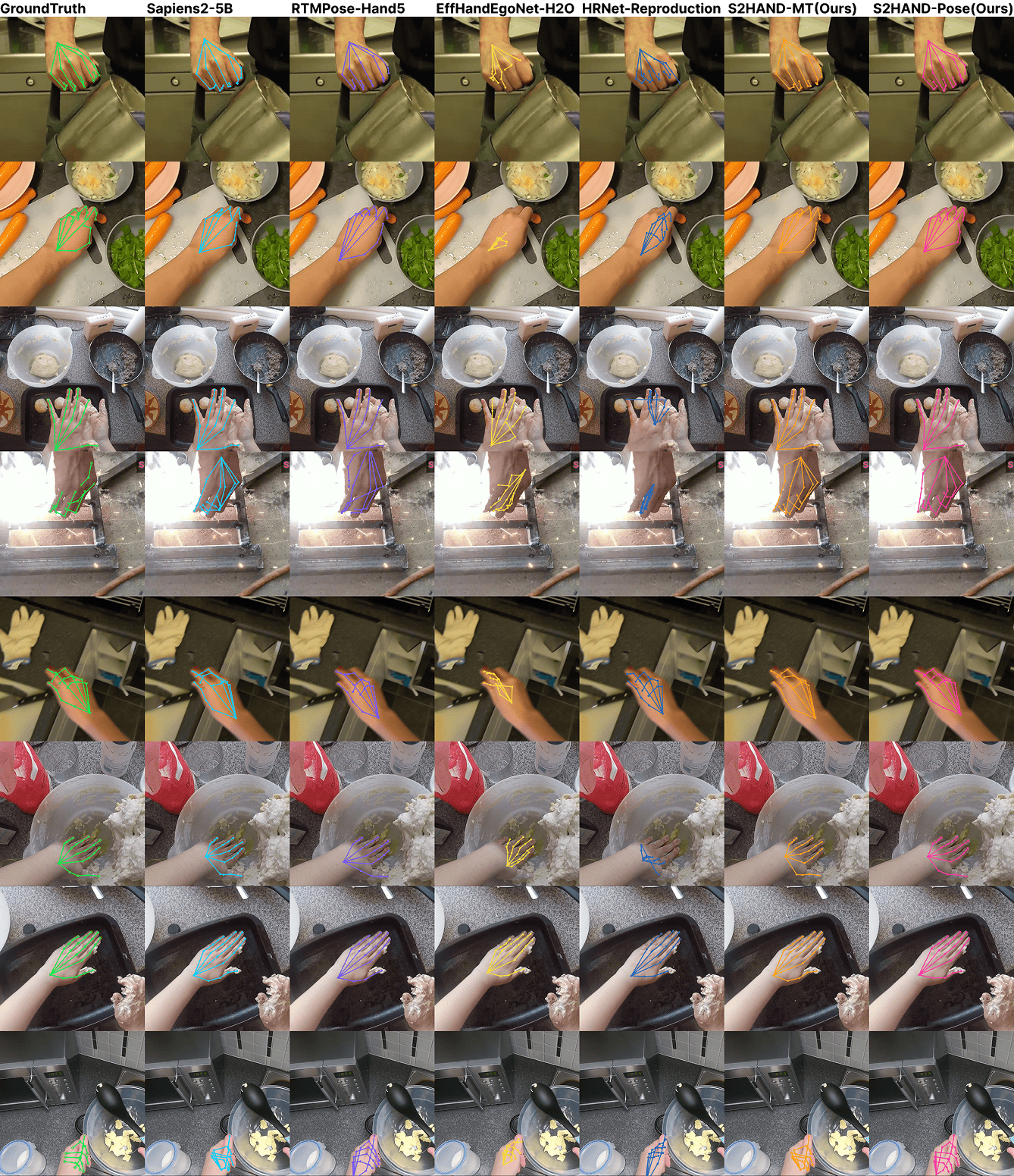}
  \caption{\textbf{HInt qualitative audit.} Eight deterministic HInt samples
  under the fixed seven-column main-model ordering.}
  \label{fig:app-main-hint}
\end{figure*}
\clearpage
\begin{figure*}[p]
  \centering
  \includegraphics[width=0.88\textwidth]{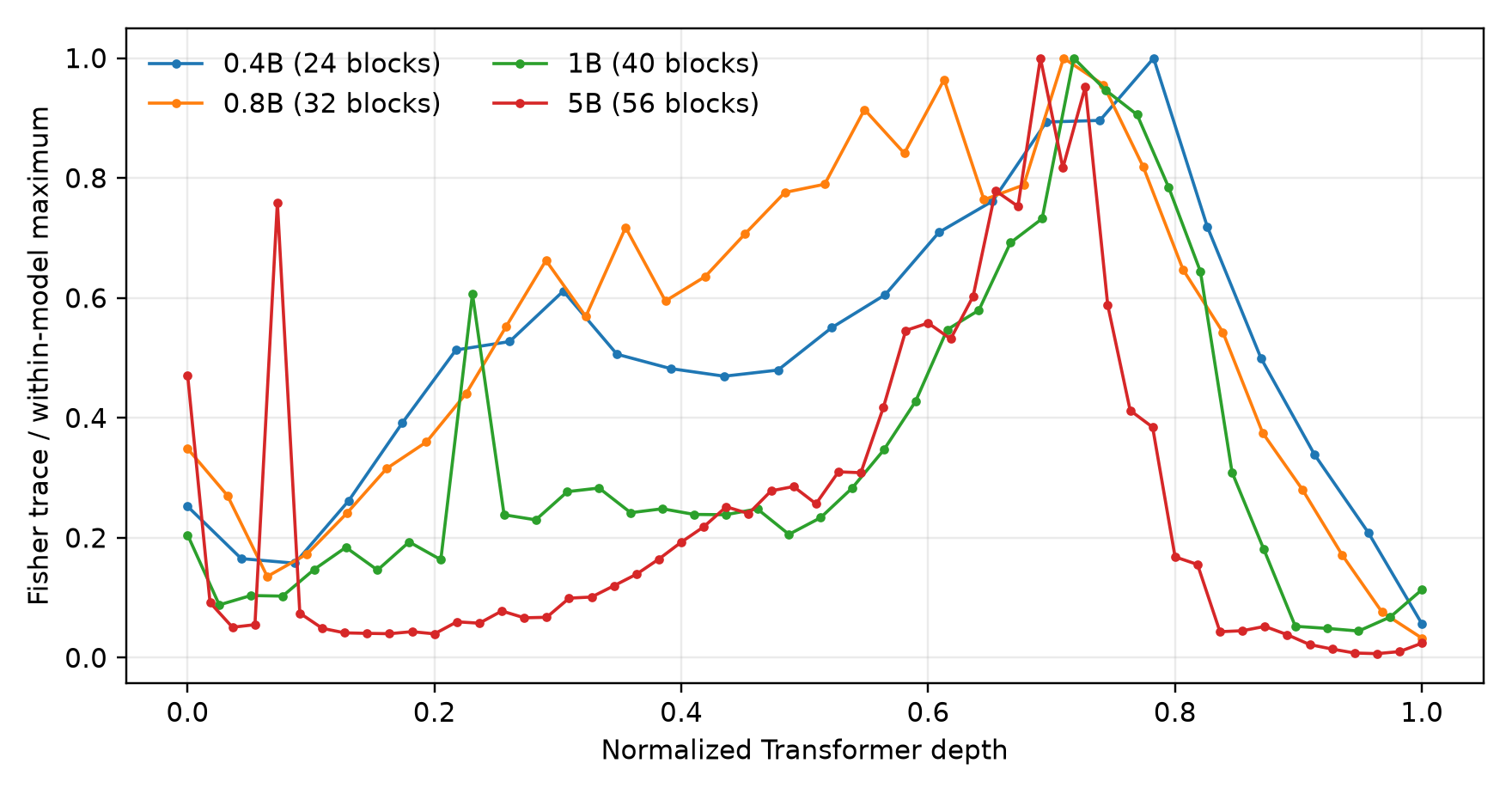}
  \caption{\textbf{Fisher profiles across the Sapiens2 model family.} Block
  depth and Fisher trace are normalized within each scale to compare the 24-,
  32-, 40-, and 56-block backbones independently of depth or
  parameter count. All curves use the same loss definition and eight cached
  HInt training batches without optimizer steps.}
  \label{fig:fisher-model-family}
\end{figure*}
\begin{figure*}[p]
  \centering
  \includegraphics[width=\textwidth]{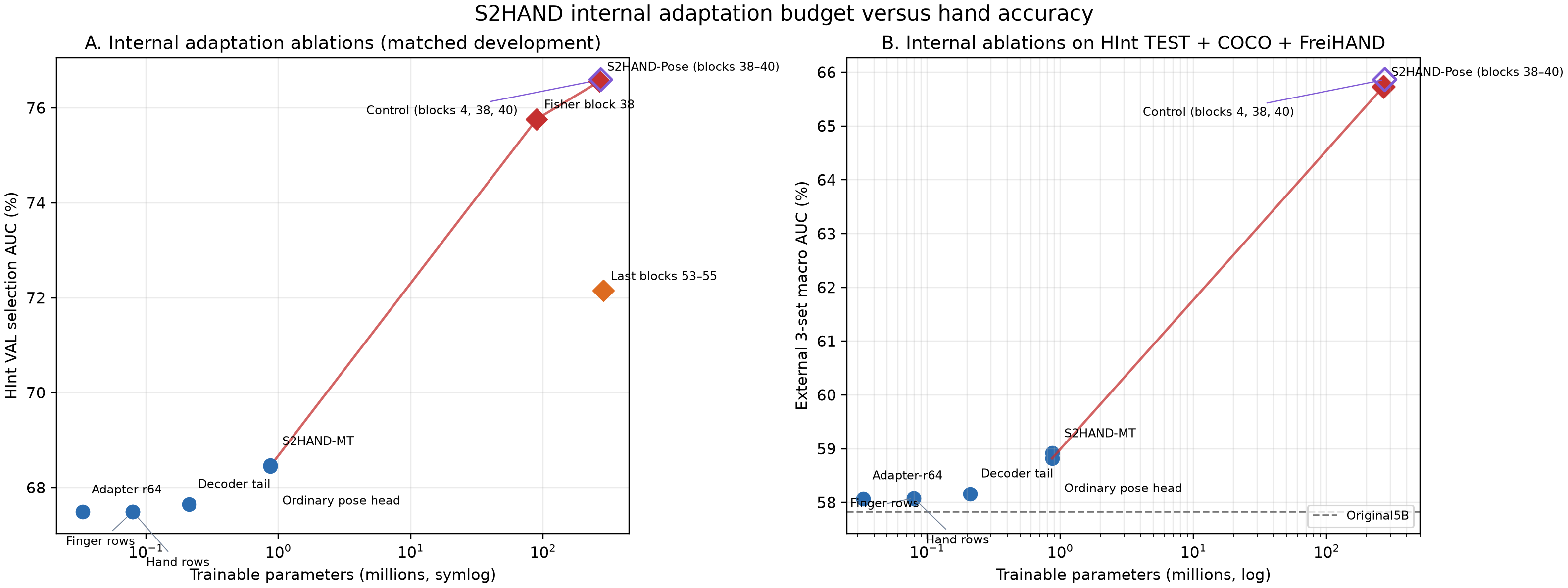}
  \caption{\textbf{Accuracy--modification trade-off.} HInt validation AUC
  across adaptation scopes and trainable-parameter budgets.}
  \label{fig:accuracy-budget}
\end{figure*}
\clearpage
\begin{figure}[p]
  \centering
  \includegraphics[width=\columnwidth]{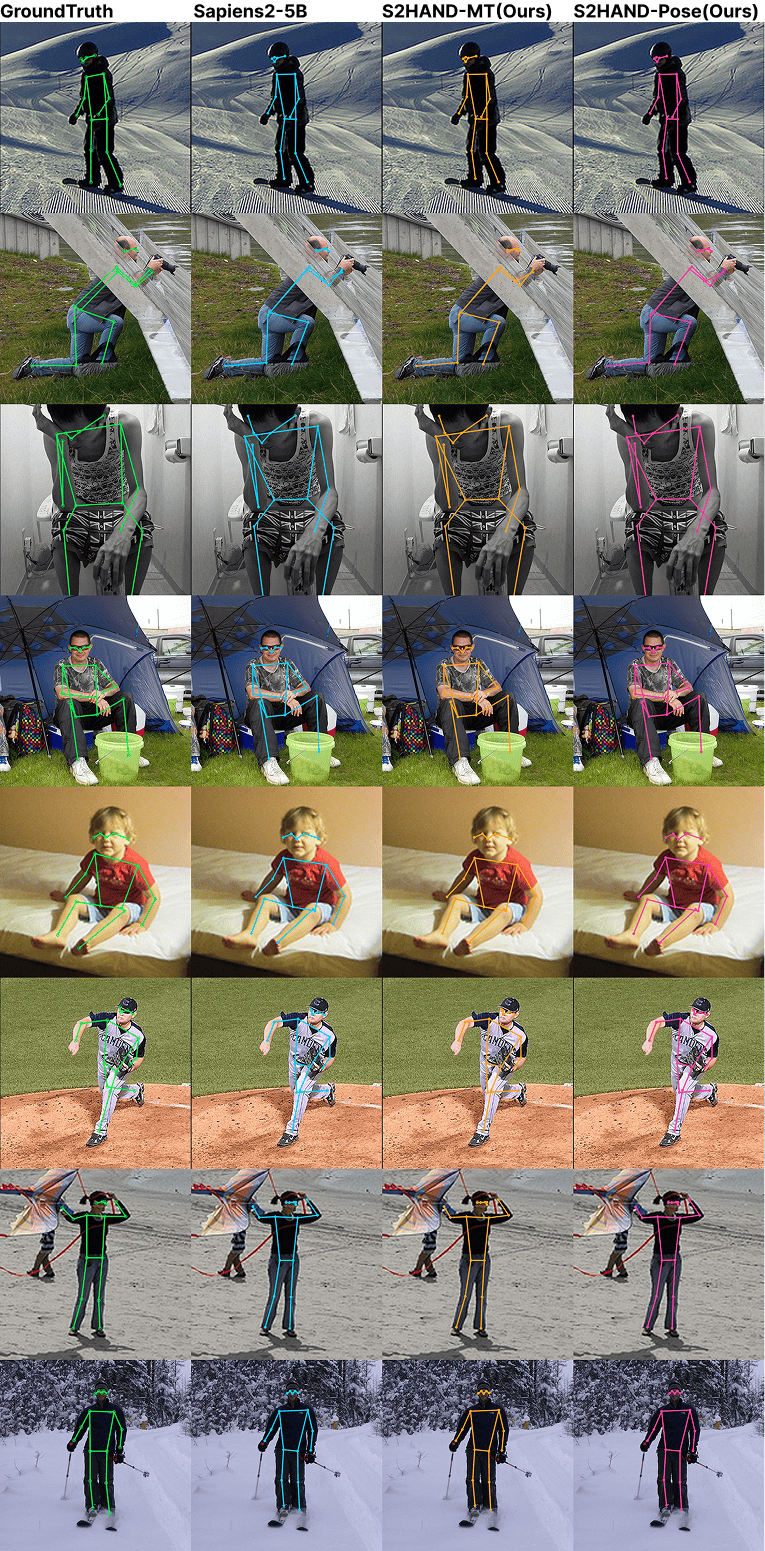}
  \caption{\textbf{Body-pose preservation on neutral cases.} Eight COCO
  WholeBody samples comparing GT, Public Sapiens2-5B, S2HAND-MT, and
  S2HAND-Pose under a fixed rendering protocol.}
  \label{fig:app-body-neutral}
\end{figure}
\begin{figure}[p]
  \centering
  \includegraphics[width=\columnwidth]{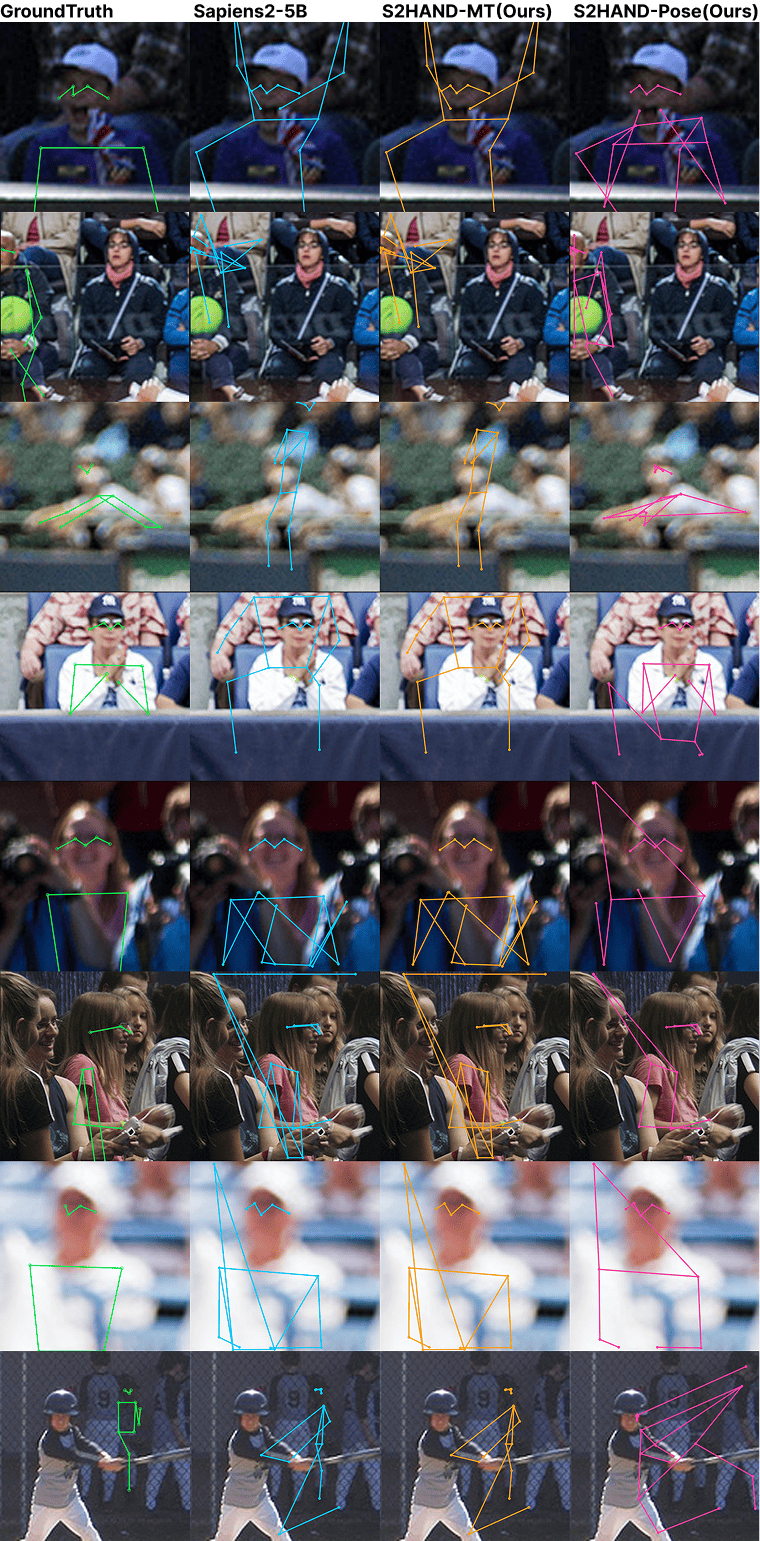}
  \caption{\textbf{Body-pose preservation on symmetric tail cases.} Eight
  cases from the frozen quantitative audit, following the four-column
  ordering in Fig.~\ref{fig:app-body-neutral}.}
  \label{fig:app-body-tail}
\end{figure}
\clearpage
\begin{figure}[p]
  \centering
  \includegraphics[width=\columnwidth]{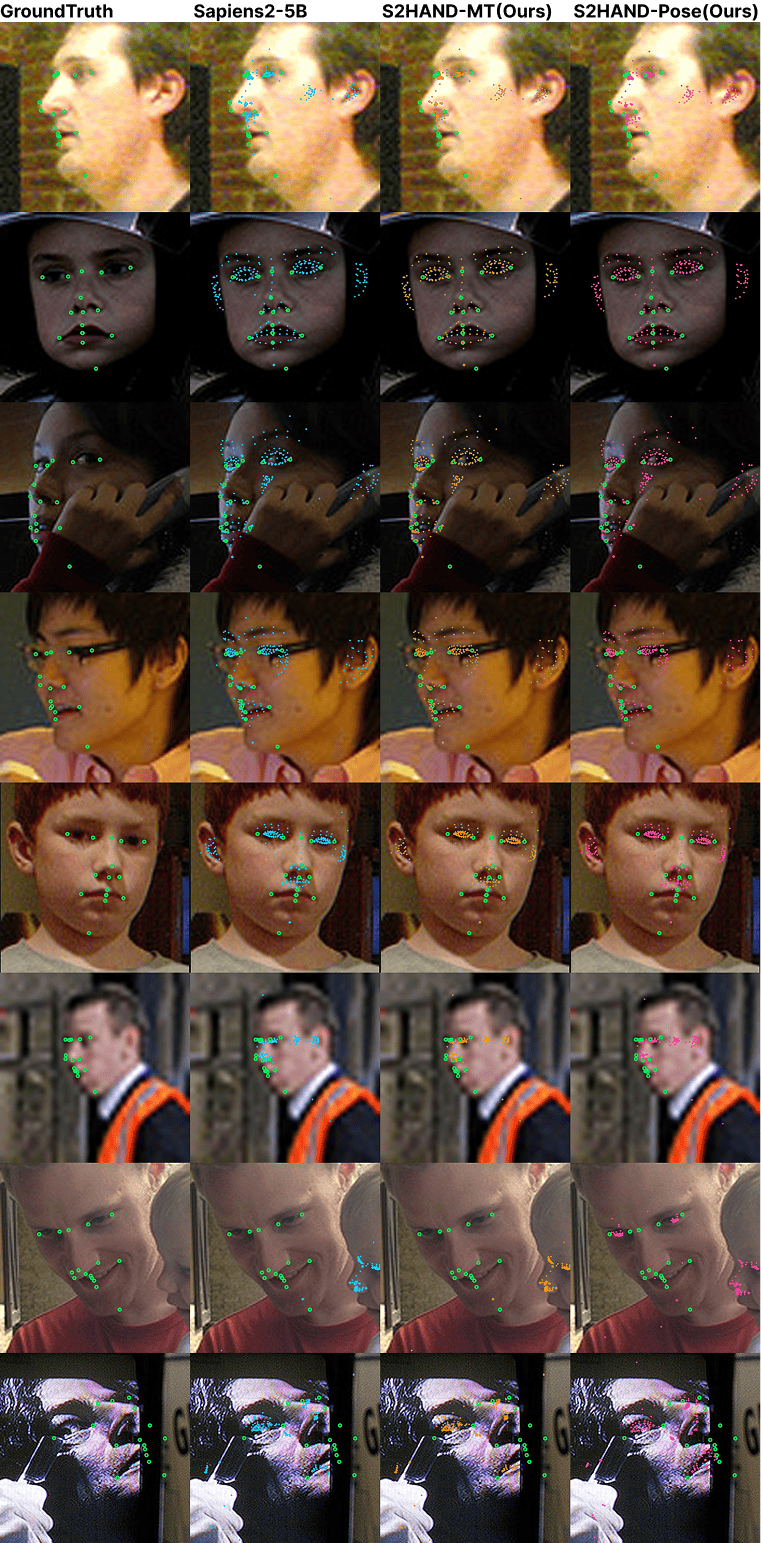}
  \caption{\textbf{Face-output preservation on a mixed audit set.} Eight COCO
  WholeBody instances comparing the 16 shared semantic anchors and decoded
  Sapiens2 dense-face outputs across the three 5B models.}
  \label{fig:app-face-mixed}
\end{figure}
\begin{figure}[p]
  \centering
  \includegraphics[width=\columnwidth]{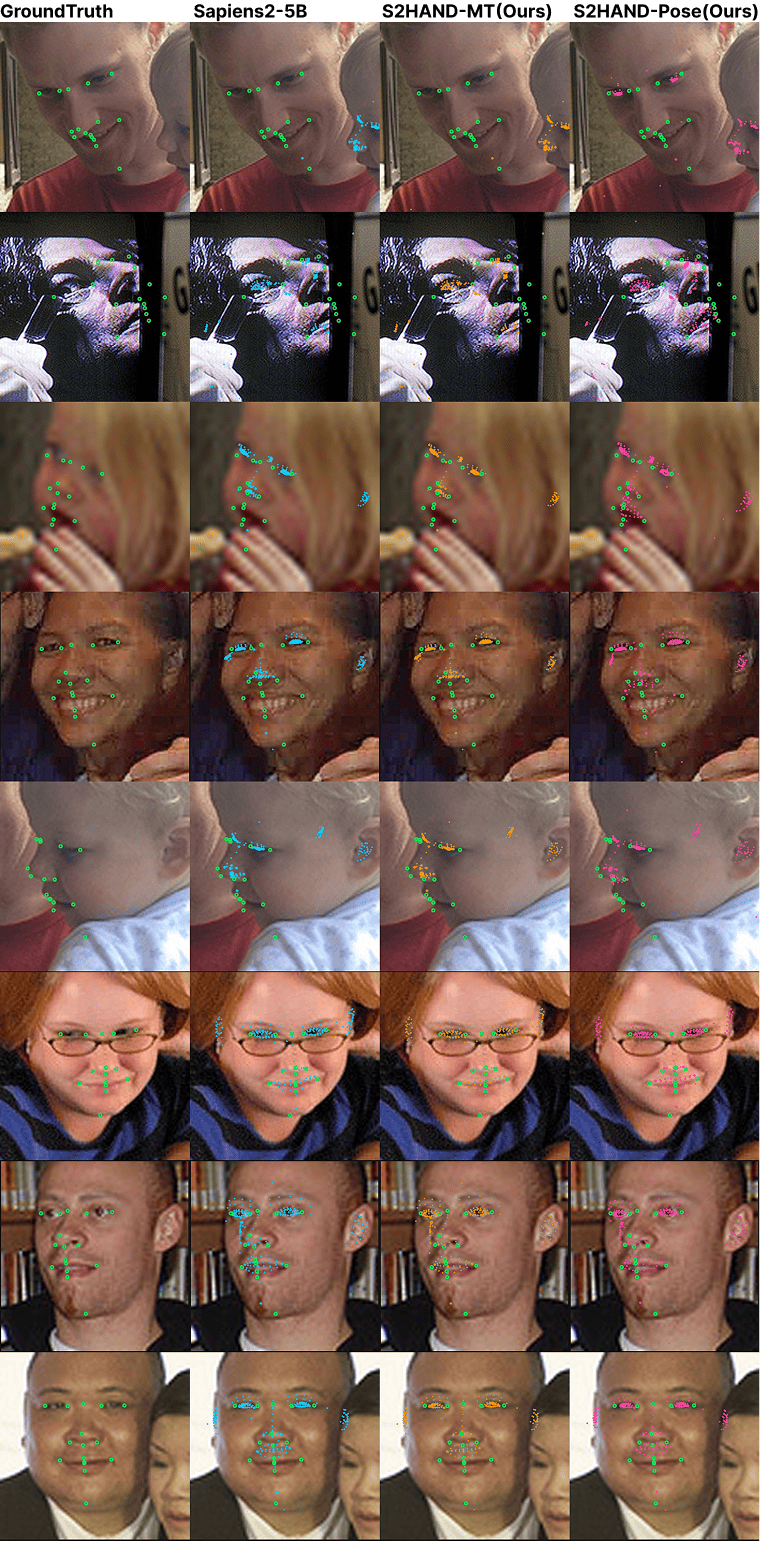}
  \caption{\textbf{Face-output preservation on symmetric tail cases.} Eight
  cases from the frozen teacher-relative drift audit, following the
  four-column ordering in Fig.~\ref{fig:app-face-mixed}.}
  \label{fig:app-face-tail}
\end{figure}
\clearpage
\begin{figure*}[p]
  \centering
  \includegraphics[width=\textwidth]{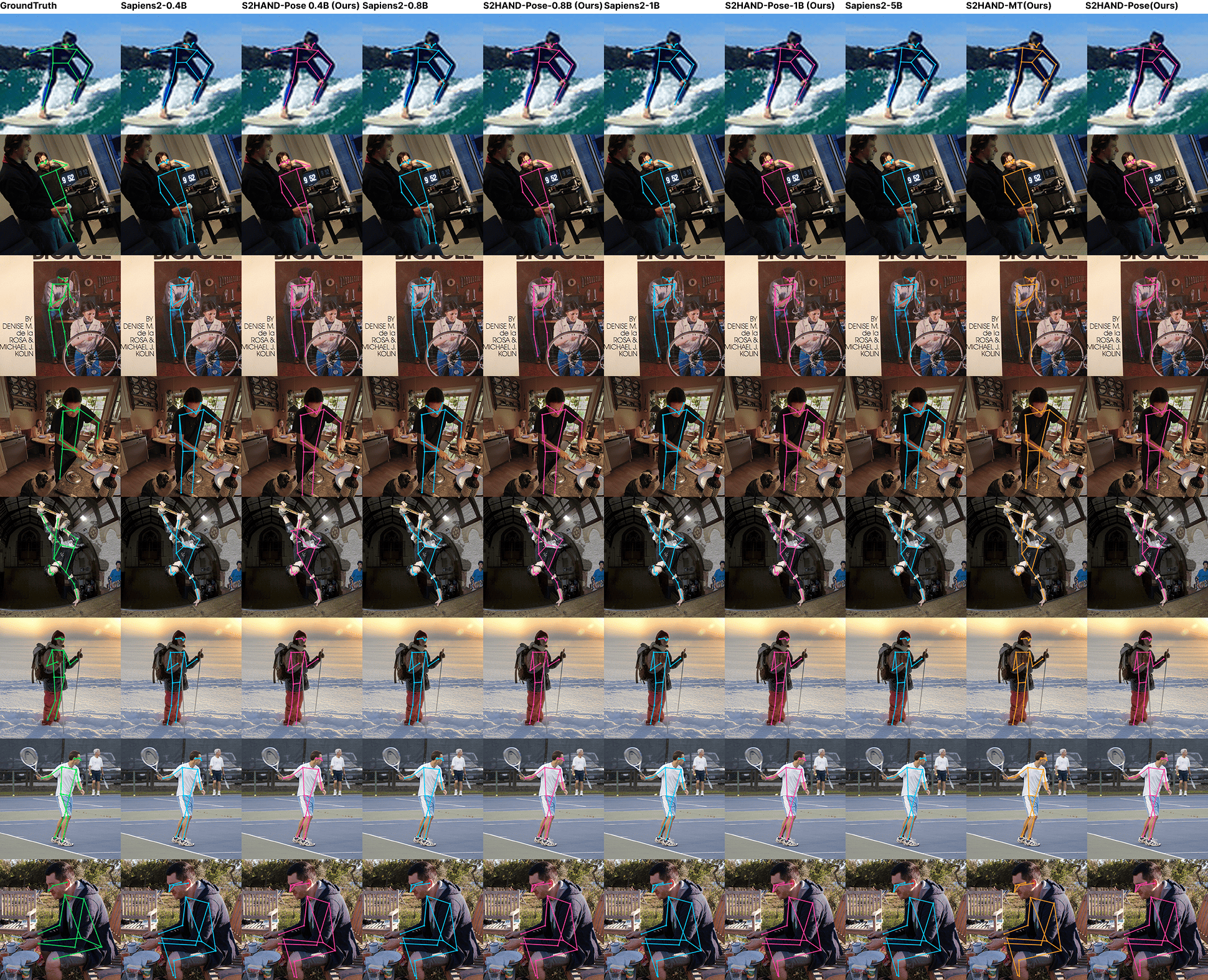}
  \caption{\textbf{Multi-scale body-output audit.} Eight COCO WholeBody
  instances comparing the public and adapted 0.4B, 0.8B, 1B, and 5B models
  (including S2HAND-MT at 5B).}
  \label{fig:app-scale-body}
\end{figure*}
\clearpage
\begin{figure*}[p]
  \centering
  \includegraphics[width=\textwidth]{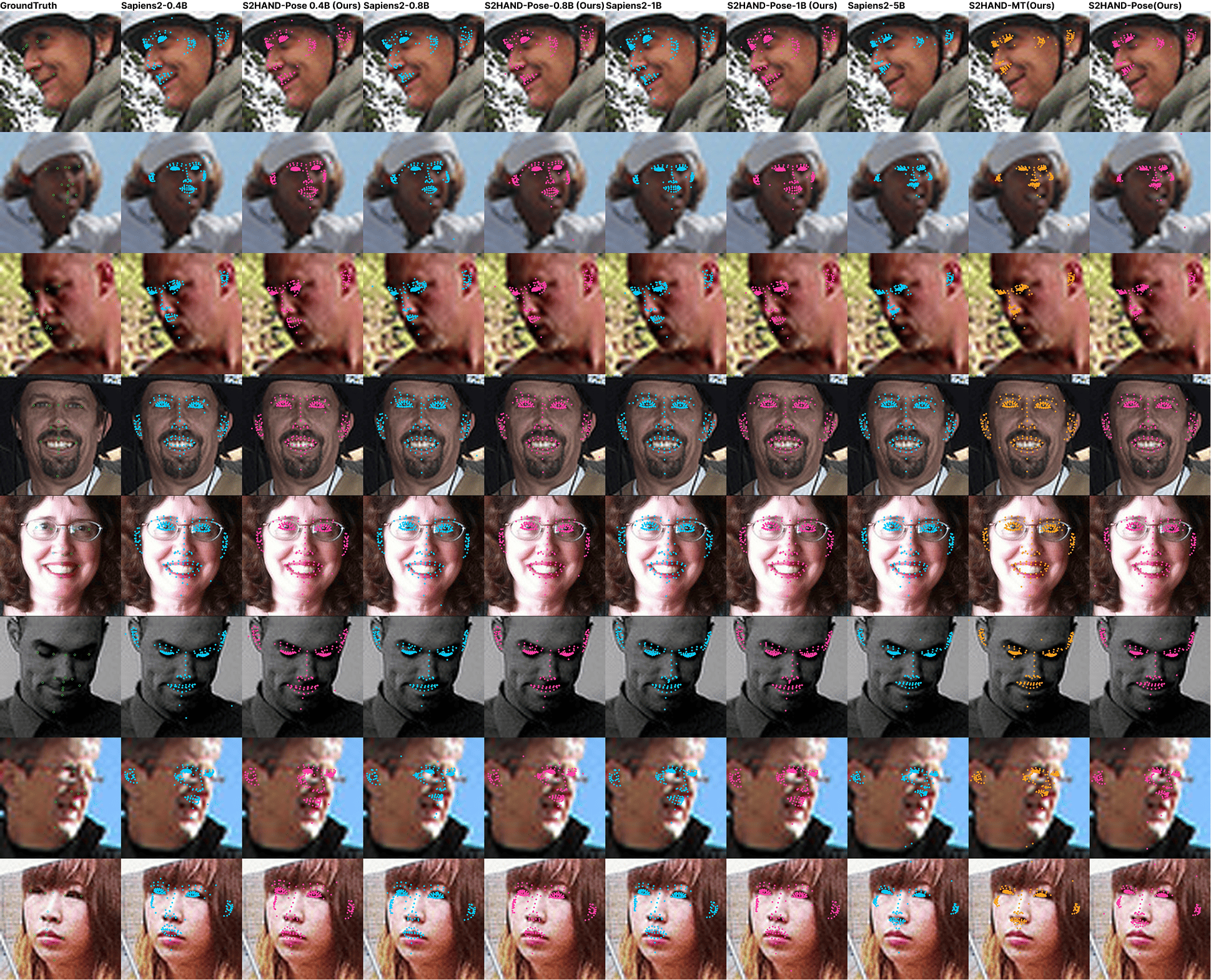}
  \caption{\textbf{Multi-scale face-output audit.} Eight COCO WholeBody
  instances following the same ten-column ordering as the multi-scale hand
  plates. GT markers show the 16 shared semantic anchors, while predictions
  display decoded Sapiens2 dense-face landmarks.}
  \label{fig:app-scale-face}
\end{figure*}
\clearpage
\begin{figure*}[p]
  \centering
  \includegraphics[width=\textwidth]{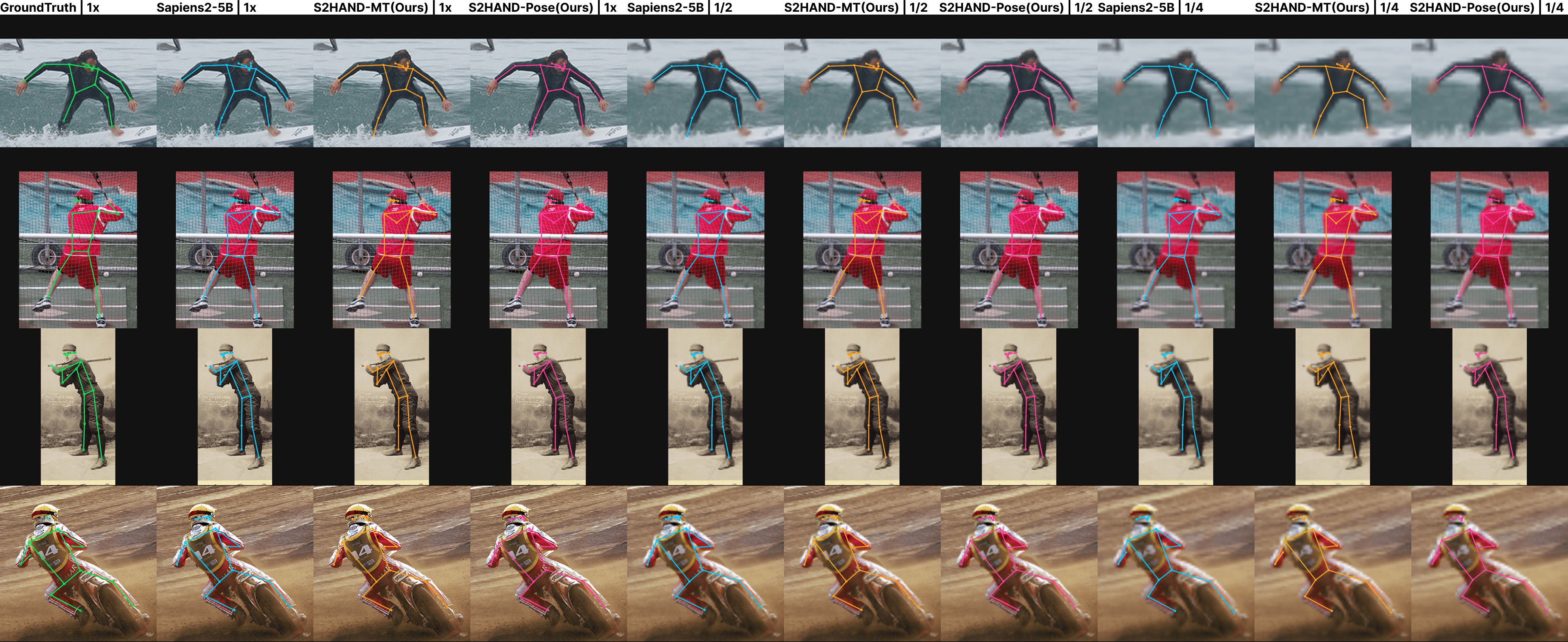}
  \caption{\textbf{Body robustness to reduced resolution.} Annotated
  samples are evaluated at native, $1/2$, and $1/4$ source
  resolutions (upsampled back to network input size prior to inference); in-plate labels report the model and
  per-sample NME.}
  \label{fig:body-resolution-app}
\end{figure*}
\begin{figure*}[p]
  \centering
  \includegraphics[width=\textwidth]{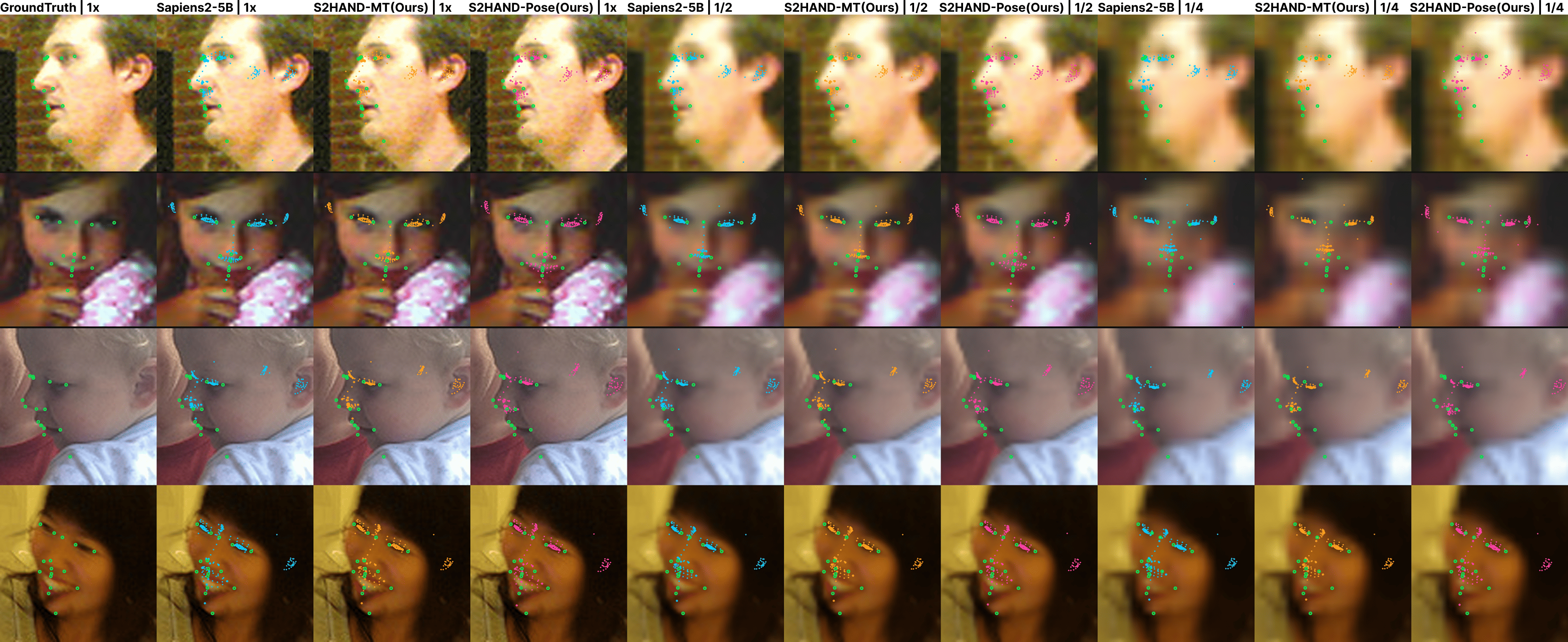}
  \caption{\textbf{Face robustness to reduced resolution.} Evaluated under the same native,
  $1/2$, and $1/4$ source-resolution protocol as
  Fig.~\ref{fig:body-resolution-app}; in-plate labels report the model
  and 16-anchor NME per sample.}
  \label{fig:face-resolution-app}
\end{figure*}
\clearpage
\begin{figure*}[p]
  \centering
  \includegraphics[width=\textwidth]{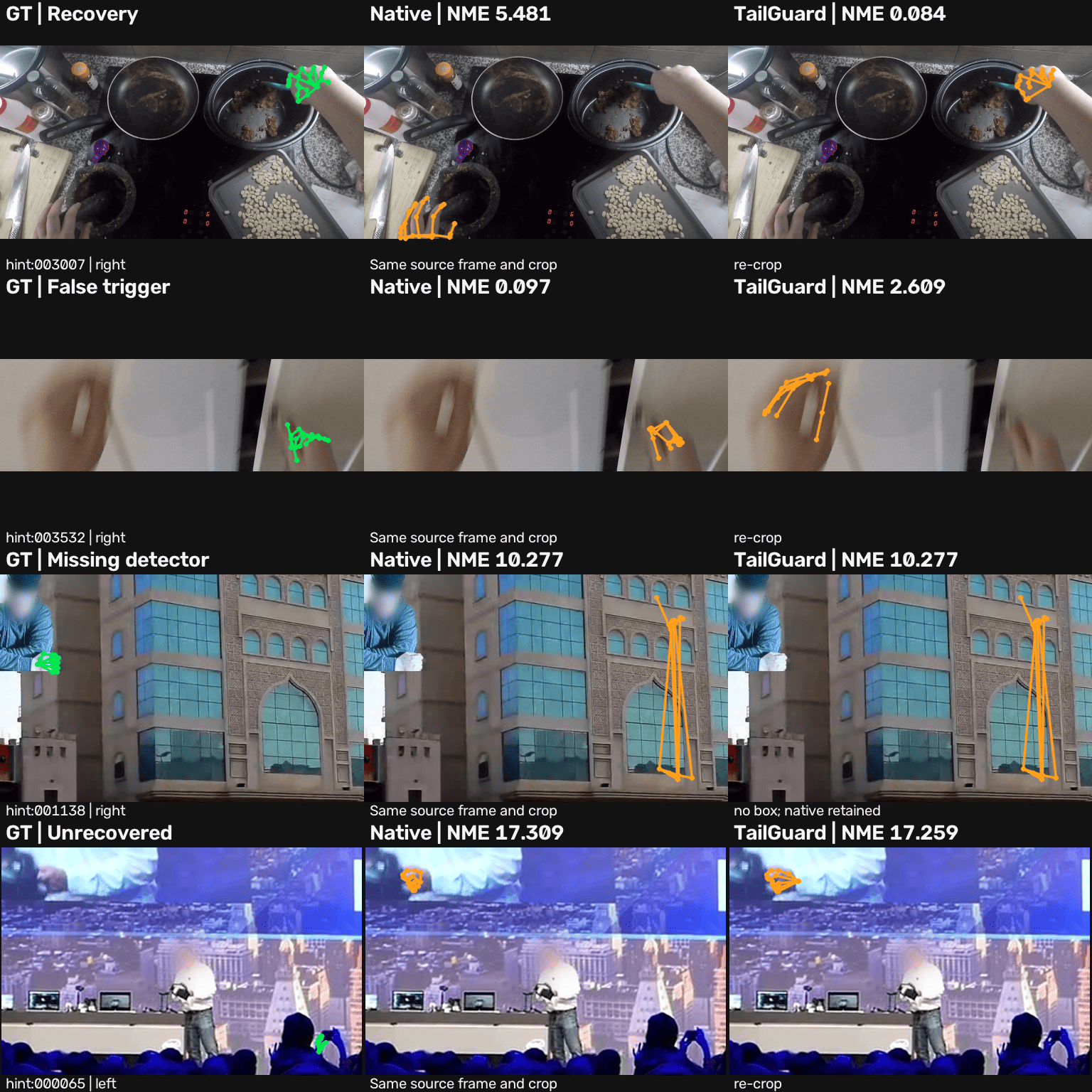}
  \caption{\textbf{TailGuard successes and failures under a common view.}
  Columns show GT, native S2HAND-MT, and S2HAND-MT with TailGuard across a fixed
  source frame, annotated hand side, visualization crop, and display scale per row
  (the display crop is for visualization only, not inference). Rows illustrate
  a successful recovery, a harmful re-crop, missing-detector fallback (which retains
  the native prediction rather than leaving an empty panel), and an unrecovered failure.
  Selected cases illustrate failure and recovery mechanisms rather than their empirical frequency.}
  \label{fig:tailguard-audit-app}
\end{figure*}
\clearpage